\documentclass[mscres,irr]{infthesis}

\course{Linguistics}
\project{Word Count: XXXX}

\usepackage{natbib}
\usepackage{graphicx}
\usepackage{booktabs}
\usepackage{multirow}
\usepackage[normalem]{ulem}
\usepackage{hyperref}
\usepackage[titletoc]{appendix}
\usepackage{subcaption}
\usepackage{amsfonts}
\usepackage{amssymb}
\usepackage{amsmath}

\useunder{\uline}{\ul}{}

\begin{document}

\begin{figure}
    \centering
    \includegraphics[width=0.3\textwidth]{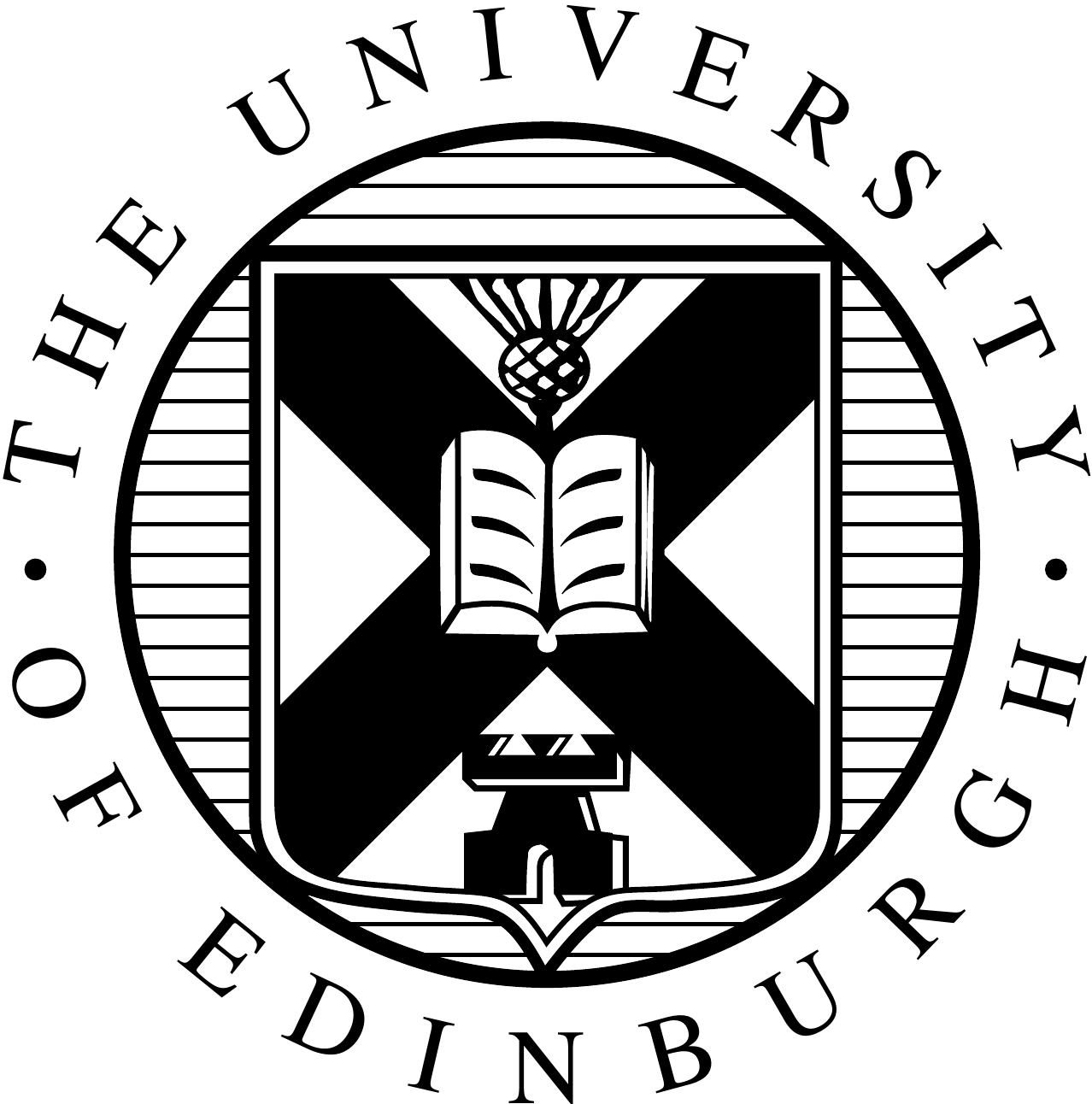}
\end{figure}
\title{Pre-trained Language Models as Re-Annotators}
\author{Chang Shu}



\abstract{%
 Annotation noise are widespread in datasets, but manually revising a flawed corpus is time-consuming and error-prone. Hence, given the prior knowledge in Pre-trained Language Models and the expected uniformity across all annotations, we attempt to reduce annotation noise in the corpus through two tasks automatically: (1) Annotation Inconsistency Detection that indicates the credibility of annotations, and (2) Annotation Error Correction that rectifies the abnormal annotations. 

We investigate how to acquire semantic sensitive annotation representations from Pre-trained Language Models, expecting to embed the examples with identical annotations to the mutually adjacent positions even without fine-tuning. We proposed a novel credibility score to reveal the likelihood of annotation inconsistencies based on the neighbouring consistency. Then, we fine-tune the Pre-trained Language Models based classifier with cross-validation for annotation correction. The annotation corrector is further elaborated with two approaches: (1) soft labelling by Kernel Density Estimation and (2) a novel distant-peer contrastive loss.

We study the re-annotation in relation extraction and create a new manually revised dataset, Re-DocRED, for evaluating document-level re-annotation. The proposed credibility scores show promising agreement with human revisions, achieving a Binary $F_1$ of 93.4 and 72.5 in detecting inconsistencies on TACRED and DocRED respectively. Moreover, the neighbour-aware classifiers based on distant-peer contrastive learning and uncertain labels achieve Macro $F_1$ up to 66.2 and 57.8 in correcting annotations on TACRED and DocRED respectively. These improvements are not merely theoretical: Rather, automatically denoised training sets demonstrate up to 3.6\% performance improvement for state-of-the-art relation extraction models, and the proposed framework is expected to be hundreds of times faster than the human re-annotators empirically.

}


\begin{preliminary}

\maketitle

\begin{acknowledgements}
\textit{This dissertation is dedicated to my father, Shengguo Shu. I wish you a happy birthday, and thank you for always being there.}

I am deeply grateful to my supervisors, Prof. Bonnie Webber, Dr. Beatrice Alex and Andreas Grivas, for bringing me to this exciting project and continuous support. I believe they are some of the best supervisors and NLP researchers on the planet, and it is such a great honour to work with them. Additionally, I would like to thank my personal tutor, Dr. Catherine Lai, for her help and advice during my master study. I also appreciate Luxi He for proofreading and Anda Zhou for discussion through this dissertation. 

I would also like to express gratitude to my previous supervisors, Dr. Rui Zhang, Dr. Tao Yu, Dr. Jian Qiu, and Prof. Zhiyuan Liu, for their past instructions and kindness in offering internship opportunities at Penn State University, Yale University, Alibaba Cloud and Tsinghua University. I also appreciate all my mentors and colleagues during these internships, Peng Shi, Jie Zhou, Taiyan Li, Yusen Zhang and Xiangyu Dong.

Finally, I would like to say my deepest thanks to my kith and kin. It was a miserable year for me physically and mentally, and I definitely would not make it without your endless support and love.

The Tao that can be told is not the eternal Tao. Though findings in this dissertation are ephemeral, I am thankful for the undiluted pleasure brought by this exploration.

\end{acknowledgements}

\standarddeclaration


\tableofcontents


\end{preliminary}


\chapter{Introduction}
Annotation noise is pervasive in datasets and becomes increasingly problematic as data-driven methods are increasingly incorporated into Natural Language Processing (NLP). This dissertation is the first to leverage prior knowledge in Pre-trained Language Models to detect annotation noise and inconsistencies and correct annotation errors. We introduce the prime motivation behind this project and outline the investigations we conducted to approach this problem. We also summarise our key contributions and the main contents of each chapter in the dissertation.
\section{Motivation}
Recent decades have witnessed  profound shifts in NLP research from symbolic methods to statistical techniques, and then to neural methods \citep{khurana2017natural}. Early research in NLP mainly relied on a finite set of hand-written rules that reflected common linguistic knowledge. In contrast, the latest paradigm of NLP involves letting neural models learn latent linguistic knowledge from vast amounts of data. As data-driven methods dominate NLP research, the importance of annotation quality is increasingly visible. However, most of the recent advances in NLP still focus on developing models with enhanced representation learning capability, underestimating the deterioration of model performance caused by annotation noise in training data \citep{DBLP:conf/coling/LarsonCMLK20, DBLP:conf/aclnmt/KhayrallahK18} and misdirection of model evaluation caused by the flawed test data \citep{DBLP:journals/corr/abs-2103-14749}. The main reason behind the phenomenon is that annotating a dataset is time-consuming, high-cost and labour-intensive, as is revising noisy and/or inconsistent datasets. Hence, an automatic re-annotator that can partially reduce labour costs or even fully replace human labour will be of benefit to the field of NLP.

The surge of Pre-trained Language Models (PLMs) is one of the most significant revolutions that has emerged in the era of neural NLP \citep{DBLP:journals/corr/abs-2003-08271, DBLP:journals/corr/abs-2111-01243}. Instead of domain-specific learning, PLMs are first unsupervised, pre-trained on the large-scale corpus, and then fine-tuned on downstream tasks. Recent studies suggest that the pre-training stage endows the PLMs with abundant commonsense \citep{DBLP:conf/emnlp/PetersNLSJSS19, DBLP:conf/emnlp/DavisonFR19, DBLP:conf/emnlp/JiangAADN20,DBLP:journals/tacl/JiangXAN20} and linguistic knowledge \citep{DBLP:conf/blackboxnlp/ClarkKLM19, DBLP:conf/blackboxnlp/ClarkKLM19,DBLP:conf/naacl/Liu0BPS19, DBLP:conf/acl/ChiHM20, DBLP:journals/tacl/Ettinger20}. The prior knowledge in PLMs has proved to be versatile in practice. For instance, PLMs can be directly used to evaluate text generation \citep{DBLP:conf/iclr/ZhangKWWA20, DBLP:conf/acl/SellamDP20} and probe factual knowledge \citep{DBLP:conf/emnlp/PetersNLSJSS19}. Considering the appealing property of PLMs, we are curious whether they can contribute to automatic re-annotation.

The prime motivation and novelty of this project involves applying Pre-trained Language Models as re-annotators to improve annotation quality with reduced cost and competitive results. As argued by \cite{DBLP:conf/eacl/DickinsonM03}, examples that have different labels but occur in very similar contexts are likely to be annotation inconsistencies or errors. Coincidentally, PLMs are well known for their outstanding capability of acquiring contextualized embedding. Therefore, the main idea of detecting or correcting diverging annotations with PLMs is to contrast the label of the target example with other examples embedded in its vicinity.

\section{Investigations}

We are the first to comprehensively study the potential of PLMs in data re-annotation using both a sentence-level and a document-level relation extraction dataset, TACRED \citep{zhang2017tacred} and DocRED \citep{yao-etal-2019-docred}. Relation extraction is the task of determining the relation holding between two entities in context -- one called the \textit{subject}, the other, the \textit{object}. We are evaluating re-annotators for sentence-level relation extraction using two datasets derived from TACRED --- TACRev \citep{DBLP:conf/acl/AltGH20a} and Re-TACRED \citep{DBLP:conf/aaai/StoicaPP21} with human revisions. To evaluate the re-annotators in document-level relation extraction, we re-annotated a subset of DocRED ourselves, to create a novel human revised dataset annotated with document-level relations.

\begin{figure}[h]
\includegraphics[width=\linewidth]{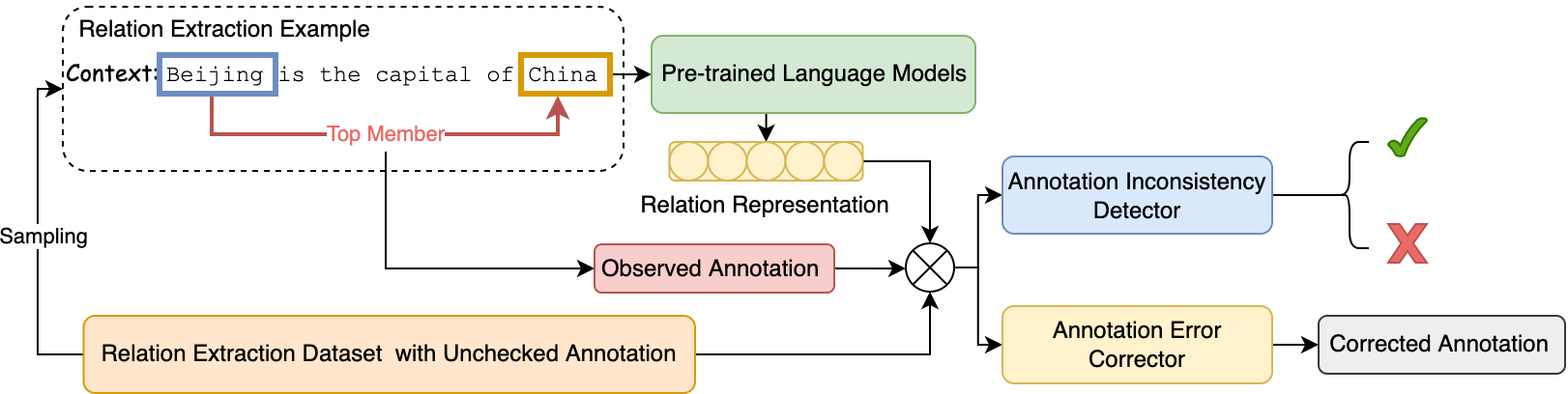}
\caption{The overview of our proposed Annotation Inconsistency Detector (AID) and Annotation Error Corrector (AEC) in the context of relation extraction. The \textbf{Annotation Inconsistency Detector} indicates whether the observed annotation is consistent with other annotations. The \textbf{Annotation Error Corrector} rectifies the annotations identified as mislabeled. }
\label{fig:intro}
\end{figure}

The re-annotation task is composed of two steps: \textbf{Annotation Inconsistency Detection} and \textbf{Annotation Error Correction} (Figure \ref{fig:intro}). Annotation Inconsistency Detection (AID) evaluates the consistency of each given annotation compared to other annotations in a similar context. Annotation Error Correction (AEC) suggests the proper annotation for the example identified as an anomaly.

Annotation Inconsistency Detection relies on the favourable property of representation methods and the desired sensitivity and specificity of the noise detector. Instead of fine-tuning the PLMs, we first researched different prompt \citep{DBLP:journals/corr/abs-2107-13586} and input modification \citep{DBLP:journals/corr/abs-2102-01373} techniques to acquire informative and distinguishable relation representation of each instance from PLMs. After optimizing the relation embedder, we leverage the K-Nearest Neighbour algorithm \citep{Mucherino2009} to determine potentially inconsistent annotations based on the local geometry of annotated examples in the embedding space. Furthermore, we propose a novel distance-based credibility score that jointly considers the labels in the vicinity and the global distribution of the assigned class of the query example. The experiments reveal that inadequately informative and overly predisposed prompts result in downgraded performance in detecting inconsistency. Empirically, our proposed credibility scores combined with relation prompts show promising agreement with human revisions in Re-TACRED and TACRev, reaching a binary $F_1$ score of 93.4 and 72.5 in detecting inconsistencies on TACRED and DocRED respectively. 

On the other hand, the PLM-based Annotation Error Corrector is fine-tuned by cross-validation \cite{stone1977asymptotic, tibshirani1996journal, doi:10.1080/00401706.1974.10489157} to make accurate revision decisions. The vanilla automatic corrector comprises PLMs stacked by a neural relation classifier. We ameliorate the learning process of AEC models with uncertain labels and neighbour-aware learning. Inspired by soft labels \citep{DBLP:conf/kes/Thiel08, DBLP:journals/jamia/NguyenVH14, DBLP:conf/emnlp/LiuWCS17, DBLP:journals/nn/ZhaoZCL14, DBLP:journals/corr/abs-2103-10869}, we construct the labels with uncertainty of samples based on their neighbours or estimated probability densities regarding each label class to replace overconfident hard labels. Specifically, one approach replaces part of hard labels with the majority labels among their neighbours, and another derives the soft-label vectors from the estimated kernel density corresponding to every class. We also explore two methods to augment the relation classifier with neighbouring knowledge: (1) rank-aware Transformer encoder \citep{DBLP:conf/nips/VaswaniSPUJGKP17} to acquire the relation embedding attended on their neighbours, and (2) distant-peer contrastive learning \citep{DBLP:conf/nips/KhoslaTWSTIMLK20} to include neighbour information to the loss function. Apart from sampling the positives and negatives in the batch, the framework of distant-peer contrastive learning selects positive examples from neighbours by our proposed peer distance computed by combining their co-occurrence frequency and squared Euclidean distance. Our empirical results show that the neighbour-aware annotation corrector trained with distant-peer contrastive learning obtains macro $F_1$ up to 66.2 on TACRED and 57.8 on DocRED. Moreover, training state-of-the-art relation extraction models on the training sets automatically denoised by our optimized annotation corrector leads to the maximum improvement of micro $F_1$ of 3.5\% on TACRED,  3.4\% on TACRev, and 1.1\% on DocRED.

Finally, we found that our proposed Annotation Error Corrector could automatically revise the annotations hundreds of times faster than human revisers with acceptable reliability. Similarly, an annotation inconsistency detector could spot dubious annotations even over ten thousand times faster than humans. Therefore, we believe that applying automatic re-annotators prior to manual revisions or even entirely relying on their revising outcomes would considerably improve the quality of data and data-driven NLP.

\section{Contributions}
The main contributions of the dissertation are:
\begin{itemize}
\item \textbf{RE-DocRED Dataset}: To study the annotation quality and evaluate our proposed automatic re-annotator in document-level tasks, we built a novel dataset, Re-DocRED, by revising 411 examples in the document-level relation extraction dataset, DocRED. It is the first re-annotated RE dataset at the document-level and will benefit future research on annotation noise.
\item \textbf{PLMs for Re-Annotation}: We are the first to leverage prior knowledge in Pre-trained Language Models to detect annotation inconsistencies and correct annotation errors. Our findings prove that factual and linguistic expertise in Pre-trained Language Models is applicable to automatic re-annotation even in the zero-shot scenario.
\item \textbf{Prompt Investigation}: We comprehensively study the impact of different forms of prompts and input modifications to the re-annotation tasks. Empirically, we found that prompts with either inadequate contexts or strong implications would mislead automatic re-annotators.
\item \textbf{Credibility Score}: We propose a novel credibility score jointly computed by the distance and reliability of neighbours. The reliability of neighbours is approximated by the estimated kernel density of their assigned classes. The experiment indicates the credibility score is very effective in spotting potential inconsistent annotations.
\item \textbf{Neighbour-based Uncertain Label}: We construct the labels with uncertainty based on the distribution of neighbours to avoid the annotation corrector from relying overly on the observed hard labels. Both the K-Nearest Neighbours based label replacements and Kernel Density Estimation based soft labels show convincing improvement to the corrector performance. 
\item \textbf{Distant-peer Contrastive Learning}: We develop a novel contrastive learning framework with augmented positive examples chosen by our newly defined peer distance. Based on their co-occurrence frequency and distance, we formulate the peer distance to select the most trustful and valuable positives from the neighbours of query examples. The results demonstrate the strength of combining distant-peer contrastive loss and the cross-entropy loss, especially when used in collaboration with with Kernel Density Estimation based soft labels.
\end{itemize}

\section{Dissertation Structure}
The summaries of the following chapters of the dissertation are listed as follows:
\begin{itemize}
\item \textbf{Chapter \ref{chap:rel_work}} introduces the previous work in (1) prior knowledge in Pre-trained Language Models and prompts for its transferability, (2) analysis of annotation noise and noise-tolerant learning methods, and (3) efforts in improving annotation quality manually and automatically.
\item \textbf{Chapter \ref{chap:data}} defines the two investigated tasks, Annotation Inconsistency Detection and Annotation Error Correction in the context of relation extraction, and describe the datasets for experiments.
\item \textbf{Chapter \ref{chap:AID}} presents our empirical study in relation representation methods and neighbouring consistency based binary classifiers for Annotation Inconsistency Detection.
\item \textbf{Chapter \ref{chap:AEC}} describes our investigation in cross-validation based Annotation Error Correction, and two improvements, uncertain labels and neighbouring awareness.
\item \textbf{Chapter \ref{chap:conclusion}} concludes our findings throughout the study and presents several charming directions to be explored in future.
\end{itemize}

\chapter{Related Work}
\label{chap:rel_work}

The three topics of most importance to this dissertation are: (1) Pre-trained Language Models; (2) identifying noise and inconsistencies that can arise in annotation, focusing on the annotation of relations rather than just on the annotation of simple strings; and (3) improving the annotation of relations through automating attempts to recognize and correct noisy or inconsistent tokens. We will address previous work on each of these topics in its own subsection, in order to clarify and justify the work we have done here.

\section{Pre-trained Language Models}
\label{sec:rel_plm}
Various Natural Language Processing tasks focus on an independent domain, but some tasks are intrinsically connected. For instance, while Dependency Parsing \citep{Dozat2017DeepBA, kubler2009dependency, nivre2005dependency, li2018seq2seq} generates syntactic dependency trees, and while Semantic Role Labeling (SRL) \citep{palmer2010semantic, marquez2008semantic} predicts the latent predicate-argument structure, both tasks require the models to have the fundamental capability of capturing the grammatical structure of the given context. Hence, the paradigm in NLP recently shifted from task-specific model designs to a pre-train and fine-tuned pipeline.

Typically, a sequence-to-sequence model with abundant parameters is trained on a massive corpus to perform language modelling tasks or text reconstruction tasks in an unsupervised learning fashion during the pre-training stage and then the model parameters are fine-tuned subtly with the task-specific data and learning objectives such as relation extraction and sentiment analysis. On the basis of its characteristics, those models are widely known as the Pre-trained Language Models (PLMs). According to the intended usage scenario, the PLMs can be roughly divided into general-purpose PLMs and special-purpose PLMs. The general-purpose PLMs are usually pre-trained on the general corpus, such as Wikipedia, and with the fundamental pre-training tasks and model architecture \citep{devlin-etal-2019-bert, joshi-etal-2020-spanbert, DBLP:conf/nips/BrownMRSKDNSSAA20, DBLP:journals/jmlr/RaffelSRLNMZLL20, DBLP:conf/acl/LewisLGGMLSZ20, DBLP:conf/nips/YangDYCSL19,DBLP:journals/corr/abs-1907-11692}. Those PLMs are more widely used and performed evenly on diverse downstream NLP tasks. Still, for those tasks demanding professional knowledge, they may be incapable of performing effectively because of lacking domain-specific pre-training. Therefore, the special-purpose PLMs normally pre-trained on a professional corpus \citep{DBLP:journals/bioinformatics/LeeYKKKSK20, DBLP:conf/emnlp/FengGTDFGS0LJZ20, DBLP:journals/corr/abs-2004-03760}, either refine the common model architecture of a general-purpose model \citep{DBLP:conf/emnlp/PetersNLSJSS19, DBLP:conf/acl/ZhangHLJSL19} or add auxiliary pre-training tasks \citep{DBLP:conf/acl/SoaresFLK19}.

Our target is to assist in mitigating the annotation noise and improve the quality of datasets. This dissertation is the first to comprehensively investigate the possible roles that PLMs can play in automatic re-annotation. To revise any existing annotations presented in the datasets, one needs to have two kinds of prior knowledge: (1) Factual and linguistic knowledge for revising the annotations that do not comply with common sense; (2) Knowledge of overall annotation distributions for detecting annotations that appear inconsistent with most other annotations. Recent studies reveal that most PLMs actually possess ample general factual and linguistic knowledge even without any fine-tuning or knowledge injection, such as \cite{DBLP:conf/emnlp/PetersNLSJSS19} and \cite{DBLP:conf/naacl/HewittM19}. Therefore, we focus on developing the framework for automatic re-annotation based on the general-purpose PLMs. The next section (Section \ref{sec:prior}) discusses the recent advances in probing the prior knowledge of general-purpose PLMs. Section \ref{sec:rel_prompt} then discusses the popular prompt-based learning for deriving the knowledge of interests from PLMs and applying it to the downstream tasks. Both directions motivate the methodologies we adopt to acquire the information-rich and highly differential representations of the examples in the dataset for automatic reexamination.

\subsection{Prior Knowledge in Pre-trained Language Models}
\label{sec:prior}
Just as bookworms can become encyclopedic by large amounts of reading, PLMs are also likely to acquire extensive linguistic and commonsense knowledge via unsupervised learning on the large-scale general corpus. Many empirical and analytical investigations have been conducted recently to justify this intuition for probing and quantifying the underlying prior knowledge in PLMs. Those findings doubtlessly breed sufficient confidence of trusting and leveraging prior knowledge in PLMs to revise the existing labels in flawed datasets.

Factual probing uncovers the hidden knowledge from PLMs firstly proposed by \cite{DBLP:conf/emnlp/PetersNLSJSS19}. To explore the solutions for this task, they introduce the LAMA (LAnguage Model Analysis) framework, which is intended to probe the commonsense knowledge in PLMs. According to the proposed framework, the entries in the knowledge bases are converted into cloze statements with templates where the relation or entity mentions are replaced with the mask. Then the prior knowledge in PLMs are assessed by their performance in completing the missing tokens. For instance, given the cloze statement "Newton was born in [MASK]", if the PLMs are able to rank the "UK" or "England" higher for filling the blank, they would be regarded as having more factual knowledge. Their experimental results convince that PLMs without fine-tuning still contain trustworthy relational knowledge and can handle open-domain questions based on their factual knowledge. \cite{DBLP:conf/emnlp/DavisonFR19} also develop a similar framework for mining the commonsense knowledge from PLMs. They first derive the masked sentences from the relational triples and then leverage the PLMs to rank the validity of a triple with the estimated point-wise mutual information between two entity mentions without fine-tuning. X-FACTOR by \cite{DBLP:conf/emnlp/JiangAADN20} attempt to generalize the cloze-style factual probing to a multi-lingual situation by composing the variations of cloze templates in 23 typologically divergent languages and proposing an improvement for probing multilingual knowledge from PLMs based on code-switching. Their experiments demonstrate the multilingual accessibility of factual knowledge in PLM. The work by \cite{DBLP:journals/tacl/JiangXAN20} further optimizes cloze-base querying processing for more accurate estimation of factual knowledge in PLMs by replacing the manually crafted prompts with an automatic pipeline that generates templates based on the paraphrasing and intention mining. They suggest that the quality or compatibility of the prompts could impact the performance of knowledge probers of PLMs. 

Aside from the commonsense knowledge, PLMs are to be able to comprehend a considerable amount of linguistic phenomena by merely pre-training on large-scale plain text without explicit linguistic annotations. \cite{DBLP:conf/naacl/HewittM19} testify syntactic knowledge in PLMs by showing that syntax trees are consistently embedded in the representation space of PLMs following certain liner transformations. \cite{DBLP:conf/blackboxnlp/ClarkKLM19} also confirm the existence of syntactic knowledge in PLMs through the empirical analysis of the attention distribution, which displays the pattern of directing objects of prepositions and verbs, determiners of nouns or co-referent mentions. \cite{DBLP:conf/acl/TenneyDP19} indicates that PLMs have the latent procedure of handling linguistic information similar to a conventional NLP pipeline of part-of-speech tagging, dependency parsing, named entity recognition, and then co-reference. Through sixteen various probing tasks, \cite{DBLP:conf/naacl/Liu0BPS19} investigate the detailed impacts of pre-training tasks on the linguistic knowledge learned by PLMs and prove that PLMs have the knowledge of semantic dependency and co-reference resolution. Based on the analysis of word embedding space by multilingual PLMs, \cite{DBLP:conf/acl/ChiHM20} discover the universal grammatical relations across languages captured by PLMs. \cite{DBLP:journals/tacl/Ettinger20} propose a novel suite of diagnostics derived from human language experiments to examine the linguistic knowledge in PLMs. They suggest that PLMs can robustly identify good from bad completions involving shared category or role reversal and retrieve noun hypernyms but are slightly hesitant compared to the human evaluators.

\subsection{Prompts for Knowledge Transferability}
\label{sec:rel_prompt}

Prompt-based learning, an straightforward method of applying the prior knowledge in PLMs to the downstream tasks, has drawn more attention recently because it supports zero-shot learning. Instead of time-consuming adaptation of all parameters in PLMs according to the downstream learning objective, prompt-based learning reformulates downstream tasks into language completion tasks with well-designed prompts. \cite{DBLP:journals/corr/abs-2107-13586} compose a comprehensive and systematic survey on recent developments of prompt-based techniques, and \cite{DBLP:conf/iclr/SaunshiMA21} formulate a solid mathematical framework to explain why prompt-based methods would work for downstream tasks. Based on the survey, we introduce prompt-based learning that forms the basis of our exploration of taking PLMs as automatic re-annotators for promoting the quality of datasets from two aspects: (1) prompt types and (2) their applications related to our task re-annotating the relation extraction datasets.

According to their form, prompts can be divided into two styles: cloze-style and prefix-style. Cloze-style prompts \citep{DBLP:conf/acl/CuiWLYZ21, DBLP:conf/emnlp/PetroniRRLBWM19} require PLMs to fill the intentionally masked span in a semi-completed sentence, and the prediction is made by the filling decisions or ranking made by PLMs. It is typically used to handle classification tasks which have clear restrictions or objectives, such as relation extraction or sentiment analysis. Conversely, prefix-style prompts \citep{DBLP:conf/acl/LiL20, DBLP:conf/emnlp/LesterAC21} encourage PLMs to generate a continuation of the given sentence. It is more suitable for those tasks involving the text generation, such as text summarization or question answering. Methods used to compose prompts can be categorized into manual template-based and automatic generated prompts. The former type of prompts is generated by manually crafted templates based on human understanding of the context of the task. For instance, as humans know that the relation mentions usually appear between the subject and object mentions in the text, inserting the mask token between the mentions of subject and object would be the most rational way to compose cloze-form prompts retrieving the possible relation held in the context. The latter type of prompts is automatically generated, searched or tuned with a small fraction of the downstream data. Considering the flexibility of expressions in natural language, generated prompts can more accurately derive task-related knowledge from PLMs to reinforce downstream performance. However, accompanying this strength, an automated template may make models relatively prone to overfitting compared to a manual crafted template because the strong implications in the prompt could easily mislead PLMs. This drawback is especially worth noting when we apply prompt-based methods for rechecking problematic annotations.

As we investigate the automatic re-annotator in the context of relation annotation, prompt applications in relation extraction could be very relevant. Relation extraction is the task of predicting the underlying relation between the given subjective and objective entities in the context. \cite{DBLP:journals/corr/abs-2104-07650} identify two major challenges of applying the prompt-based method to the relation extraction task: (1) the difficulties in prompt engineering caused by the enlarged label space of relations, and (2) the elusive importance of the tokens appeared in the context. To overcome these two obstacles, they developed the KnowPrompt framework that constructs the learnable prompt for relation extraction with virtual template words and answers words. They further inject the knowledge of entity and relation via marking entity spans by wrapping the entity mentions with special markers such as \texttt{[E]}. Correspondingly, \cite{DBLP:journals/corr/abs-2105-11259} leverage the technique of prompt composition to form the prompt with abundant entity type information. The prompt composition technique is to compose the final prompt by synthesizing several sub-prompts based on logic rules. For instance, to extract the relation between "Google" and "Alphabet" with the context "Google became a subsidiary of Alphabet", we can compose the complete prompt as "The [MASK] Google [MASK] the [MASK] Alphabet" from the sub-prompts "The [MASK] Google", "The [MASK] Alphabet", " Google [MASK] Alphabet". The first two sub-prompts enable the PLMs to associate the supplementary knowledge about the entity "Google" and "Alphabet" independently before guessing their relation. 

\section{Annotation Noise}
Annotation noise in the form of inconsistent or incorrect labels appears to be common in most machine learning datasets. \cite{DBLP:journals/glottometrics/Kusendova05} reckon that the noise can follow from unclear or insufficient instructions, unclear contexts, bias on the part of the annotator, or deviations from what the instruction writers expect. Nowadays, training neural network models usually requires vast amounts of annotated data, so this issue has become increasingly prominent in machine learning. The annotators of these datasets often lack expertise in the related domain because the most commonly used method for composing large-scaled supervised data involves crowdsourcing the annotation with the platforms like Amazon Mechanical Turk \citep{DBLP:conf/ifip8-2/Crowston12}, and the annotation is done by cheap labour rather than well-paid expert annotators.  Specifically, the platforms distribute a small fraction of the whole annotation task with brief instructions to the non-expert crowd workers and then merge the partial annotations together to form the massive dataset. Since the crowd workers may have their own annotation standards, the quality of crowdsourced datasets is worrisome due to potential inconsistencies and errors and endangers the robustness of machine learning systems. Considering these facts, we believe an automatic pipeline for denoising the annotations in datasets is a worthwhile and meaningful way to resolve the bottleneck in machine learning. Furthermore, since PLMs are pre-trained on unlabeled data in an unsupervised learning fashion without any human intervention, applying PLMs to improve the data quality is an appealing method for reducing the uncontrollable human factor in machine learning.

In Section \ref{sec:analysis_noise}, we introduce the annotation noise in the context of classification tasks from four perspectives: definition, taxonomy, sources and its downstream influence, which specify the context of our investigation of automatic re-annotation. In Section \ref{sec:noise_tolerant_learning}, we describe plausible methods to learn on imperfect datasets without alternating the annotation noise. Compared to noise-tolerant learning, we believe that revising error annotations with PLMs brings more interpretability and insights for handling the annotation noise.

\subsection{Analysis of Label Noise}
\label{sec:analysis_noise}
Based on the previous work on the label noise \citep{DBLP:conf/ranlp/SharouLS21, DBLP:conf/coling/LarsonCMLK20, DBLP:journals/tnn/FrenayV14, beck2020representation}, we briefly demonstrate the label noise involved in classification tasks from the following four facets: 
\paragraph*{Definition of Annotation Noise} For supervised multi-class classification, each sample corresponds to a true class, but the identification of this class would be passed into a noise process to become the observed labels presented to the classification models. Observed annotations may be different from the true labels of samples. In this process, the compromised annotations are called \textit{label noise} \citep{DBLP:journals/ml/AngluinL87}, in contrast to feature noise, which is the perturbation of feature values. However, as noisy labels often have a relatively lower probability of occurrence in their vicinity than the normal labels, the mislabelled examples may be defined as the anomalies or outliers of the distribution of their assigned class in some cases. Therefore, anomaly detection \citep{DBLP:conf/nips/ScholkopfWSSP99, DBLP:conf/nips/HaytonSTA00, DBLP:journals/neco/ScholkopfPSSW01, DBLP:journals/pr/Hoffmann07, DBLP:journals/csur/ChandolaBK09} and outlier detection \citep{barnett1978study, sebert1997outliers, DBLP:conf/emnlp/Zhou0C21, DBLP:journals/air/HodgeA04, DBLP:conf/aici/NiuSSH11, DBLP:books/sp/Hawkins80, DBLP:journals/corr/abs-2007-05566} are of great relevance to our task. For instance, \cite{DBLP:journals/corr/abs-2007-05566} motivate us to combine the cross-entropy loss with a contrastive loss to enhance the performance of the automatic re-annotator, and we further deliberate the sampling strategy of contrastive learning via a novel distant-based criterion. The framework developed by \cite{DBLP:conf/emnlp/Zhou0C21} is also based on contrastive learning and PLMs, detecting the out-of-distribution in relation extraction by the Mahalanobis distance \citep{mclachlan1999mahalanobis} of the hidden representations of examples in the penultimate layer. However, our task, re-annotating problematic annotations, is significantly distinguished from the out-of-distribution detection, which intends to solve the problems caused by the different distributions of training and real-world test data.

\paragraph*{Taxonomy of Annotation Noise}
Although the taxonomy of annotation noise is potentially large, we only define two types of annotation noise here to reduce the scope of the investigation: annotation inconsistency and annotation error. Most of the time, the expression of inconsistency and error can be interchangeable, but we would give narrow definitions for them in the context of our project. Annotation inconsistency is the particular annotation that is divergent from the annotations shared by examples with similar context \citep{DBLP:conf/coling/LarsonCMLK20, DBLP:conf/lrec/HollensteinSW16, DBLP:conf/sigdial/QianBLDGYS21, DBLP:conf/icdm/LiQWYS20}, but the inconsistent annotation is not necessarily incorrect. For instance, the relation between "James" and "Bob" in the context "James has a son called Bob", could be annotated as \texttt{father} or \texttt{family\_member\_of} rationally. However, if in most similar cases, we choose to annotate the example with the most precise plausible relation, then \texttt{family\_member\_of} label here may be taken as inconsistent. In contrast, annotation error is a broad concept, referring to annotations that are counter to common sense or that contradict the given context \cite{DBLP:conf/conll/ReissXCME20, DBLP:journals/ieicet/SuzukiKM17, DBLP:journals/csl/MatousekT17, DBLP:conf/depling/HaverinenGLKVNS11, DBLP:phd/ethos/Bryant19}. Therefore, in the former example, neither label  (\texttt{father} or \texttt{family\_member\_of}) is an annotation error.

\paragraph*{Sources of Annotation Noise}
Typical sources of annotation noise are: (1) The annotators do not have sufficient information or knowledge to successfully complete the annotation task \cite{DBLP:journals/ai/Hickey96, DBLP:journals/jair/BrodleyF99, DBLP:conf/cbms/PechenizkiyTPP06}, which is not uncommon when, for example, medical or legal text data is annotated. (2) The inconsistency and errors are introduced in a crowd-sourcing annotation scenario, when a large number of non-experts are involved in the annotation process \citep{DBLP:conf/coling/LarsonCMLK20, DBLP:conf/emnlp/SnowOJN08, DBLP:journals/jmlr/RaykarYZVFBM10, DBLP:conf/socialcom/YuenKL11}. (3) The target of the annotation task is ambiguous or subjective, such as  annotation for image classification or medical analysis \citep{DBLP:conf/acl-louhi/GrivasAGTW20, DBLP:journals/bioinformatics/MalossiniBN06, DBLP:journals/prl/Smyth96, DBLP:conf/naacl/FornaciariUPPHP21}. (4) There are distractions during  annotation or problems in the design of the annotation interface, such as lack of feedback mechanism \citep{DBLP:conf/ceas/SculleyC08}. Intuitively, automatic re-annotation is expected to mitigate the second and fourth sources of annotation noise effectively. The first source may be solvable by  PLMs pre-trained on domain-specific data \citep{DBLP:journals/bioinformatics/LeeYKKKSK20, DBLP:conf/emnlp/FengGTDFGS0LJZ20, DBLP:journals/corr/abs-2004-03760}, but we leave this for future exploration.

\paragraph*{Influence on Model Performance}
The negative impact of annotation noise can be considerable. Experiments conducted by \cite{DBLP:conf/coling/LarsonCMLK20} show that any type of inconsistency in crowd-sourced data would downgrade model performance in slot-filling, though different types of inconsistency may have a different level of impacts. \cite{DBLP:conf/aclnmt/KhayrallahK18} empirically study the impact of diverse annotation noise in a parallel corpus for machine translation and reveal that neural machine translation models are more error-prone to annotation noise than statistical machine translation methods. \cite{DBLP:conf/icml/ChenLCZ19} suggest that the accuracy on the Test set could be used as the quadratic function to evaluate the noise ratio in the dataset if the annotation noise can be categorized into the symmetric noise \citep{DBLP:conf/nips/RooyenMW15}. \cite{DBLP:conf/acl/AltGH20a} and \cite{DBLP:journals/corr/abs-2103-14749} prove that annotation noise in the Test set significantly misleads the process of model evaluation and selection. Additionally, \cite{DBLP:journals/corr/abs-2103-14749} argue that less powerful models with fewer parameters or simpler model architecture have more resistance and regularization to robustly learn on the data with asymmetric distribution of noise than large models with more advanced representation learning capability. \cite{DBLP:journals/corr/abs-2001-01987} present a mathematical explanation for the deterioration of softmax classifiers caused by annotation noise by reformulating a softmax classifier as K-means clustering and deducing the relation between prediction distortion and annotation noise based on Lipschitz Continuity \cite{sohrab2003basic}. 

\subsection{Noise-tolerant Learning}
\label{sec:noise_tolerant_learning}
Instead of improving the annotation quality like automatic re-annotators, noise-tolerant learning intends to minimize the negative impact of the annotation noise during the training period, typically by learning to treat the labels with different degrees of credibility. The most commonly used techniques for mitigating the perturbation of noisy data are soft-labeling \citep{DBLP:conf/kes/Thiel08} and curriculum learning \cite{DBLP:journals/corr/abs-2101-10382, DBLP:conf/ijcai/PortelasCWHO20, DBLP:journals/corr/abs-2010-13166, DBLP:conf/icml/BengioLCW09}. Though noise-tolerant learning empirically leads to improved robustness for handling annotation noise, our work improves its fundamental component of suspicious annotation detection with an enhanced PLM-based detector and the interpretability of its black-box learning process with annotation correction.

The core idea of applying soft-label to relieve the noise in training data is to avoid heavily and blindly relying on the observed labels. For example, \cite{DBLP:conf/kes/Thiel08} show that the benefit of soft-label in terms of improving noise resiliency of the classifier compared to the hard labels. \cite{DBLP:conf/emnlp/LiuWCS17} construct the soft-label of examples in the noisy distant-supervised relation extraction dataset by jointly considering both the credibility of observed labels and entity-pair correlation in the context, which results in excellent improvement of the model performance. Similarly, \cite{DBLP:journals/corr/abs-2103-10869} derive the soft-label from the features of examples in training data and gradually update the soft-label with meta-objective at the beginning of each training iteration, where the meta-objective is obtained by cleaning a small fraction of training data. 

The main intuition behind curriculum learning for noise-resistant learning is to let the model learn on the trustworthy data with a large learning rate first and then subtly adjust the parameters based on possibly noisy data. MentorNet proposed by \cite{DBLP:conf/icml/JiangZLLF18} is a vivid example to exemplify this idea. The noise-resistant learning procedure involves two paired neural networks, MentorNet and StudentNet. MentorNet learns the curriculum that can help StudentNet focus on the training examples with a relatively high probability of being correctly labelled. The curriculum taught by MentorNet is initially learnt from a tiny dataset with checked labels and then iteratively deliberated based on the learning feedback of StudentNet. \cite{DBLP:journals/jair/NorthcuttJC21}  further take MentorNet as the baseline for exploring confident learning in the general domain. Similarly, \cite{DBLP:conf/iclr/ZhouWB21} leverage the curriculum learning based on the data parameters to produce noise-resilient keyword spotting models. To enhance the learning outcomes, \cite{DBLP:conf/icassp/HiguchiSSTDD21} utilize the time ensemble of the model and data augmentations to generate pseudo labels for composing noisy data as the harder curriculum for models to learn to handle annotation noise.

\section{Improving Annotation Quality}
Aside from noise-tolerant learning, we could also directly improve or examine the quality of annotation. According to the degree of automation, we categorize the methods for improving annotation quality into three classes: (1) Annotation Process Improvement, that is to manually optimize the factor and steps in the process of annotation, (2) Annotation Inconsistency Detection, that is to spot suspected labels or indicate the reliability of labels based for downstream manual or automatic correction, and (3) Annotation Error Correction, that is to automatically correct the mislabeled examples.

In Chapter \ref{chap:AID}, we introduce a novel PLM-based approach for Annotation Inconsistency Detection, and in Chapter \ref{chap:AEC} we further develop contrastive methods for Annotation Error Correction based on cross-validation. Our proposed models have two significant properties that enable them to stand them apart from previous work in Annotation Inconsistency Detection and label rectification: (1) We are the first to leverage the distribution of contextualized embedding from PLMs to indicate the annotation inconsistency and correct the label errors. (2) Our proposed methods do not rely on any explicit linguistic knowledge, pre-defined rules, or cleaned data, which enable them to be generalized to different domains.

\subsection{Improving the Annotation Process}
Improving the process of collecting the annotations is the most straightforward way to alleviate annotation noise, but it usually requires more human labour and increases annotation cost. To reduce the noise caused by non-expert annotators, we could follow previous work in choosing the candidates of annotator who pass a qualification test, systematically training annotators, or cyclical annotation \citep{DBLP:conf/acl/RoitKSMMSZD20, DBLP:conf/hcomp/LiL15, DBLP:conf/acllaw/AlexGSK10}. To alleviate the noise raised by internal inconsistency between multiple annotators, we could overlap a small fraction of their annotation work to measure their agreement or repeatedly collect the annotations of each example from multiple annotators and then aggregate their decisions comprehensively \citep{DBLP:conf/naacl/HovyBVH13, DBLP:journals/tacl/PassonneauC14, DBLP:conf/emnlp/PardeN17, DBLP:conf/mir/NowakR10, DBLP:conf/emnlp/JamisonG15}. As for the noise introduced by intricate annotation objectives, we may reformulate the annotation targets or equip the annotation platform with an annotation assistant based on active learning \citep{DBLP:journals/fdgth/DobbieSPFJATL21, DBLP:conf/eacl/NghiemBA21, DBLP:journals/corr/abs-2112-11914}. The noise due to the distraction during the annotation period also could be improved by monitoring if the annotators could correctly label the probe examples with known golden annotations \citep{oppenheimer2009instructional}.
\subsection{Detecting Annotation Inconsistency}
There have been extensive investigations in Annotation Inconsistency Detection for a better understanding of the distribution of annotation noise. Early work in this area focussed on detecting inconsistent part-of-speech tags \citep{DBLP:conf/emnlp/AbneySS99, DBLP:conf/anlp/Eskin00, DBLP:conf/lrec/MatsumotoY00, DBLP:conf/coling/NakagawaM02, DBLP:conf/lrec/MatsumotoY00, DBLP:conf/icann/MaLMII01}. Since then, the statistical and rule-based methods have been effective in correcting the corpus for other syntactic prediction tasks, such as semantic role labeling \citep{DBLP:conf/lrec/DickinsonL08}, and dependency parsing \citep{DBLP:conf/acl/Dickinson10, DBLP:conf/iwpt/DickinsonS11}. However, the relation extraction task that we investigated as the semantic task is comparatively complicated to detect the inconsistency in its context because of the flexibility of semantic expression. For instance, detecting inconsistency in multi-word-expression or word sense datasets needs more powerful models. \cite{DBLP:conf/acllaw/DligachP11} screen out the annotations that are worth rechecking based on the in word sense data. They regard the examples that are repeatedly predicted as suspicious by both the machine tagger based on support vector machine and the ambiguity detector based on trainable probabilistic classifier as the inconsistent annotations. \cite{DBLP:conf/lrec/HollensteinSW16} propose an algorithm based on the ranking of absolute frequency and entropy of the label distribution to automatically spot inconsistencies in multiword expression and supersense tagging datasets. \cite{DBLP:conf/sigdial/QianBLDGYS21} develop an automatic inconsistency detector for the task-oriented dialogue dataset, MultiWOZ, based on scripts using regular expressions. Our work is, to the best of our knowledge, the first attempt to detect annotation inconsistencies in the relation extraction domain.
\subsection{Correcting Annotation Error}
Annotation Error Correction is a further step from Annotation Inconsistency Detection, demanding the models to not only identify the suspect labels but also to suggest a rational modification at the same time \citep{DBLP:conf/cikm/ZhangSWFW15, DBLP:conf/dsaa/NicholsonZSW15, DBLP:journals/ijon/BhadraH15}. As one of the recent advances in label correction, \cite{DBLP:conf/aaai/WuSX0M21} develop a meta-learning framework that first approximates the soft label of each example in datasets under the guidance of the small meta dataset with cleaned labels, and then derive the meta learner for annotation correction from the meta-process of soft label estimation. Similarly, \cite{DBLP:conf/aaai/ZhengAD21} also carry out the annotation correction process on a small set of data with checked labels as the meta-process and develop the meta-learning based framework composed of two network models, where the meta-model is for correcting the noisy annotations, and the main model is for exploiting the rectified label. The two network models in this framework are collaboratively trained as a bi-level optimization problem. Conversely, \cite{DBLP:conf/sigir/ZouZHS21} propose an unsupervised ensemble framework for annotation correction without the requirement of checked golden data. They first aggregate the annotations for holistic examples in the dataset by expectation-maximization algorithm, then screen out the hard case to form the targeting dataset with a two-step filtering approach, and finally apply an Adaboost classifier trained on the low-risk remaining dataset to predict the label corrections on the target dataset. In conclusion, aside from being the first to explore the Annotation Error Correction in relation extraction, we also present an appealing research direction of replacing the meta-learning process of label correction with the powerful prior knowledge in PLMs. In the future, we will also explore whether screening out the most suspicious annotations first can aid our proposed automatic annotation rectification.

\chapter{Tasks and Data}
\label{chap:data}

Annotation Inconsistency Detection and Annotation Error Correction are the two most vital tasks of automatic re-annotation. To clarify the scope of this dissertation, we give both conceptual and mathematical definitions of these two tasks. We also discuss how Pre-trained Language Models can set the foundation for successful re-annotation. With clear objectives, we describe the details of two kinds of datasets we used for conducting the experiments: target datasets for learning and revised datasets for evaluation.   

\section{Re-Annotation in Relation Extraction}
Relation Extraction (RE) is the task to predict the semantic relations between given subjective and objective entities, namely head and tail entities, in the context \cite{DBLP:conf/icsim/WangLYQ21, DBLP:journals/corr/abs-2007-04247, DBLP:conf/ccks/CuiLWY17}. A typical RE example can be to discern the relation between the head entity "SpaceX" and the tail entity "Elon Musk", given the sentence "SpaceX was founded in 2002 by Elon Musk". Ordinarily, RE datasets contain a textual context and spans of head and tail entities in the context and have pre-defined types of plausible relations. Some datasets also provide the auxiliary information of entities, such as Named Entity Recognition (NER) types. Hence, annotation in RE is the process of chosing relations between head and tail entities in the context from a collection of relation types. Relation Extraction is an indispensable component for composing the knowledge graphs \cite{DBLP:conf/apweb/LiWWZ019} that are useful to various downstream NLP applications such as question answering \citep{DBLP:phd/dnb/Dubey21, DBLP:conf/emnlp/SaffariOSA21, DBLP:conf/emnlp/SenOS21} and dialogue system \citep{DBLP:journals/ijon/LiuTLC21, DBLP:conf/esws/ChaudhuriR021, DBLP:conf/semweb/GaoZZFL21}. 

The re-annotation task presupposes that any annotation observed in datasets is not necessarily the true labels of the example in reality because of the annotation noise (Section \ref{sec:analysis_noise}). Therefore, re-annotation in RE is to re-examine the annotated relations, suggest their credibility, and recommend better relations if possible.    

To formally explain the task, we also present the mathematical description of RE re-annotation. Let $\mathcal{R}$ denote the set of relation types, $\mathcal{A}$ denote all annotations, $sub$ and $obj$ denote the subjective and objective entities (head and tail entities) respectively, and $c$ denotes the context of each example. $\mathcal{R}$ contains finite $n$ types of relation $t$, namely $\mathcal{R} = \{t_i\}^n$. Given context $c_i$, there is a true valid relation $ r_i(sub_i, obj_i) \in \mathcal{R}$ holding between entities $sub_i$ and $obj_i$. However, due to the possible annotation noise, we can only see $ r'_i(sub_i, obj_i) \in \mathcal{R}$  which is the observed relation in datasets. Each example in the datasets with its annotation can be denoted as $a_i(r'_i(sub_i, obj_i), c_i ) \in \mathcal{A}$. Thus, re-annotation in RE is to testify whether $r = r'$or reveal the true relation $r$, based on the observed $r'$, overall annotations $\mathcal{A}$ and implicit real-world knowledge. The corresponding tasks are: Annotation Inconsistency Detection and Annotation Error Correction.

\subsection{Definition of Annotation Inconsistency Detection}
\label{sec:aid_def}
Annotation Inconsistency Detection (AID) is the task to verify if the annotation of each example is consistent with other annotations. AID on Relation Extraction datasets aims to detect whether the observed relation of each example is consistent with other examples with similar entities and similar contexts.

AID could also be conducive for downstream model training in various aspects. It helps a human re-annotator start by looking at the most error-prone annotations to save time. The models may benefit from differentiating the reliable data and unreliable data during training in the curriculum learning fashion \citep{DBLP:journals/corr/abs-2101-10382, DBLP:journals/corr/abs-2010-13166}, or by adjusting the weights of training data \citep{DBLP:conf/emnlp/WangSLLLH19}. 

In this project, we define the AID as the binary classification task with the target $y \in [0, 1]$ where $1$ means the annotation is consistent and $0$ means the annotation is inconsistent. For instance, there is a RE example with the context $c$ "Alan Turing died in 1954" and the observed relation $r'(\texttt{Alan Turing},\texttt{1954})$ as \texttt{date of birth}. AID models are trained to predict $y =0$  to indicate it as inconsistent, if we have sufficient reasons to believe  $r' \neq r$ based on the context $c$, observed relation $r'(\texttt{Alan Turing},\texttt{1954})$ and the global distribution of all annotations.

\subsection{Definition of Annotation Error Correction}
\label{sec:aec_def}
Annotation Error Correction (AEC) task is considered to be more complicated than AID task. It not only testifies the annotations but also attempts to predict the true labels for annotations identified as incorrect simultaneously. As for Relation Extraction datasets, AEC models aim to validate the annotated relations between entities in the context and rectify the invalid relations. 

Unquestionably, the latest models with millions of parameters could easily capture the representations from the data on an unprecedented level. However, a growing body of research revealed that the data quality is a non-negligible factor that significantly downgrades the model performance or misleads the evaluation \citep{DBLP:conf/coling/LarsonCMLK20, DBLP:conf/aclnmt/KhayrallahK18, DBLP:journals/corr/abs-2103-14749}. Hence, AEC is expected to become an appealing alternative for manually enhancing data quality. It is common to take human annotators hundreds of hours to revise the large-scale datasets, while AEC systems would vastly shorten the revising time and reduce the cost. 

In this dissertation, we define AEC as the multi-class classification task with the target classes that is the same as the set of pre-defined relations $\mathcal{R}$. For example, if an observed relation $r'(\texttt{Alan Turing},\texttt{1954})$ in the context $c$ of "Alan Turing died in 1954" is \texttt{date of birth}, AEC models are expected to predict the true relation $r = \texttt{date of death}$, where $r \in \mathcal{R}$.

\subsection{Basis of Automatic Re-Annotation}
\label{sec:basis}
\begin{figure}[h]
\includegraphics[width=\linewidth]{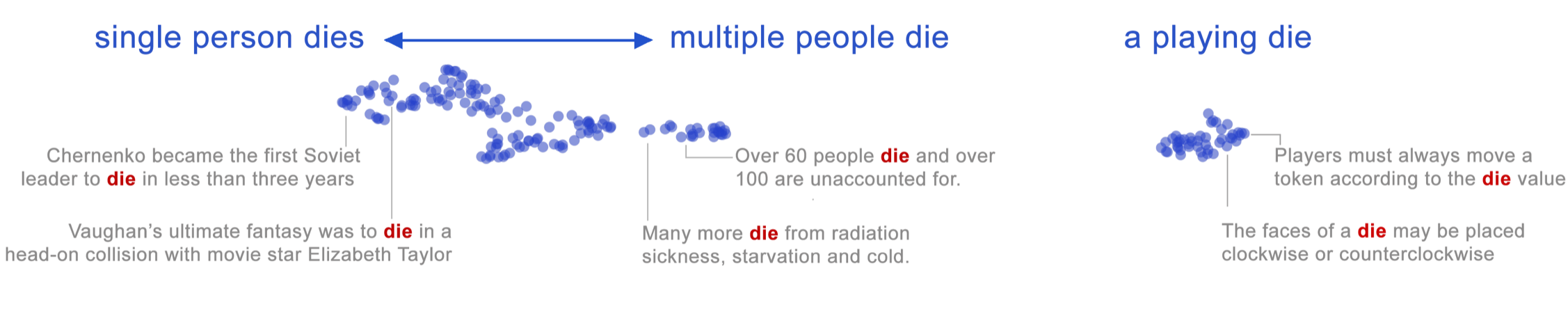}
\caption{Word embeddings from Pre-trained Language Models can capture subtle semantic variations in the context. Hence, Pre-trained Language Models raise novel possibilities in automatic re-annotation. Figure from \cite{DBLP:conf/nips/ReifYWVCPK19}.}
\label{fig:plm_basis}
\end{figure}
The essential idea of automatic re-annotation is to contrast the query annotation with other annotations in a similar context through models. As shown in Figure \ref{fig:plm_basis}, Pre-trained Language Models (PLMs) demonstrate extraordinary sensitivity to the subtle variations in context. Therefore, we believe it is possible to acquire sufficiently distinguishable relation representations for automatic re-annotation based on the embeddings by PLMs.

Mathematically, \cite{DBLP:conf/iclr/SaunshiMA21} uncover the key factors for successfully applying PLMs to the task of interests by reformulating the PLM-based classification tasks into sentence completion tasks. They give a criterion, Definition 3.2 in their paper, for judging whether a downstream classification task $\mathcal{T}$ can be considered as a natural task, namely an analogous sentence completion task, with regards to PLM-based embeddings $\Phi \in \mathbb{R}^{d \times V}$, where $V$ is the vocabulary size. If we assume $p \in \Delta \mathcal{C}$ to indicate probability distribution over context $\mathcal{C}$, and $p_{\cdot|c}$ to denote the true conditional distribution over words in vocabulary $\mathcal{W}$ in given context $c$, the criterion can be formulated as the inequality:
\begin{equation}
\min_{\mathbf{v}\in \text{row-span}(\Phi) ,\|\mathbf{v}\|_{\infty} \leq B} \hspace{2mm}\mathrm{l}_{\mathcal{T}}(\{\mathbf{p}_{\cdot|c}\}, \mathbf{v}) \leq \tau 
\end{equation}, where $\mathrm{l}_{\mathcal{T}}$ is the 1-Lipschitz surrogate \citep{DBLP:conf/icml/Mairal13} to the classification loss of task $\mathcal{T}$, $\{\mathbf{p}_{\cdot|c}\}$ denotes a language model, and $\tau$ and $B$ are two constrains. Through the detailed proofs, they give an intuitive interpretation of $\tau$ and $B$: (1) $\tau$ indicates the ambiguity of downstream task measured by Bayes error \citep{franklin2005elements}, and (2) $B$ is inversely proportional to the probability mass of the set of indicative words. Thus, the less ambiguous downstream task which mainly involves frequent words will have smaller $\tau$ and $B$. Qualitatively, \cite{DBLP:conf/iclr/SaunshiMA21} suggest two positive factors that help PLM to solve downstream tasks: (1) the PLM-based embeddings can capture the needed semantic meaning in the context and (2) tasks of interest is solvable by distinguishing words with obviously different meanings.

Based on the enlightenment, we develop the frameworks for AID and AEC tasks described in Chapter \ref{chap:AID} and Chapter \ref{chap:AEC} respectively.
 
\section{Datasets in Relation Extraction}
\label{sec:data}
To comprehensively study the automatic re-annotators, we need two types of datasets: (1) target datasets with unchecked annotations and (2) their corresponding revised datasets with cleaned annotations. 

Target datasets is the real-world data with observed annotations that are not necessarily correct or consistent. Therefore, the first usage of target datasets is to ask the re-annotators to learn on them, and then assess if re-annotators can detect inconsistencies or correct errors automatically. In this case, we combine all data splits with unchecked annotation, namely original Train, Dev and Test sets together as the new Train set for our proposed re-annotator. The second usage of target datasets is for downstream evaluation, namely to testify whether the state-of-the-art (SOTA) RE models can benefit from the data denoised by our proposed annotation corrector. In this case, we follow the original data splits to conduct the downstream RE experiments. More details about the downstream evaluation will be introduced in Section \ref{sec:downstream}.

On the other hand, we assume the revised versions of target datasets as the golden standard of revision. If we consider the rechecked annotations in revised datasets as ground truths, we can evaluate our proposed re-annotators by comparing their predictions with the manual revisions  in revised datasets with classification metrics. Therefore, we derive the Dev and Test sets with checked annotations for our proposed re-annotator from the revised datasets.

Since we decide to investigate the re-annotation in RE, we choose the popular sentence-level RE dataset TACRED \citep{zhang2017tacred} and document-level RE datasets DocRED \citep{yao-etal-2019-docred}, as two target datasets. The sentence-level dataset TACRED already has two revised versions called TACRev \citep{DBLP:conf/acl/AltGH20a}  and Re-TACRED \citep{DBLP:conf/aaai/StoicaPP21}. Nevertheless, to the best of our knowledge, there is no existing manual revision of document-level datasets DocRED. Therefore, to study the new challenges posed by the context length and inter-sentence complexity, we manually re-annotated a subset of DocRED ourselves, called Re-DocRED.

\subsection{Target Datasets}
\subsubsection*{TACRED}
TACRED \footnote{\url{https://nlp.stanford.edu/projects/tacred/}} \citep{zhang2017tacred}, The TAC Relation Extraction Dataset, is one of the largest and most widely used datasets for sentence-level RE task, containing examples from web text to news articles from TAC Knowledge Base Population (TACKBP) challenges. Examples in TACRED cover 41 positive relation types which is identical to TACKBP challenges (e.g., \texttt{per:title\_of}) and one negative type (\texttt{no\_relation})  if no defined relation is held between given head and tail entities. These examples are created by combining available human annotations from the TACKBP challenges and crowd-sourcing. However, as shown by \citet{DBLP:conf/acl/AltGH20a} and \citet{DBLP:conf/aaai/StoicaPP21}, TACRED is a typical dataset with pervasive annotation errors and inconsistencies. Since TACRED originally have 68,124 Train samples, 22,631 Dev samples and 15,509 Test samples, there are in total 103, 738 examples for training the sentence-level re-annotators.

\subsubsection*{DocRED}
DocRED \footnote{\url{https://github.com/thunlp/DocRED}} \citep{yao-etal-2019-docred}, Document-Level Relation Extraction Dataset, is a document-level RE dataset derived from Wikipedia and Wikidata. DocRED requires reading multiple sentences in a document to extract entities and infer their mutual relations by common-sense reasoning and aggregating contextual information of the document. Compared to the TACRED dataset, the examples in DocRED datasets generally exhibit more complex inter-sentence relations. They can be more intricate than the existing relation extraction (RE) methods that focus on extracting intra-sentence relations for single entity pairs. Moreover, DocRED has 96 valid relation types which is also siginificantly more than TACRED. 

DocRED contains 132,375 entities and 56,354 relational facts on 5,053 human-annotated Wikipedia documents. We are the first to examine the annotation quality and spot the annotation deficiency on DocRED. As shown in Table \ref{tab:datarev}, we eliminate some examples with multiple relations to reduce the ambiguity while training, and there are 50,503 examples in the Train set for document-level re-annotators.

\begin{table}[h]
\centering
\begin{tabular}{@{}llllll@{}}
\toprule
Dataset                    & Split & \#Example & \#Positive & \#Negative & \#Relation \\ \midrule
TACRED                     &  Train     & 103,738    & 19,247      & 84,491      & 42         \\ \midrule
TACRev                     & Dev   & 1,263      & 596        & 667        & 40         \\
                           & Test  & 1,263      & 628        & 635        & 39         \\ \midrule
\multirow{2}{*}{Re-TACRED} & Dev   & 5,364      & 3,596       & 1,768       & 36         \\
                           & Test  & 5,365      & 3,608       & 1,757       & 36         \\ \midrule \midrule
DocRED                     &   Train    & 50,503     & 50,503      &      0      & 96         \\ \midrule
\multirow{2}{*}{Re-DocRED} & Dev   & 206       & 206        &       0    & 55         \\
                           & Test  & 205       & 205        &    0       & 55         \\ \bottomrule
\end{tabular}
\caption{The statistics of target datasets TACRED and DocRED and their revised datasets TACRev, Re-TACRED, and Re-DocRED datasets. The example is negative if no defined relation is held between head and tail entities.}
\label{tab:datarev}
\end{table}

\subsection{Existing Revised Datasets}

\subsubsection*{TACRev}
TACRev \citep{DBLP:conf/acl/AltGH20a} was the first investigation on the annotation noise in the TACRED dataset. Linguists re-examined the most challenging 5,000 examples in TACRED, and 2,526 of them were revised from the original labels, of which roughly 57\% of the negative labels were modified into the positive labels. They proved that the performance ceiling of previous SOTA models on TACRED datasets is largely due to the annotation noise, showing that 4 SOTA models can improve 8\% absolute $F_1$ test score by evaluating on refined TACRev dataset. 

The TACRev dataset only releases the 2,526 revised examples without the annotations proved to be correct.
We first shuffle the 2,526 revised examples and then evenly split them into Dev set and Test set (Table \ref{tab:datarev}). The Dev set contains 40 relations including \texttt{no\_relation}, while Test set contains 39 relations. 

\subsubsection*{Re-TACRED}
Compared to TACRev, Re-TACRED \citep{DBLP:conf/aaai/StoicaPP21} is the crowd-sourced version of TACRED revision instead of being relabeled by several linguists. The entire TACRED dataset was rechecked by an improved and cost-efficient crowd-souring annotation strategy with a quality control mechanism. During the process of relation definition refinements intended to resolve the ambiguous relation definition in TACKBP, the authors further introduced new relations (e.g. \texttt{org:-member} as the inverse relation of \texttt{org:members} and \texttt{org:subsidiaries}) and rename several initial relation (e.g. \texttt{per:alternate\_name} to \texttt{per:identity}). Since the automatic re-annotators are impossible to predict the new relations by merely learning on the vanilla TACRED dataset without these new relations, we delete all examples with the newly introduced relations in Re-TACRED to accommodate our task.
Trained and evaluated on the Re-TACRED dataset, an average of 14.3\% improvement of $F_1$-score could be observed, which indicates that Re-TACRED essentially enhanced the annotation quality and could assess relation extraction models more faithfully.

Similarly, we exploit the Re-TACRED to appraise to what extent our proposed automatic re-annotator can relabel the dataset with higher consistency by merely learning on noisy data. We recompose the Re-TACRED dataset for upstream evaluation following the same strategies on TACRev. The dataset includes 10,729 examples that have been relabeled and still held relations already existed in the original TACRED and TACKBP relation set. To make the results on Re-TACRED comparable to TACRev, we only keep the revised examples, shuffled and then evenly split into Dev set with 5264 examples and Test set with 5365 examples that covered 36 different relations.

\subsection{New Revised Dataset: Re-DocRED}
We create the first dataset containing manual re-annotations in document-level RE, called Re-DocRED. It is a revised subset of DocRED, containing the revision of examples with challenging annotation inconsistencies and errors.

Since the labels of the Test set of DocRED are non-public, the Re-DocRED is derived from the Train and Dev sets of DocRED, which only have 50503 relation facts. Re-DocRED includes 411 re-annotations totally and is evenly split into Dev and Test set with the coverage of 55 relation types (Table \ref{tab:datarev}). Re-DocRED dataset is sampled and annotated in two phases:
\paragraph*{Data Selection} 
We firstly follow the similar data selection strategy proposed in TACRev \citep{DBLP:conf/acl/AltGH20a} to make maximum use of our restricted human labour.
The potentially challenging examples for annotation are screened out according to the disagreement of the predictions by multiple state-of-the-art RE models. Specifically, we fine-tuned the CorefBERT \citep{ye-etal-2020-coreferential} with BERT-base, ATLOP \citep{DBLP:conf/aaai/XuWLZM21} with RoBERTa-large, and SSAN \citep{DBLP:conf/aaai/XuWLZM21} with RoBERTa-base and RoBERTa-large on the Train set and contrast their predictions with the human annotations on the Dev set of DocRED to figure out the error-prone labels from Dev examples. Unlike the criteria applied in TACRev, which is to select the misclassified examples by at least half of the models, we empirically observe that if human annotations are different from all these four model predictions, the annotations are likely to be incorrect or inconsistent on DocRED. This automatic data selection procedure narrows the scope for human revision to 560 examples.
\paragraph*{Manual Annotation}
The selected examples are firstly validated by a recent graduated undergraduate student in Linguistics and then inspected by the author of this dissertation, a master by research student in Linguistics with a bachelor degree in Computer Science. The validation and inspection are independently conducted via the online annotation platform based on INCEpTION \footnote{\url{https://inception-project.github.io}} \citep{tubiblio106270}. It is an open-source semantic annotation platform built by UKP Lab at TU Darmstadt. The manual validation and inspection procedures roughly take 35 and 10 hours, respectively. Consequently, 411 high-quality re-validated examples from DocRED Dev set are selected to compose the Re-DocRED dataset, including  57.5\% examples with revised labels and 42.5\% examples with original labels. 

\chapter{Annotation Inconsistency Detection}
\label{chap:AID}
Annotation Inconsistency Detection testifies whether each annotation is consistent with other majority annotations in a similar context. We explore various relation representation methods and neighbouring consistency measurements for the zero-shot inconsistency detectors in relation extraction. The results suggest that our proposed credibility score combined with the relation prompt effectively detects inconsistencies.

\section{Overview}
This chapter will discuss the Annotation Inconsistency Detection (AID) based on the static relation representations directly from the Pre-trained Language Models (PLMs) without fine-tuning. The detailed definition of the AID task and the mathematical notations are introduced in Section \ref{sec:aid_def}.

To achieve the best results from zero-shot AID models, we first explore the prompt-based and entity-based methods to acquire the relation representations with full utilization of prior knowledge in PLMs. Based on the semantic sensitive PLM-based relation representations, we approximate the neighbouring consistency with the K-Nearest Neighbours algorithm and a novel credibility score jointly computed by the distance and trustworthiness of retrieved neighbours. Consequently, we propose two approaches for detecting annotation inconsistencies: (1) compare the observed annotation with the majority annotations of K-Nearest Neighbours, and (2) consider the observed annotation with credibility lower than a certain threshold as inconsistent. 

The empirical experiments show that both prompt-based and entity-based methods can result in the PLM-based relation embedding with needed semantic sensitivity for distinguishing inconsistent relation labels. Our proposed credibility scores combined with proper relation representations demonstrate impressive capability in detecting annotation inconsistencies, reaching the binary $F_1$ up to 92.4 on TACRev, 92.1 on Re-TACRED and 72.5 on Re-DocRED. Through the visualization, we reveal that overconfident prompts could downgrade the sensitivity and specificity of AID models.

\section{Methodology}
\begin{itemize}
\item Section \ref{sec:plm} briefly introduces the technical details of PLMs, such as types of special tokens and pre-training tasks, for better understanding our proposed relation representation methods. 
\item Section \ref{sec:rel_rep_meth} describes prompt-based and entity-based methods for obtaining PLM-based relation representation. 
\item Given noise-sensitive relation representations, Section \ref{sec:neigh} proposes the K-Nearest Neighbours and credibility score based AID models.
\end{itemize}

\subsection{Pre-trained Language Models}
\label{sec:plm}

Large-scale Pre-trained Language Models (PLMs) as the contextualized word embedding techniques have become the dominant workhorse in NLP because in various downstream tasks, they lead to a better performance than traditional fixed word embedding methods such as one-hot embedding Glove \citep{pennington-etal-2014-glove} and Word2Vec  \citep{NIPS2013_9aa42b31}. The strength of PLMs could be roughly attributed to: (1) massive learnable parameters allow models to capture richer interactions between each token in the context and (2) large-scale pre-training endows models with abundant common and linguistic knowledge. Recent studies \citep{petroni-etal-2019-language, DBLP:conf/acl/GaoFC20, DBLP:conf/iclr/CaoI0P21, DBLP:journals/corr/abs-1909-03193, hewitt-manning-2019-structural, shin-etal-2020-autoprompt, DBLP:conf/emnlp/DavisonFR19, DBLP:journals/tacl/JiangXAN20,DBLP:journals/tacl/TalmorEGB20} imply that properly designed prompts for formatting the text input could explicitly leverage intrinsic knowledge from PLMs to enhance few-shot learning tasks. 

\begin{figure}[h]
\subfloat[The architecture of the multi-layer bidirectional Transformer encoder.]{\label{fig:transformer_encoder}
\includegraphics[height=0.3\textheight]{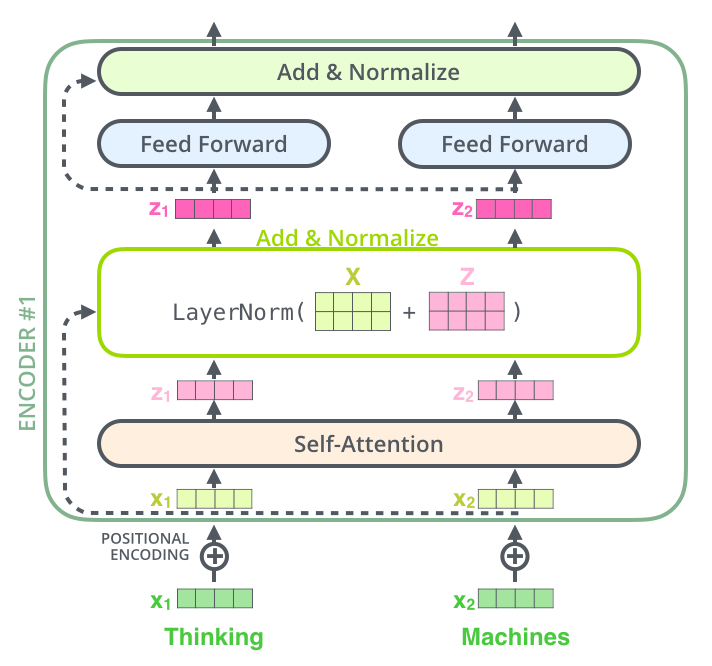}}
\hspace*{\fill}
\subfloat[The architecture of multi-head attention, the basic component of self-attention layer.]{\label{fig:attention_head}
\includegraphics[height=0.3\textheight]{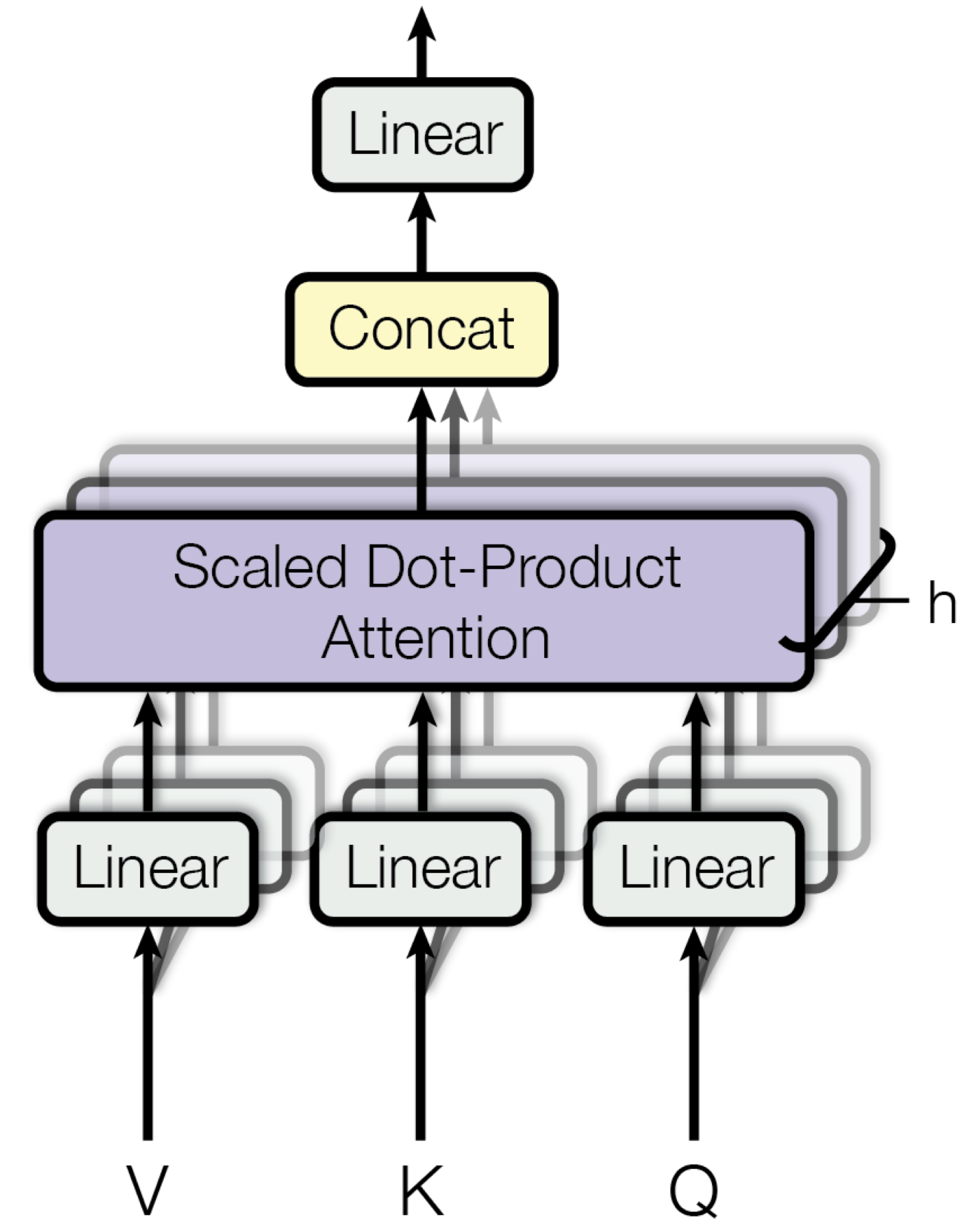}}
\caption{The architecture of the backbone model of BERT. Figures from \citet{DBLP:conf/nips/VaswaniSPUJGKP17}. }
\label{fig:transformer}
\end{figure}

Since only BERT was used in this project, this section will only introduce the details of architectures and pre-training tasks proposed by \citet{devlin-etal-2019-bert}. The architecture of its backbone model is identical to the multi-layer bidirectional Transformer encoder described by \citet{DBLP:conf/nips/VaswaniSPUJGKP17}. The encoder typically is composed of a stack of self-attention layer and position-wise fully connected feed-forward layer with residual connections (Figure \ref{fig:transformer_encoder}). The self-attention layer exploits the multi-head mechanism with scaled dot-product attention (Figure \ref{fig:attention_head}) to capture the mutual interaction within the input sequence. 

Several special tokens are introduced to the vocabulary to support BERT in diverse downstream tasks: \textbf{\textsc{[CLS]} token:} special classification token used to acquire the sentence-level representation for sequence classification tasks (e.g. sentiment classification); \textbf{\textsc{[SEP]} token:} Separator token used to differentiate the consecutive sentences; \textbf{\textsc{[MASK]} token:} mask token used to replaced the tokens indented to be masked during pre-training.

\begin{figure}[h]
\includegraphics[width=\linewidth]{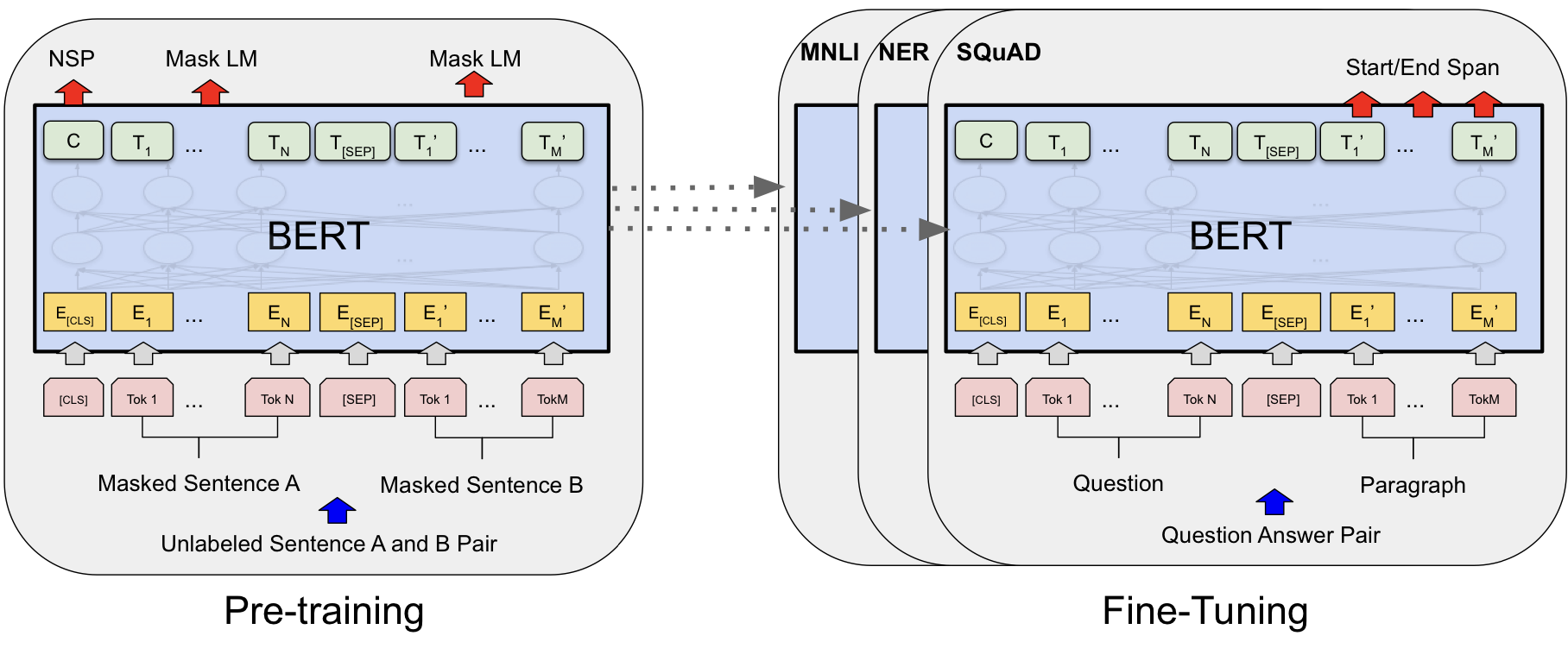}
\caption{The PLMs framework includes two stages: (1) pre-training on general purpose tasks such as Masked Language Model on large-scale corpus; (2) fine-tuning for the downstream tasks on domain-specific datasets. Figure from \citet{devlin-etal-2019-bert}.}
\label{fig:plm}
\end{figure}

The pre-training data of BERT is two large-scale document-level corpora, BooksCorpus \citep{Zhu_2015_ICCV} with 800M words and English Wikipedia with 2,500M words. As shown in Figure \ref{fig:plm}, BERT is pre-trained with two unsupervised tasks:
\begin{itemize}
\item \textbf{Masked Language Model (MLM):} Some percentage of the input tokens are masked randomly, and then the model learns to predict the masked tokens similarly to the Cloze-filling task. As the \textsc{[MASK]} token usually does not appear in the downstream tasks, only 80\% of the time masked tokens are replaced by \textsc{[MASK]} token, while 10\% of the time by random tokens and 10\% of the time by unchanged tokens, to avoid mechanical memorization.
\item \textbf{Next Sentence Prediction (NSP):} Given sentence \texttt{A}, the model learns to predict its next sentence \texttt{B}. 50\% of the time \texttt{B} is the actual next sentence that follows \texttt{A}, while 50\% of the time \texttt{B} is not. Argued by \citet{DBLP:journals/corr/abs-1907-11692}, this task is not as informative as the MLM task.
\end{itemize}

\subsection{Relation Representation Methods}
\label{sec:rel_rep_meth}

The basis of AID in Relation Extraction (RE) is to acquire the relation representations with desired semantic meaning from the context and entity mentions. As mentioned in Section \ref{sec:basis}, the main assumption of the AID task is that the representations of inconsistent samples should have different observed relations with their neighbours in the embedding space. Thus, we explore two methods to derive the relation representation from PLMs without fine-tuning: (1) prompt-based method, and (2) entity-based method. In this zero-shot scenario, guaranteeing their capability of fetching needed prior knowledge from PLMs is challenging and consequently determines the final performance of AID models. However, as PLMs perform only the inference stage, it will be viable if the computational resource is limited.

\subsubsection*{Prompt-based Representations}
\label{sec:prompt}
As described in Section \ref{sec:rel_prompt}, PLMs acquire abundant prior knowledge from a large corpus through pre-training tasks MLM and NSP. However, the gap between pre-training tasks and downstream tasks restricts knowledge transferability. Hence, Prompt-based Learning with PLMs becomes the new paradigm in natural language processing, narrowing this gap by formulating downstream tasks into mimic LM tasks and introducing inductive words. \citep{DBLP:journals/corr/abs-2107-13586}. The conventional approach trains PLMs to directly predict the output $y$ given the input context $c$ by $P(y \mid c)$, and $y$ is from a different label set without explicit connection with the vocabulary of PLMs. Nevertheless, prompt-based methods modify the input $c$ using a template into a new textual string prompt $c'$ with unfilled slots, and then the predictions are acquired by observing the discrete guesses or processing latent representations of PLMs on these masked slots. In the AID task, we explore methods to acquire the relation representation from the latent representations of masked slots. If the prompt $c'$ is properly designed, this approach naturally narrows the gap between the pre-training and downstream domain-specific tasks, such as AID. It enables our models to readily perform few-shot, or even zero-shot learning with minimum loss of prior knowledge of PLMs learnt from the large-scale pre-training data.

The prompt we designed for AID comprises two parts, the context given in each example and a template following it. The context part is a single sentence on TACRED-based datasets or a document on DocRED-based datasets. It relieves the disparity between the pre-training MLM task and our target AID task by providing explicit context. The prompts contain the mentions of head and tail entities and a \textsc{[MASK]} token in the position $m$ for deducing the relation representation $\mathbf{e}_r$ by:
\begin{equation*}
\mathbf{e}_r = \textbf{h}(\text{[MASK]}) = PLM(\text{prompt})_{m}
\end{equation*}, where $\textbf{h}(\text{[MASK]}) $ is the last hidden states in the masked position obtained from a PLM.

As shown in Table \ref{tab:pos_temp}, we explored five different variations from three types of template:
\begin{itemize}
\item \textbf{Fixed Templates}: The first and second methods fix a single template for all relation types.
\item \textbf{Hand-written Templates}: The third method manually assigns every relation type with hand-written templates.
\item \textbf{Generated Templates}: The fourth and fifth methods generate the templates for each example. They ask the PLMs to fill one token between head mention and [MASK], and another one token between [MASK] and tail entity mentions, with the following steps: (1) the example likes $r'(\texttt{Bill Gates}, \texttt{Microsoft})$ is converted into the template \texttt{Bill Gates [MASK] Microsoft}; (2) If the PLMs fill the masked slot with the token \texttt{,}(comma), we fill the [MASK] in the previous template with \texttt{,}(comma) and insert a new [MASK] after this decoded token to convert the previous templates into \texttt{Bill Gates, [MASK] Microsoft}; (3) We repeat above steps until three consecutive tokens are filled between head and tail mentions, like \texttt{Bill Gates, CEO of Microsoft}; (4) We then mask the middle token of the generated three consecutive tokens to become the prompt used for acquiring relation representations, like \texttt{Bill Gates, [MASK] of Microsoft}. The fourth method is to directly fill each masked slot with the top candidate given by PLMs, while the fifth method randomly selects one of the top three candidates to fill each slot.
\end{itemize}

\begin{table}[h]
\centering
\resizebox{\textwidth}{!}{%
\begin{tabular}{@{}ll@{}}
\toprule
Templates                                     & Examples                              \\ \midrule
{[}HEAD{]} is {[}MASK{]} of {[}TAIL{]}        & Bill Gates is {[}MASK{]} of Microsoft       \\
{[}HEAD{]} {[}MASK{]} {[}TAIL{]}              & Bill Gates {[}MASK{]} Microsoft             \\
{[}Human-tuned template for each relation type{]}      & Bill Gates works as {[}MASK{]} of Microsoft \\
{[}Auto template filled with top candidate{]} & Bill Gates, {[}MASK{]} of Microsoft         \\
{[}Auto template filled with top  3 candidates randomly{]} & Bill Gates was {[}MASK{]} of Microsoft                        \\ \bottomrule
\end{tabular}%
}
\caption{The examples of different templates for prompt-based relation representation. In the examples, subjective entity is "Bill Gates" and objective entity is "Microsoft".}
\label{tab:pos_temp}
\end{table}

\subsubsection*{Entity-based Representations}
\label{sec:marker}
Aside from deriving the relation representations from masked tokens as prompt-based methods, fully exploiting the entity-wise embeddings is an alternative to acquiring the relation representations. Generally, salient notation of head and tail entity mentions could provide considerable hints for PLMs to discern the authentic relation between them. We follow the entity representation techniques discussed by \citet{DBLP:journals/corr/abs-2102-01373}, to obtain the relation embedding $\textbf{e}_r$ by concatenating the contextualized embeddings $\textbf{e}_{sub}$ and $\textbf{e}_{obj}$ of subject and object entity as follows:
\begin{gather*}
\textbf{e}_{entity} = \textbf{h}_{e} = PLM([x_0, ...., x_n])_e \\
\textbf{e}_r = [\textbf{e}_{sub} : \textbf{e}_{obj}]
\end{gather*}
, where $x_i$ is the input token, $\textbf{h}_{\text{e}} $ is the last hidden states of the input token in the position of \textit{entity pointer} $e$. Considering the entities usually appear more than once in the context with the markers in different places $[\textbf{e}_0, ..., \textbf{e}_n]$ on DocRED-based datasets, we use the average of all the hidden states of entity markers $\textbf{h}_{e_n}$ as the entity representation:
\begin{equation*}
\textbf{e}_{entity} = avg([\textbf{h}_{e_0}, .., \textbf{h}_{e_n}])
\end{equation*}

As exemplified in Table \ref{tab:input_format}, the investigated techniques for entity representations include:
\begin{itemize}
\item \textbf{Entity position}: It takes the first tokens of head and tail entities as the \textit{entity pointers} without modifying the given context.
\item \textbf{Entity marker} \citep{DBLP:conf/acl/ZhangHLJSL19, DBLP:conf/acl/SoaresFLK19}: It introduces two special tokens pair \texttt{[H]}, \texttt{[/H]} and \texttt{[T]}, \texttt{[/T]} to enclose head and tail entities, which indicates the positions and spans of entities. The tokens \texttt{[H]} and \texttt{[T]} are \textit{entity pointers}.
\item \textbf{Entity marker (punct)} \citep{WenxuanZhou2020DocumentLevelRE}: It is similar to the entity marker but replaces the special tokens pair by punctuation existing in the vocabulary, such as \# and @. The punctuation in front of each entity mention is the \textit{entity pointer}.
\item \textbf{Entity mask}\citep{YuhaoZhang2017PositionawareAA}: It adds the new special tokens \texttt{[SUBJ-TYPE]} and \texttt{[OBJ-TYPE]} to mask the spans of the head and tail entities, where \texttt{TYPE} is substituted by their named entity types. The special tokens \texttt{[SUBJ-TYPE]} and \texttt{[OBJ-TYPE]} are \textit{entity pointers}.
\item \textbf{Typed entity marker} \citep{ZexuanZhong2020AFE}: It is similar to the entity marker but further provides the information regarding entity types.
\item \textbf{Typed entity marker (punct)}: It is similar to the Entity marker (punct) but further provides the information of named entity types.
\end{itemize}

\begin{table}[h]
\centering
\resizebox{\textwidth}{!}{%
\begin{tabular}{@{}ll@{}}
\toprule
Input Formats                        & Examples                                                                                   \\ \midrule
{Entity position} & { \underline{B}ill Gates founded \underline{M}icrosoft.} \\
{Entity marker} & { {[}H{]} Bill Gates {[}/H{]} founded {[}T{]} Microsoft {[}/T{]}.} \\
Entity marker (punct)                & @ Bill Gates @ founded \# Microsoft \#.                                                    \\
Entity mask                          & {[}SUBJ-PERSON{]} founded {[}OBJ-CITY{]}.                                                  \\
Typed entity marker &
 \textless{}S:PERSON\textgreater Bill Gates \textless{}/S:PERSON\textgreater founded \textless{}O:CITY ....
  \\
Typed entity marker (punct)          & @ [ person ] Bill Gates @ founded \# ! city ! Microsoft \#.                                \\ \bottomrule
\end{tabular}%
}
\caption{Different input formats to highlight the entity mentions for entity-based relation representations.}
\label{tab:input_format}
\end{table}

\subsection{Neighbouring Consistency}
\label{sec:neigh}
Neighbouring consistency is a vital clue for detecting abnormal annotations. Intuitively, the annotation reliability of each example may be judged by the similarity between its embedding and the embeddings of other examples with the same label. Hence, we investigate the neighbouring consistency based on the distribution of all relation embeddings encoded by PLMs in the representation space without fine-tuning PLMs on downstream tasks. We propose two approaches in order to capture annotation inconsistencies based on neighbouring correspondence: (1) K-Nearest Neighbours (KNN) and (2) Kernel Density Estimation (KDE) based detectors.

\begin{figure}[h]
\includegraphics[width=\linewidth]{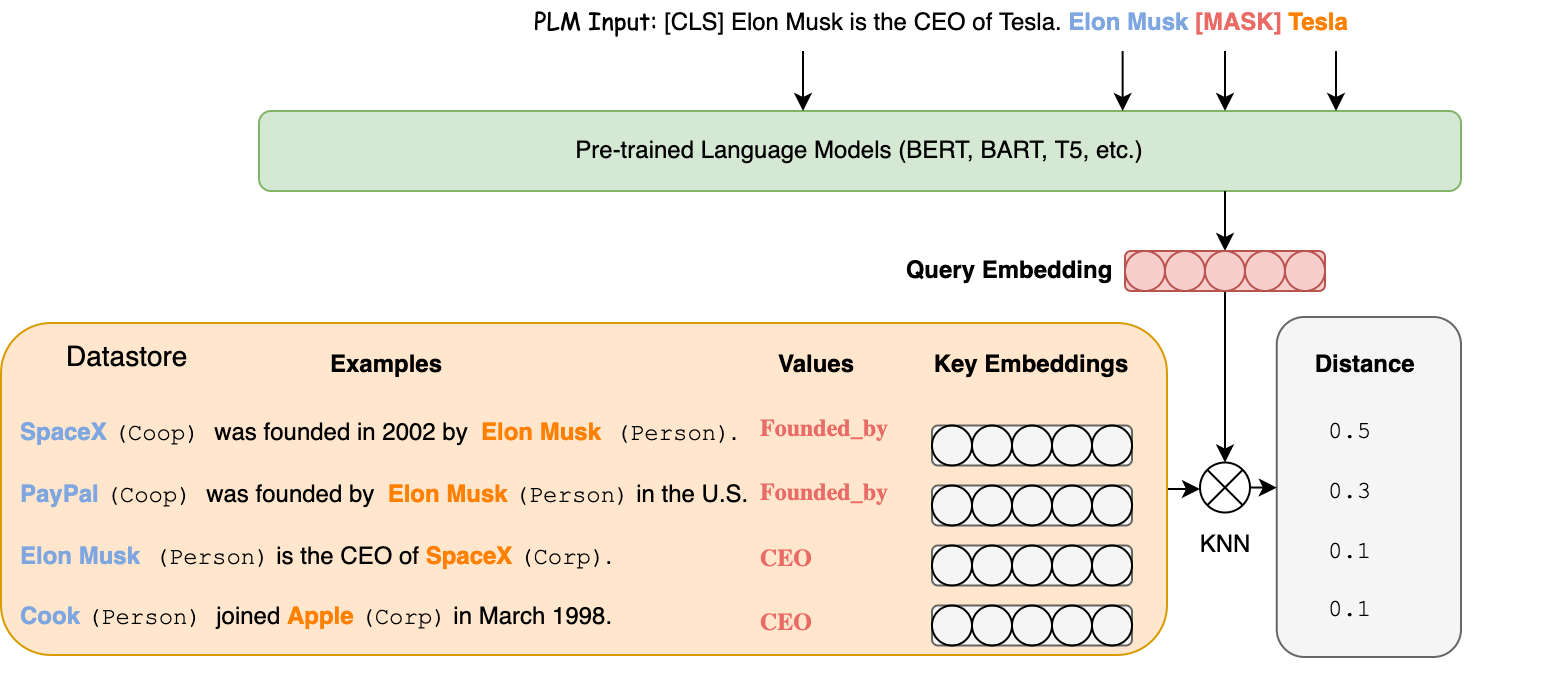}
\caption{An example of retrieving neighbours of a query with prompt-based representation based on the K-Nearest Neighbours algorithm.}
\label{fig:knn}
\end{figure}

The K-Nearest Neighbours (KNN) algorithm \citep{fix1989discriminatory} is a non-parametric method for classification by a plurality vote of its neighbours. As suggested by \citet{DBLP:conf/acl-louhi/GrivasAGTW20}, analyzing the nearest neighbour annotations retrieved by KNN may effectively reveal potential annotation inconsistencies. We exploit the framework similar to \citep{DBLP:conf/iclr/KhandelwalLJZL20} to retrieve neighbours of the query embedding by squared Euclidean distance (Figure \ref{fig:knn}). The query embedding is the relation representation of the example in the Test and Dev set, acquired by PLMs with prompts or entity markers. The datastore contains the keys $k$ and values $v$ : (1) keys $k$ are the relation representations of all instances in the Train set, and (2) values $v$ are observed relations. Therefore, given a query, the KNN algorithm would return a list of nearest neighbours sorted by the distance metrics that measure the similarity between the query and stored keys. Then, we further propose the following vote-based and credibility-based methods to verify the query annotations by analyzing the retrieved neighbours.

\subsubsection*{Vote-based Detection}
The vote-based detection follows the conventional approach of KNN-based classification that determines if the annotation of each example is consistent with the majority voting of their neighbours. That means, if the annotation of the example is the same as the most frequent annotations of its neighbours, the vote-based models predict the annotation as valid.

\subsubsection*{Credibility-based Detection}
Since the vote-based methods only rely on the closest neighbours and consider them as isolated, we propose a credibility-based approach for testifying the annotation consistency.
Inspired by the inference function in \cite{DBLP:conf/iclr/KhandelwalLJZL20}, we define a novel credibility score combining both the local neighbouring and the global distributional information. 

Assuming each instance in retrieved neighbours $\mathcal{N}$ has relation $r_{n_i}$ and relation representation $\textbf{e}_{n_i}$, the credibility score $\psi_i$ of each annotation with relation $r_i$ and relation embedding $\textbf{e}_i$ could be computed by first aggregating probability mass $s_i$ across all its neighbours with the same annotation and then rescaling to $\psi \in [0, 1]$:
\begin{gather} 
s_i =  \sum^{r_i = r_{n_i}}_{ (r_{n_i}, \textbf{e}_{n_i})\in \mathcal{N}} f_K(\textbf{e}_{n_i})exp(-d(\textbf{e}_{n_i} ,\textbf{e}_{i})), \hspace{2mm} s_i \in \mathcal{S} \label{eq:entropy} \\
\psi_i= norm(s_i) = \frac{s_i - min(\mathcal{S})}{max(\mathcal{S}) - min(\mathcal{S})}
\end{gather}, where $exp$ is the exponential function.

The $d$ in Equation \ref{eq:entropy} is the normalized squared Euclidean distance function, computed as follows:. 
\begin{gather}
d(\textbf{e}_{n_i} ,\textbf{e}_{r_i}) = \frac{|\textbf{e}_{n_i} ,\textbf{e}_{r_i}|}{max(\mathcal{D})} , \hspace{2mm} |\textbf{e}_{n_i} ,\textbf{e}_{r_i}|\in \mathcal{D}
\end{gather}It gathers the local information by measuring the distance between an example and its neighbours with the same relation type. We normalize the distance to $d(\textbf{e}_{n_i}, \textbf{e}_{r_i}) \in [0, 1]$ to prevent $exp(-d(\textbf{e}_{n_i} ,\textbf{e}_{i}))$ from being too close to 0, which helps keeping the information from remote neighbours. The closer neighbours result in higher $exp(-d(\textbf{e}_{n_i} ,\textbf{e}_{r_i}) )$, which means the closer neighbours contribute more significantly to the credibility score.

The $f_K$ in Equation \ref{eq:entropy} is the Probability Density Function using a Gaussian Kernel \citep{DBLP:books/lib/Murphy12}. We utilize Kernel Density Estimation (KDE), a method to estimate the probability density function, to capture the global information from the overall embedding distribution of each relation type. If the probability density $f_K(\textbf{e})$ is high, the embedding is near the centroid of distribution of all examples with identical relation types. Intuitively, we believe that the neighbours with higher probability density regarding their annotated relation type are more trustworthy for contributing to the credibility score. Noted the $\mathcal{K}$ as the standard normal distribution function, the KDE function $f_K$ could compute the probability density of an example with the embedding $e$ and relation $r$ as the relation type $t$ as follows:
\begin{gather}
f_K(\textbf{e}) = f^t_K(\textbf{e}) = \frac{1}{|\mathcal{E}_t|h}\sum_{\textbf{e}_{i} \in\mathcal{E}_t}\mathcal{K}(\frac{\textbf{e} - \textbf{e}_i}{h}) \label{eq:kde}
\end{gather}, where $h$ is the bandwidth, and $\mathcal{E}_t$ is the set of embeddings of all examples with relation type $t$.

Finally, the prediction $y_i$ is obtained by comparing the credibility score $\psi_i$ with a threshold $\beta\in [0, 1]$.
\begin{gather}
y_{i} = 
\begin{cases}
    inconsistent, &  \psi_i  < \beta\\
    consistent, &     \psi_i \geq \beta
    \label{eq:threshold}
\end{cases}
\end{gather}
\section{Evaluation}
The performance of the proposed AID approaches is evaluated based on the consistency between human revisions and model predictions. Hence, by regrading the decisions made by human revisers as the ground-truth, the performance of different relation embedders and the results of AID models can be quantified with rank-based metrics and classification metrics, respectively.

\paragraph*{Rank-based Metrics}
Rank-based Metrics are able to measure the sorted retrieval results obtained by KNN  models directly. The relation representation methods are expected to map the embeddings of examples with the same true annotations to be close together. Hence, given a particular relation representation as input, the examples with the true annotation of the query example in its neighbours are expected to be ranked as high as possible by the KNN models. The Hit@1, Hit@5, Hit@10, and Mean Reciprocal Rank (MRR) \citep{DBLP:conf/lrec/RadevQWF02} are used to evaluate the relation representation methods.

\paragraph*{Classification Metrics}
Classification Metrics are sensible to evaluate the model performance on AID tasks because the AID is a binary classification task. Accuracy and binary $F_1$ score \citep{grishman1996message} are utilized to assess if the AID models are in agreement with human revisers. Accuracy intuitively reflects the proportion of predictions identical to the manual revisions. At the same time, $F_1$ score is the harmonic mean of the precision and recall, demonstrating a better measure of the incorrectly classified cases.

\section{Experiments}
\begin{itemize}
\item Section \ref{sec:aid_implement} presents the implementation details of all experiments in AID. 
\item Section \ref{sec:exp_rel_rep} demonstrates the experimental results regarding different relation representation techniques described in \ref{sec:rel_rep_meth}.
\item Section \ref{sec:aid_classify} shows the outcomes of different inconsistency detecting strategies mentioned Section \ref{sec:neigh}.
\end{itemize}

\subsection{Implementation Details}
\label{sec:aid_implement}
\paragraph*{Pre-trained Language Model}
The off-the-shelf backbone model we used is the pre-trained \textsc{bert-base-cased}\footnote{\url{https://huggingface.co/bert-base-cased}} from Hugging Face. It has the maximum length for the input sequence as 512 tokens, and the tokenizer is based on WordPiece\cite{DBLP:journals/corr/WuSCLNMKCGMKSJL16}. The multi-head encoder has 12 attention heads, and the dropout probability between adjacent hidden layers is 0.1. It totally contains 12 hidden layers, and both the embedding and hidden layers have dimensions of 768. Therefore, the dimension of the relation representations acquired by prompt-based methods (Section \ref{sec:prompt}) would be 768, whereas 1,536 of the embeddings obtained by entity-based methods.

\paragraph*{K-Nearest Neighbours Searching}
The K-Nearest Neighbours retrieval is implemented with Faiss\footnote{\url{https://github.com/facebookresearch/faiss}} \citep{JDH17}. Faiss is the library including several methods for high-performance similarity search in the embedding space. As for the Faiss implementation, each instance is assumed to be a vector embedding and indexed by an integer, where the vectors can be compared with squared Euclidean distances. Examples that are similar to a query are those that have the vector embeddings with the lowest squared Euclidean distance with the query vector.

\paragraph*{Handling Long Sequences}
As the contexts of the examples in DocRED is the entire document, the input sequences sometimes would exceed the maximum input length of \textsc{bert-base-cased} model. Therefore, we incrementally truncate the long sequences with the following modes in order until the input lengths meet the requirement:
\begin{itemize}
\item \textbf{Mode 1}: It remains the sentences between the first and the last sentences that contain the mentions of arbitrary entities.
\item \textbf{Mode 2}: It remains the sentences including the mentions of arbitrary entities.
\item \textbf{Mode 3}: It iteratively decreases the number of sentences containing entities until the input length is shorter than the maximum input length.
\end{itemize}

\paragraph*{Datasets}
Both the vote-based KNN detector and the credibility detector require the information of the existing relation embedding distribution. All 103,738 examples of TACRED and 50,503 examples of DocRED are used to produce the embedding datastore. As for KNN-based methods, the vector representations of relations are used as the keys, and the original annotations of examples are regarded as the values for the neighbour search. As for KDE-based methods, the relation embeddings contribute to the training of the KDE model of their belonging classes according to the original annotations in TACRED and DocRED. We evaluate the performance of different AID systems based on the manual revisions provided in TACRev, Re-TACRED and Re-DocRED: If the manual revision is identical to the original annotations, the target label is \texttt{True} and vice-versa. First, we use the Dev and Test sets of TACRev, both containing 1,263 human revisions for systematical evaluation on relation encoders. Then, additionally to TACRev, we take 5,364 and 5,365 revisions from the Dev and Test sets of Re-TACRED, and 206 and 205 revisions from the Dev and Test sets of Re-DocRED as the ground-truth to optimize the hyperparameters of inconsistency detectors and comprehensively evaluate their performance. More details of the datasets used to conduct the experiments are presented in Section \ref{sec:data}.

\paragraph*{Hyper-parameters}
The hyper-parameters of KNN-based methods include the number $k$ of voting neighbours, and we only explore the cases of $k=1$ and $k=3$. The hyper-parameters of credibility-based methods are the number of retrieved neighbours for computing the credibility score, the bandwidth $h$ of the KDE model in Equation \ref{eq:kde} and the threshold $\beta$ of the credibility-based classifier in Equation \ref{eq:threshold}. The number of retrieved neighbours was manually set to be 250. By tuning hyper-parameter with Bayesian optimization \citep{snoek2012practical} on the Dev sets of TACRev, Re-TACRED and Re-DocRED, the bandwidth $h$ was optimized as 0.25, and the threshold $\beta$ was set as 0.5 for all datasets.

\paragraph*{Baselines}
To properly evaluate the performance of different relation representation methods, we regard the three sentence-level relation representations as to the baseline models: (1) the sentence representation by feeding the last hidden states of the \texttt{[CLS]} token into the pre-trained BERT pooling layer for the NSP task; (2) the sentence embedding by average pooling over the hidden states of all tokens in the sequence; (3) the aggregated embedding by feeding the last hidden states of all tokens into the maximum pooling layer.

\paragraph*{Experimental Environments}
The following experiments were conducted on a single GeForce GTX 1080 Ti with 12GB graphic memory and CUDA version of 11.0. The proposed models are implmented with the PyTorch 1.9.0\footnote{\url{https://pytorch.org}} \citep{DBLP:journals/corr/abs-1912-01703} , Transformers 4.3.3\footnote{\url{https://github.com/huggingface/transformers}} \citep{wolf-etal-2020-transformers}, Scikit-learn 1.0.1\footnote{\url{https://scikit-learn.org/}} \citep{scikit-learn} , Numpy 1.20.3\footnote{\url{https://numpy.org}} \citep{ harris2020array}  and GPU-based Faiss 1.7.1.

\begin{table}[h]
\centering
\resizebox{\textwidth}{!}{%
\begin{tabular}{llllll}
\hline
\multicolumn{2}{l}{Representation Method}                                           & Hit@1 & Hit@5         & Hit@10        & MRR  \\ \hline
\multirow{3}{*}{\begin{tabular}[c]{@{}l@{}}\textit{Baseline}\\ Sentence-level\end{tabular}} & Pooler Layer & 41.1 & 64.4 & 70.3 & 51.5 \\
                                & Average Pooling                                   & 41.3  & 67.4          & 77.1          & 53.0 \\
                                & Max Pooling                                       & 40.0  & 67.6          & 77.6          & 52.8 \\ \hline
\multirow{5}{*}{Entity-based}   & Entity position                                 & 57.2  & 86.7          & 92.7         & 70.0\\
                                &  Entity marker                                 & 54.2  & 88.0          & 94.1          & 68.7 \\
                                & Entity marker (punct)                             & 56.6  & \textbf{89.5} & \textbf{94.5} & 70.3 \\
                                & Entity mask                                       & 57.9  & 88.9          & 93.7          & 70.7 \\
                                & Typed entity marker                               & 55.2  & 88.9          & 94.2          & 69.7 \\
                                & Typed entity marker (punct)                       & 56.8  & 87.5          & 93.2          & 70.2 \\ \hline
\multirow{5}{*}{Prompt-based} & {[}HEAD{]} is {[}MASK{]} of {[}TAIL{]}                    & 65.1          & 88.5 & 93.6 & 75.8          \\
                              & {[}HEAD{]} {[}MASK{]} {[}TAIL{]}                          & \textbf{67.2} & 88.4 & 93.5 & \textbf{76.9} \\
                                & {[}Human-tuned template for each relation type{]} & 5.3   & 8.1           & 8.4           & 6.4  \\
                                & {[}Auto template filled with top candidate{]}     & 2.3   & 4.5           & 6.8           & 3.7  \\
                              & {[}Auto template filled with top 3 candidates randomly{]} & 10.2          & 23.8 & 31.1 & 16.7          \\ \hline
\end{tabular}%
}
\caption{The accuracy and binary $F_1$ of the relation representations obtained by sentence-level, prompt-based, entity-based embedders on TACRev Test set.}
\label{tab:rank}
\end{table}

\subsection{Representation Methods of Relation}
\label{sec:exp_rel_rep}
Table \ref{tab:rank} shows the Hit@1, Hit@5, Hit@10 and MRR of the KNN query results based on the embedding acquired by sentence-level embedders, entity-based embedders and prompt-based embedders. The sentence-level representation approaches are regarded as the baselines. Generally, the average pooling of all token embeddings in the context is the best representation method among the sentence-level encoders, reaching the Hit@1 of 41.3 and MRR of 53.0.

As the embeddings of the newly introduced marker tokens are randomly initialized without downstream fine-tuning, it is noticeable that using the punctuation to mark the entity spans slightly outperforms the methods with new special tokens. For instance, the \textsc{entity marker (punct)} that marks the entity with punctuation likes \texttt{\#} increases the Hit@1 by 2.4 and MRR by 1.6 compared to the \textsc{entity marker} that marks the entity with special tokens like \texttt{[H]}. Similarly, \textsc{Typed entity marker (punct)} enables the KNN-based detector to achieve higher Hit@1 and MRR than \textsc{Typed entity marker}. In contrast, the \textsc{Entity position} and \textsc{Entity mask} without including any special tokens or new language patterns that PLMs have not been exposed to in the pre-training tasks are the most competitive methods in the entity-based relation encoders. The KNN-based detector with \textsc{Entity mask} technique reaches the highest Hit@1 of 57.9 and MRR of 70.7 among all entity-based embedders, outperforming the best baseline by 40.2 \% in Hit@1 and 33.4 \% in MRR. The results indicate that methods that smooth the learning curve of PLMs lead to better relation representations, which potentially supports the arguments by \cite{DBLP:conf/iclr/SaunshiMA21} as briefly described in Section \ref{sec:basis}.

In contrast, the appropriate prompts result in even better compatibility between AID tasks and pre-training tasks than the entity-based representation methods. Both \textsc{[HEAD] is [MASK] of [TAIL]} and \textsc{[HEAD] [MASK] [TAIL]} templates show the better overall performances than the sentence-level and entity-based encoders. The prompt generated by \textsc{[HEAD] [MASK] [TAIL]} template leads to the optimal result of all relation representations with Hit@1 of 67.2 and MRR of 76.9, exceeding the best entity-based model by 8.8\% and the best baseline by 45.1\% in term of MRR. However, the human-tuned and automatically searched prompts dramatically downgrade the performance of relation embedders. According to the empirical studies, we find that this degradation could be accused to that the obvious hints in the prompts or contexts make the PLMs to be less sensitive to the annotation inconsistencies. A detailed discussion of the trade-off of relation representations will be presented in Section \ref{sec:trade_off}.

\subsection{Classification by Neighbouring Agreements}
\label{sec:aid_classify}
According to the experimental results of different relation representations, we finalize three representation methods for all following experiments: (1) \textbf{Relation Prompt}: the prompt generates by the \textsc{[HEAD] [MASK] [TAIL]} template, (2) \textbf{Entity marker}: the entity spans are wrapped by \texttt{[H]} or \texttt{[T]}, and (3) \textbf{Entity marker (punct)}: the entity spans are wrapped by \texttt{@} or \texttt{\#}. \textsc{[HEAD] [MASK] [TAIL]} is selected because of its overwhelming performance among prompt-based methods. As the entity-based methods, including the external NER knowledge, do not bring prominent advantages over other methods, we decline those methods to focus on the information that can be directly grabbed by the PLMs from the context. \textsc{Entity marker} and \textsc{Entity marker (punct)} are selected because their overall performances are archetypal among entity-based methods, and comparing the influences of different types of markers is of interest.

\begin{table}[h]
\centering
\resizebox{\textwidth}{!}{%
\begin{tabular}{ll|ll|ll|ll}
\hline
                               &               & \multicolumn{2}{l|}{TACRev} & \multicolumn{2}{l|}{Re-TACRED} & \multicolumn{2}{l}{Re-DocRED} \\ \cline{3-8} 
Detection Methods          & Relation Representation & Acc     & $F_1$      & Acc     & $F_1$      & Acc    & $F_1$     \\ \hline
\multirow{3}{*}{KNN-based (k=1)} & Relation Prompt           & 31.0 & 47.4 & 47.3 & 64.2 & 43.4 & 49.6 \\
                           & Entity marker           & 28.4 & 44.3 & 45.1 & 62.2 & 48.3 & 55.1 \\
                           & Entity marker (punct)   & 26.6 & 42.0 & 45.5 & 62.5 & 47.8 & 54.5 \\ \hline
\multirow{3}{*}{KNN-based (k=3)} & Relation Prompt           & 45.8 & 62.9 & 64.7 & 78.5 & 44.9 & 56.0 \\
                           & Entity marker           & 46.2 & 63.2 & 64.2 & 78.2 & 49.8 & 61.7 \\
                           & Entity marker (punct)   & 43.0 & 60.1 & 63.4 & 77.6 & 51.2 & 63.2 \\ \hline
\multirow{3}{*}{Credibility-based ($\beta$ = 0.5)} & Relation Prompt & \textbf{85.9}      & \textbf{92.4 }     & \textbf{85.3 }        & \textbf{92.1}         & 59.0        & 69.1        \\
                           & Entity marker           & 78.1 & 87.7 & 71.9 & 83.6 & 60.0 & 72.1 \\
                           & Entity marker (punct)   & 84.4 & 91.5 & 72.6 & 84.1 & \textbf{60.1} & \textbf{72.5} \\ \hline
\end{tabular}%
}
\caption{The accuracy and binary $F_1$ of KNN-based and credibility-based inconsistency detection methods with different relation encoders are evaluated on TACRev, Re-TACRED, Re-DocRED Test set. Credibility-based AID models obviously outperform other methods.}
\label{tab:aid}
\end{table}

Table \ref{tab:aid} illustrates the performances of the KNN-based detectors and credibility-based detectors on the Test set of TACRev, Re-TACRED, Re-DocRED. The KNN-based models with $k=3$ consistently surpass those with $k=1$, indicating that the annotation of the closest neighbour may be deceptive. The credibility-based indicators evidently outshine all KNN-based detectors with the increment of 39.7 in accuracy and 29.2 in $F_1$ score on TACRev, 20.6 in accuracy and 13.6 in $F_1$ score on Re-TACRED, and 8.9 in accuracy and 9.3 in $F_1$ score on Re-DocRED at least. It proves that our proposed credibility-based score is more effective in detecting annotation inconsistencies by jointly considering the local geometry of neighbours and the global embedding distributions of each class.

\begin{figure}[h]
\subfloat[The distribution of the credibility scores on TACRED.]{\label{fig:tacred_ent}
\includegraphics[height=0.3\textheight]{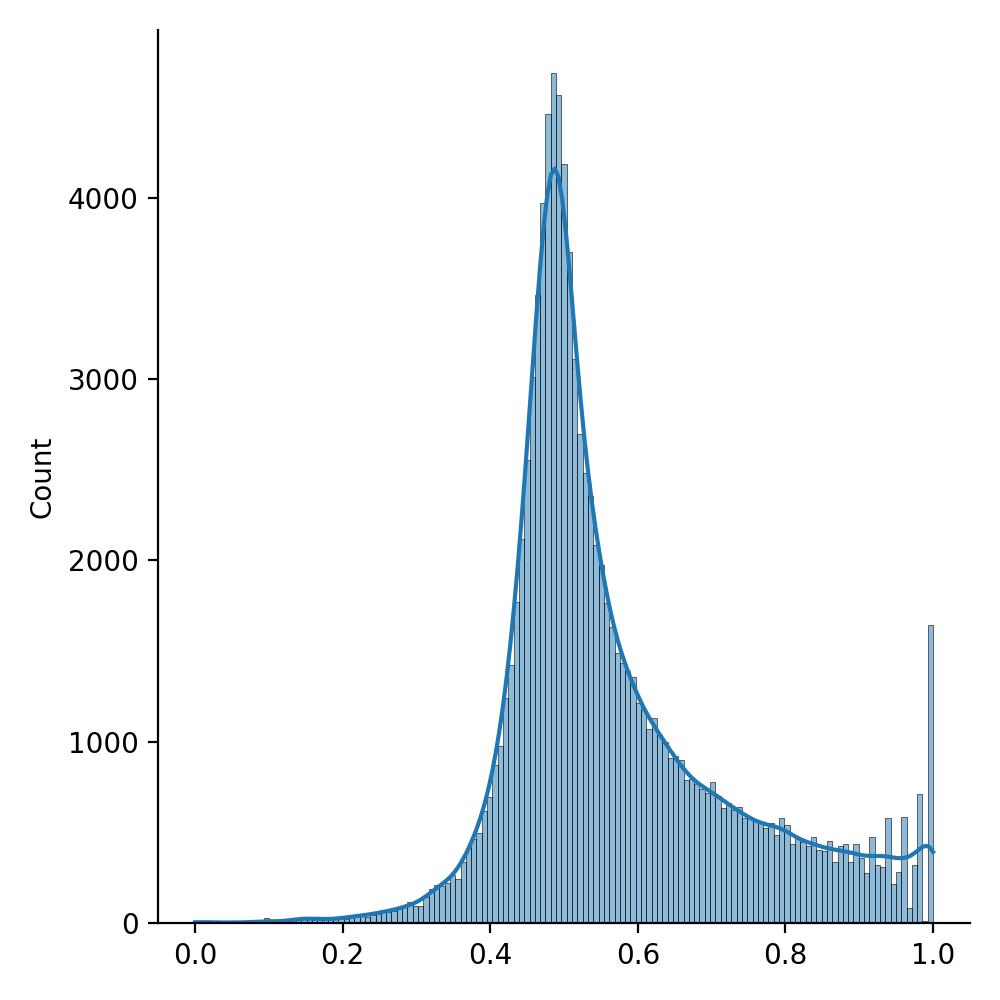}}
\hspace*{\fill}
\subfloat[The distribution of the credibility scores on DocRED.]{\label{fig:docred_ent}
\includegraphics[height=0.3\textheight]{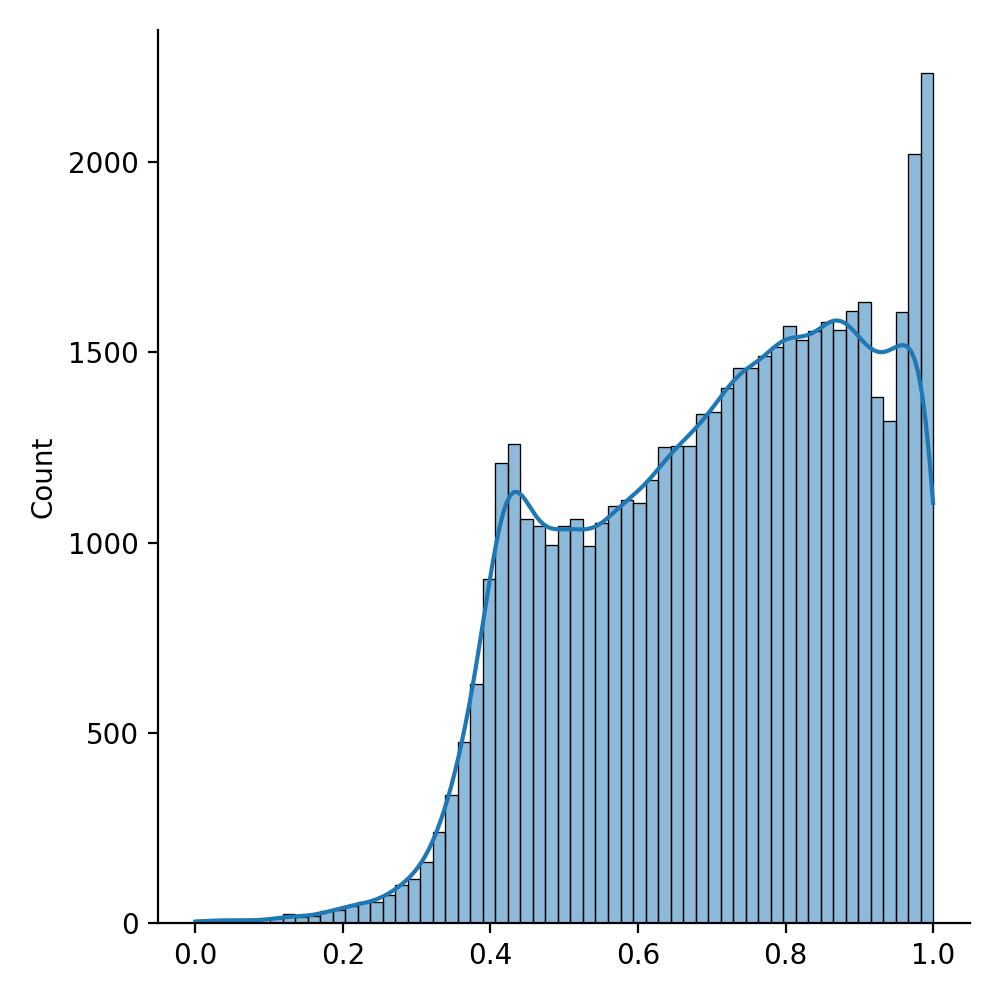}}
\caption{The long tails of the distributions of the credibility scores on the entire TACRED and DocRED datasets are apparent when the score is under certain threshold. }
\label{fig:ent_dist}
\end{figure}
In Figure \ref{fig:ent_dist}, it is clear that the examples with credibility scores under certain threshold form the long tail of the distribution. This phenomenon is adhered to the intuition due to the fact that the inconsistent annotations should never be in the majority of any sound corpus. The adaptive threshold for credibility scores will be left for future explorations.

\section{Discussions} 
\subsection{Trade-off of Relation Representations}
\label{sec:trade_off}
Mindset is a set of assumptions, methods, or notions held by people which are incentives to continually adapt the prior behaviours or choices \citep{argyris2004reasons, taylor1995effects}. As observed, the bias of mindset could happen to the PLMs when they encounter prompts with strong implications. It is especially fatal for the AID task because it relies on the neutral prior knowledge of the PLMs to reexamine the annotations with due impartiality. However, the strong indications in the prompts possibly misdirect the detectors to make arbitrary decisions on the ambiguous annotations.

\begin{figure}[h]
\subfloat[]{\label{fig:sent_max_emb}
\includegraphics[width=0.33\textwidth]{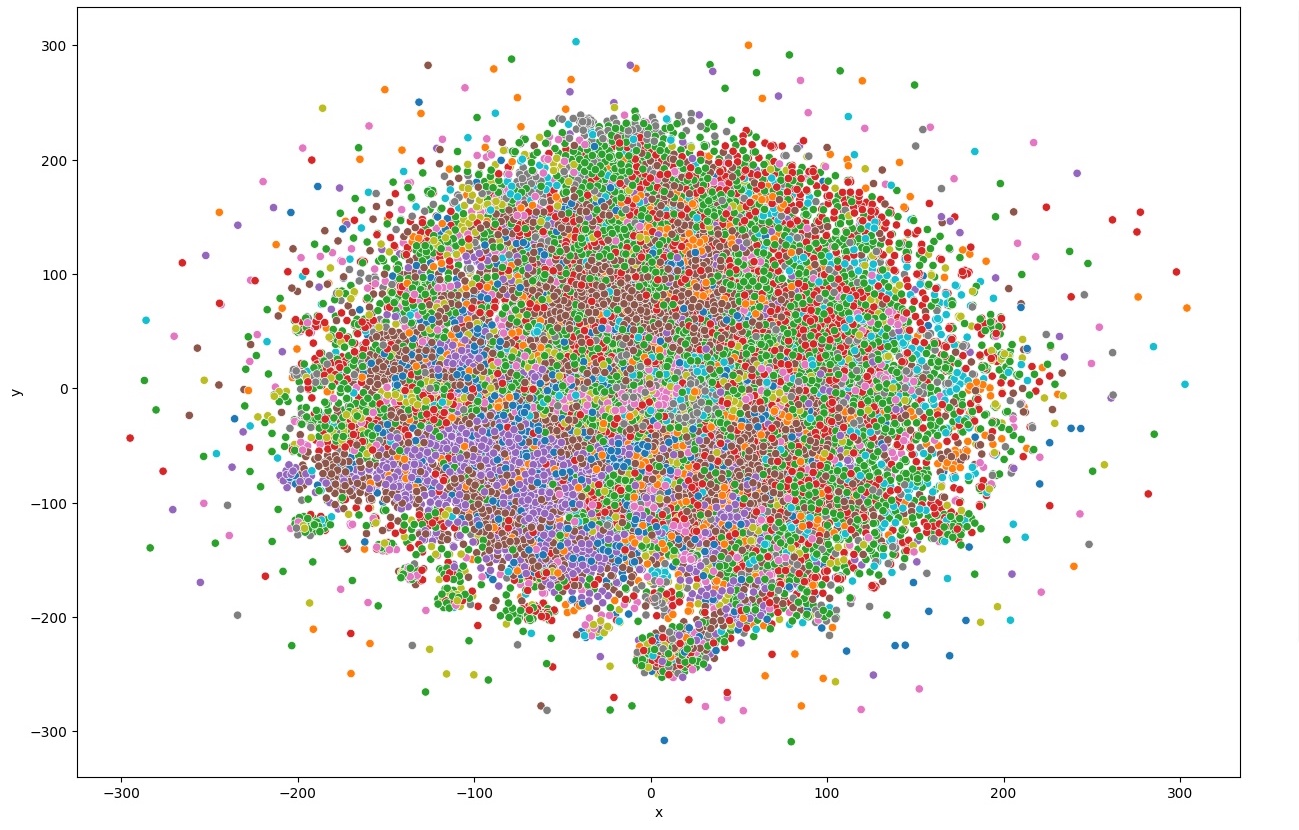}}
\hspace*{\fill}
\subfloat[]{\label{fig:mask_h_mask_t_emb}
\includegraphics[width=0.33\textwidth]{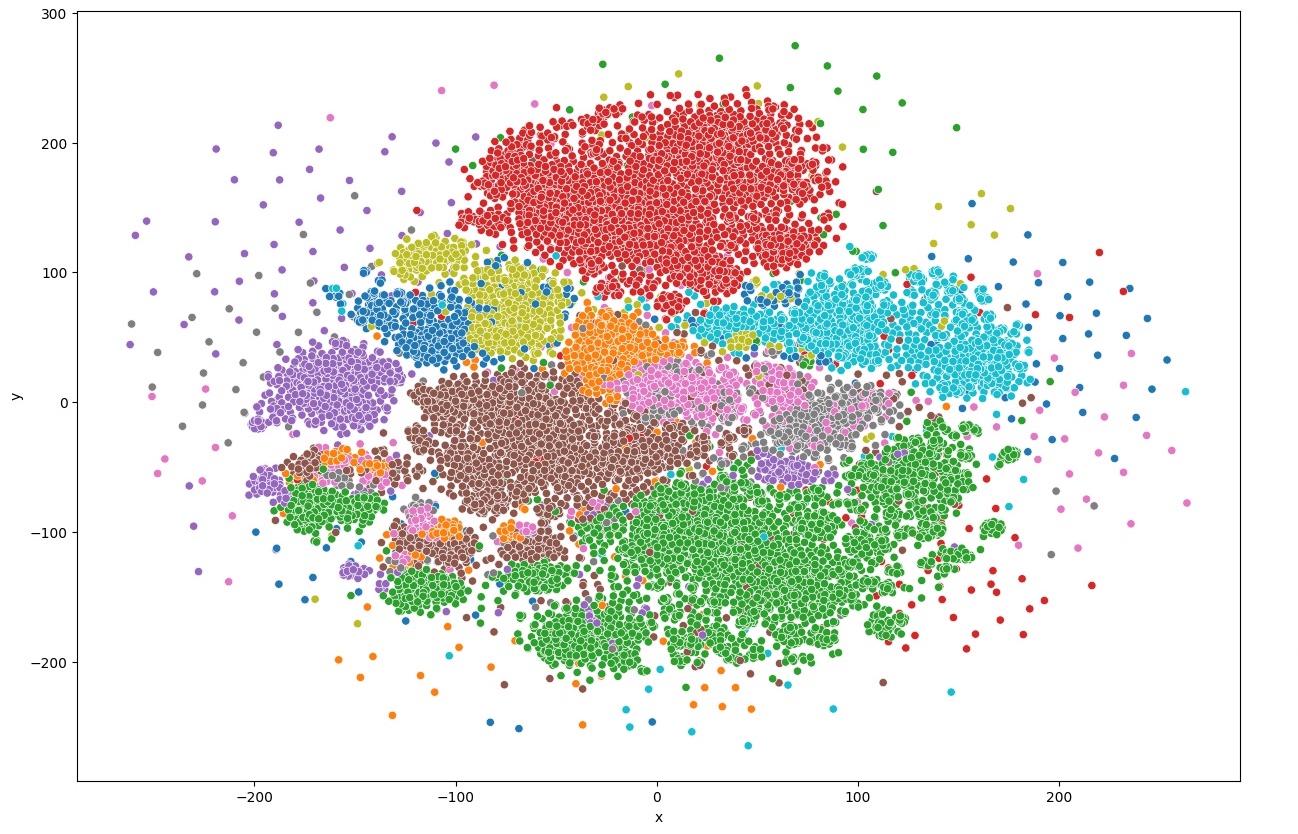}}
\hspace*{\fill}
\subfloat[]{\label{fig:mask_h_mask_t_emb}
\includegraphics[width=0.33\textwidth]{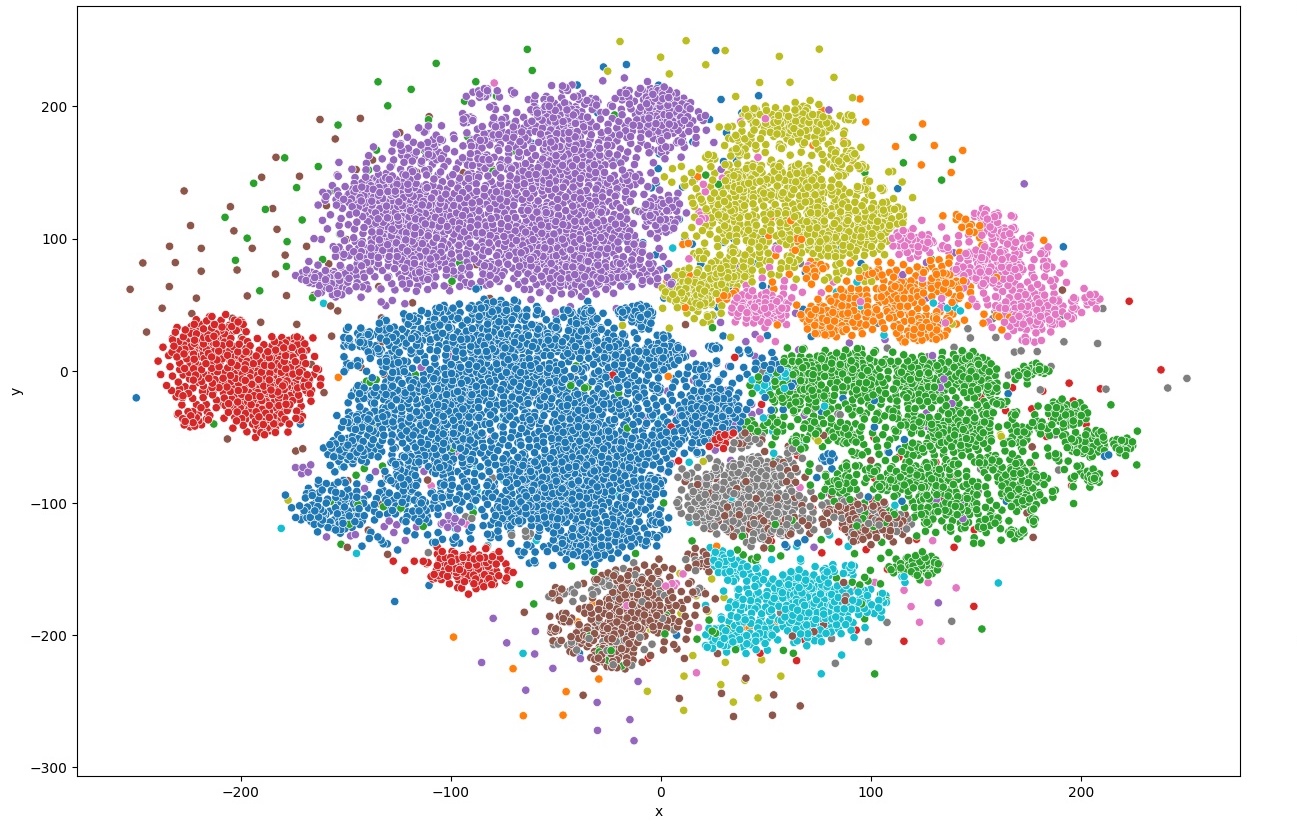}}
\caption{The T-SNE visualization of the relation representations by sentence-level and prompt-based embeddings, and the colors stands for the labels of examples. \textbf{(a):} The relation representations by sentence-level max pooling. \textbf{(b): }The relation representations by the prompts with \textsc{{[}HEAD{]} {[}MASK{]} {[}TAIL{]}} template. \textbf{(c):} The relation representations by prompts with human-tuned template. }
\label{fig:emb_vis}
\end{figure}

We leverage the T-SNE \citep{JMLR:v9:vandermaaten08a} to visualize the relation embedding acquired by different methods we mentioned in Section \ref{sec:prompt}. As shown in Figure \ref{fig:emb_vis}, the cluster of the embedding belonging to different relation types become more differential with the growth of the prior knowledge in prompt designs. Since the sentence-level methods do not provide any additional knowledge about neither the relation extraction task nor the AID task, its final relation representations almost have not shown any distinguishable cluster. In the sharp contrast, the relation representations obtained by \textsc{{[}HEAD{]} {[}MASK{]} {[}TAIL{]}} and human-tuned template form differential clusters according to relation types. Compared with the clusters by \textsc{{[}HEAD{]} {[}MASK{]} {[}TAIL{]}} templates with a slight overlap between adjoining classes, the clusters by human-tuned templates are separated by clearer boundaries. Counter-intuitively, the AID task actually requires the embedding distribution to meet a subtle balance between distinctive and farraginous. The strong inclination of clustering usually means the detectors would be overconfident with the observed annotations. 

\begin{figure}[h!]
\includegraphics[width=\linewidth]{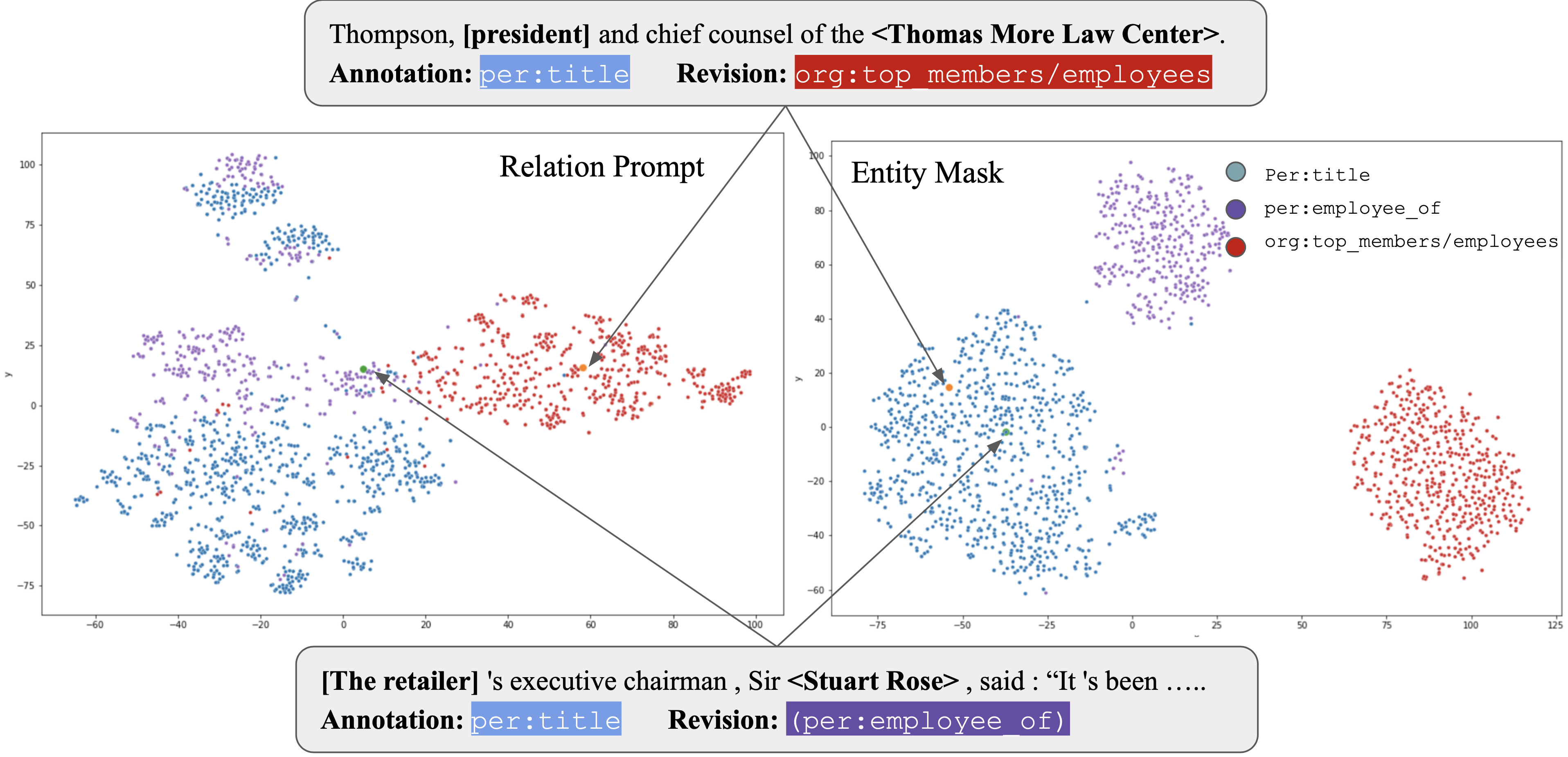}
\caption{The T-SNE visualization of two examples with the same incorrect annotation of \texttt{per:title} that should be revised into \texttt{org:top\_members/employees} and \texttt{org:employee\_of} respectively.}
\label{fig:emb_error}
\end{figure}
The Figure \ref{fig:emb_error} illustrates a typical detection error caused by the strong implication in prompts. There are more obvious three clusters, \texttt{per:title}, \texttt{org:top\_members/employees} and \texttt{org:employee\_of} clusters, of the relation representations by \textsc{entity mask} than \textsc{relation prompt}. Nevertheless, two examples with the same error annotation \texttt{per:title} are still located in the inappropriate \texttt{per:title} cluster  by \textsc{entity mask}, but they are mapped into the correct \texttt{org:top\_members/employees} and \texttt{org:employee\_of} clusters by \textsc{relation prompt}.

Prompt tuning, as one of the emerging paradigm in NLP, doubtless unleashes the potential of massive pre-trained models \citep{DBLP:conf/acl/GaoFC20, DBLP:conf/acl/LiL20, DBLP:conf/naacl/ZhongFC21}. However, our experiments remind that the over artefact prompts may easily lead to the dilemma in the era of feature engineering: the trade-off between generalizability and overfitting.

\chapter{Annotation Error Correction}
\label{chap:AEC}
Annotation Error Correction detects suspicious annotations and recommends true annotation for them simultaneously. Instead of zero-shot learning, we fine-tune the error corrector by cross-validation. We introduce the uncertainty to the observed hard label, effectively mitigating the negative impact of annotation noise during cross-validation. We also develop the rank-aware neighbouring encoder and distant-peer contrastive loss to enhance the neighbour awareness of AEC models. Empirically, distant-peer contrastive loss with uncertain soft labels is the optimal configuration of AEC models throughout our study. 

\section{Overview}
This chapter will demonstrate the Annotation Error Correction (AEC) based on the dynamic relation representations by fine-tuning the PLM-based neural relation classifier with cross-validation. The comprehensive definition of the AEC task and the formal notations are described in Section \ref{sec:aec_def}.

Compared to zero-shot learning, PLM-based neural relation classifiers fine-tuned by cross-validation enable AEC models to suggest better annotations for suspected examples precisely. However, as AEC models cross-validate on the noisy observed data, to alleviate the noise problems of softmax classifier, we convert the overconfident observed labels in training folds into the labels with uncertainty by Kernel Density Estimation (KDE) and K-Nearest Neighbours (KNN). We also enhance AEC models with neighbouring information based on the rank-aware Transformer encoder and a novel distant-peer contrastive loss. 

Empirically, we observe that fine-tuning leads to better AEC performance than zero-shot methods introduced in Chapter \ref{chap:AID}. Uncertain labels show better properties over the hard labels for fine-tuning AEC models on noisy observed datasets. The neighbouring awareness is also conducive to the AEC models generally. Consequently, the AEC model augmented with distant-peer contrastive loss and Kernel Density Estimation based uncertain labels outperforms other configurations, achieving macro $F_1$ of 66.2 on TACRev, 47.7 on Re-TACRED and 57.8 on Re-DocRED. 

Practically, cleaning the training set of relation extraction datasets with our proposed AEC framework leads to up to 3.6\%  downstream improvement for state-of-the-art relation extraction models. According to the statistics of needed time per examples, our proposed AID and AEC models beat the human revisers with significantly shortened re-annotation time. 
\section{Methodology}
\begin{itemize}
\item Section \ref{sec:cross_val} introduces the cross-validation for fine-tuning the AEC models on observed data. 
\item Section \ref{sec:uncertain_label} proposes the uncertain labeling for relieving the noise problem encountered during cross-validation. 
\item Section \ref{sec:classifier} suggests two methods for injecting the neighbouring information to the learning process of AEC models.
\end{itemize}

\subsection{Cross Validation}
\label{sec:cross_val}
Though well-designed relation embedders and neighbour-based detectors are proved to be effective in revealing the potential annotation inconsistencies, the static relation representations may not be sufficient to correct the invalid annotations. The credibility-based scores can indicate the consistency of a given annotation, but it is impossible to suggest another annotation. Though the KNN-based detectors are able to speculate the true labels based on the majority voting by retrieved neighbours, they are usually over-sensitive to the noisy neighbours. Especially for the AEC task, as there is no guarantee on the quality of training data, KNN-based classifiers can hardly predict the true label, if the example is surrounding by compromised neighbours.

In order to address the shortcoming of KNN for AEC, we attempt the AEC task by predicting the verified labels for the annotated example with the conditional probability:
\begin{equation}
r = \arg max_{r\in \mathcal{R}} \ Pr(r|r'(sub, obj), c, \mathcal{A}) \label{eq:aec}
\end{equation}, which means the AEC models suggest the relation that is the most compatible with both the observed annotation and the internal congruity over the entire dataset.

 Cross-validation \citep{stone1977asymptotic, tibshirani1996journal, doi:10.1080/00401706.1974.10489157} uses different portions of the data to test and train a model on different iterations, typically for estimating the model performance in practice or optimizing the hyper-parameters. In contrast, we leverage cross-validation to learn the conditional probability in Equation \ref{eq:aec} on target datasets with unchecked annotations.
 
 As for the Leave-one-out cross-validation, the entire dataset is usually evenly split into several folds. Then, in each iteration, one fold of data is sequentially selected to simulate the unseen data for hyper-parameter searching and testing, and other folds are merged together to train the models. Inspired by the cross-validation, we explore similar methods to train the PLM-based AEC models.
 
The neural relation classifier is built by the hidden layer and softmax classifier. The relation embedding vector $\mathbf{e}_r \in \mathbb{R}^d$ acquired by the prompt-based methods (Section \ref{sec:prompt}) or the entity-based methods (Section \ref{sec:marker}), is first fed into the hidden layer with the $ReLU$ non-linear activation:
\begin{gather}
\mathbf{h} = ReLU(\mathbf{W}_{proj}\mathbf{e}_r) \label{eq:hid_layer}
\end{gather}, where $\mathbf{W}_{proj} \in \mathbb{R}^{d\times d}$ is the linear projection matrix and $\mathbf{h}$ represents the hidden states from the hidden layer. Then, based on the hidden representation $\mathbf{h}$, the softmax classifier predicts the conditional probability of relation $r$ given context $c$ and observed annotation $r'$:
\begin{equation}
Pr(r \mid c) = \frac{exp(\mathbf{W}_r\mathbf{h} + b_r)}{\sum_{r' \in \mathcal{R}}exp(\mathbf{W}_{r'}'\mathbf{h} + b_{r'}')} \label{eq:softmax}
\end{equation}, where $\mathbf{W}_r, \mathbf{W'}_{r'} \in \mathbb{R}^{d \times R}$ and $b_r, b'_{r'} \in \mathbb{R}^{R}$. After training with cross-validation algorithm, the neural classifiers are expected to learn the relation with the maximum probability as the predicted rectified annotation.
 
\begin{figure}[h]
\includegraphics[width=\linewidth]{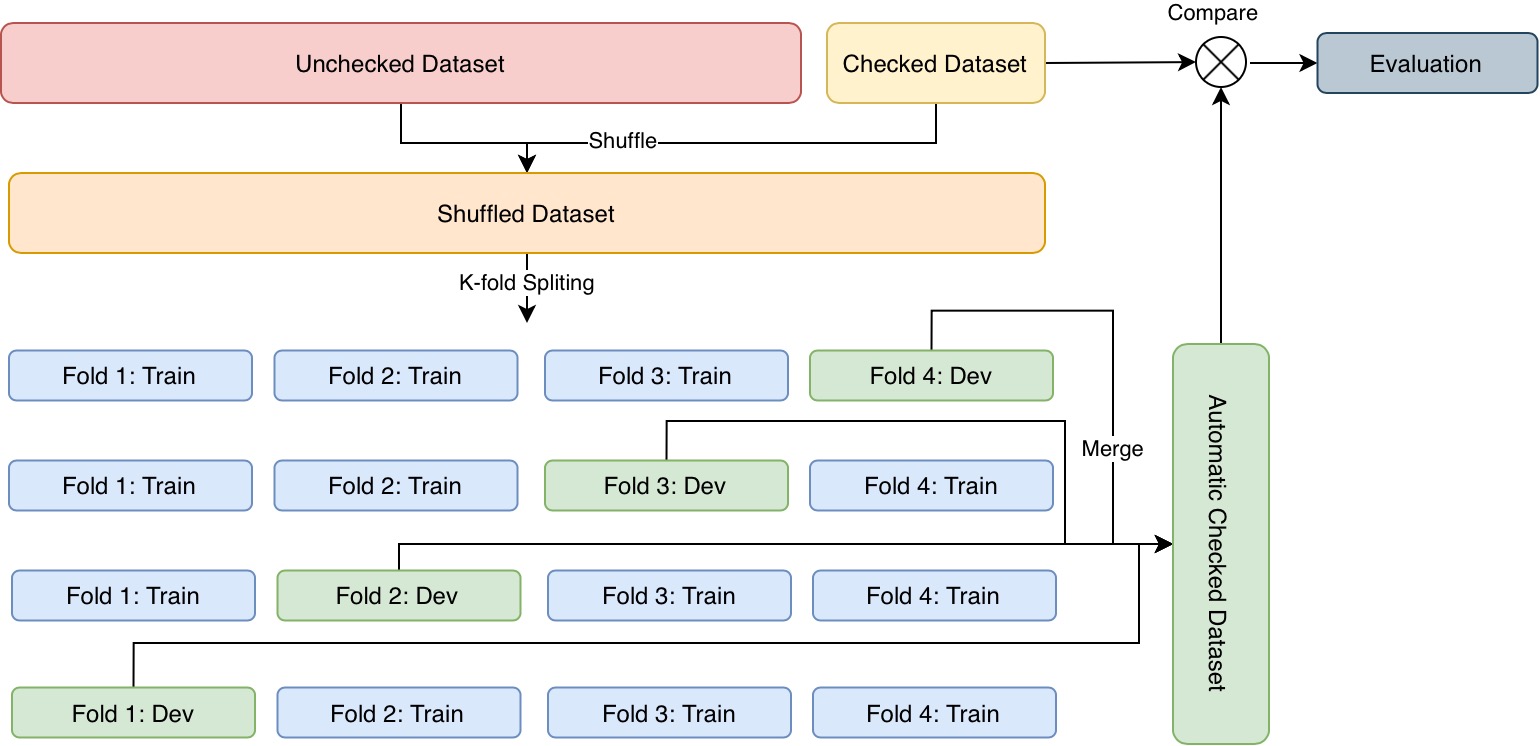}
\caption{The Pre-trained Language Models are fine-tuned by cross validation on the unchecked dataset. The automatically corrected annotations are obtained by merging the held-out Dev sets from all iterations of the cross validation.}
\label{fig:cross_val}
\end{figure}

Under the AEC setting, we train the models with the targets of unchecked annotations on training folds but expect the models to predict the labels on the held-out fold as the same as the human revised label (Figure \ref{fig:cross_val}). Statistically, the annotations on training fold $\mathcal{A_{\text{Train folds}}}$ could estimate the characteristics of the overall annotations $\mathcal{A}$ in Equation \ref{eq:aec} by sampling. Since the training process totally relies on the original annotations in the datasets without being exposed to any kind of revisions, there is no data leakage in the cross-validation.

As for the AID task, the PLMs already show impressive ability to detect abnormal annotations solely relying on their extensive inner common-sense knowledge learnt by pre-training on the massive general domain corpus. However, the cross-validation could further reinforce this strength to accommodate prior knowledge in PLMs to AEC task in three ways: (1) Fine-tuning endows the PLMs with the both task- and domain-specific knowledge in Relation Extraction; (2) Fine-tuning allows \texttt{[MASK]} tokens in prompts and new special tokens in entity-based methods (e.g. \texttt{[H]} in \textsc{Entity Marker})  to be adapted for relation verification.; (3) neural relation classifiers give better predictions based on the supervised learning, instead of only relying on the error-prone neighbouring information.

\subsection{Uncertain Labeling}
\label{sec:uncertain_label}
Although the softmax classifier empirically leads to excellent classification performance, it is also not absolutely robust to noise. Thus, we further introduce uncertainty to labels to avoid cross-validation based AEC model from over trusting the observed labels during training on the target datasets. We first present the mathematical explanation of noise sensitivity of softmax classifier by \cite{DBLP:journals/corr/abs-2001-01987}, and then describe two possible solutions to reduce the inducement of noise based on their theoretical insights.

According to \cite{DBLP:journals/corr/abs-2001-01987}, noise sensitivity of softmax classifier can be mathematically illustrated with Lipschitz Continuity and the relation between softmax classifier and $k$-means Clustering. First, the Equation \ref{eq:hid_layer} and Equation \ref{eq:softmax} of neural classifier could be expressed by a more general form: 
\begin{equation}
F(x) =  \sigma(f_p(x)^\intercal W) \label{eq:softgen}
\end{equation}, where $\sigma$ stands for softmax function and $W \in \mathbb{R}^{d\times c}$ is the matrix of weights. The $f_p$ denotes the penultimate layer of neural network that maps the $n$-dimensional input space to $c$-dimensional probability vector, and $c$ should be the same as the number of valid classes. This formulation omits the expression of bias vector and affine function.

Lipschitz Continuity \citep{NEURIPS2018_48584348} can theoretically measure the robustness of models by demonstrating the effect of the perturbations of the input. If the function $f: \mathbb{R}^n \rightarrow \mathbb{R}^c$ is Lipschitz continuous with modulus L, for every $x_1, x_2 \in \mathbb{R}^n$ it should satisfy:
\begin{equation*}
\|f(x_1) - f(x_2) \| \leq L\|x_1 - x_2 \|
\end{equation*}  The Lipschitz modulus of function $F$ in Equation \ref{eq:softgen} is decided by $L_p\|W\|$, where $L_p$ is the modulus of function $f_p$ because the modulus of softmax function is less than one:
\begin{equation*}
\| F(x_1) - F(x_2) \|^2 \leq \|f_p(x_1)^\intercal W - f_p(x_2)^\intercal W\| \leq L_p \|W\|\|x_1 - x_2 \|
\end{equation*} As for the neural networks, the small Lipschitz modulus, which implies that the adjacent data points have close function values, could also indicate the model robustness by restricting the effect on the classification of small distortions of data points with inequalities.

\cite{DBLP:journals/corr/abs-2001-01987} prove two theorems revealing the connections between softmax classifier and  $k$-means Clustering:

\textbf{Theorem 1}: \textit{Let the dimension of the penultimate layer $d$ be larger than or equal to the number of classes: $d \geq c - 1$. Assumed a network output is $y = \arg max_k f_p(x)^\intercal W_{\cdot k}$, there exist $c$ class centroids $Z_{\cdot k} \in \mathbb{R^d}$ with equal distance to the origin, such that every $x$ is classified to the class whose center is nearest in the transformed space:}
\begin{equation*}
y = \arg min_k \|f_p(x) - Z_{\cdot k}\|^2
\end{equation*}

\textbf{Theorem 2}: \textit{Let $Z$ be the center matrix in Theorem 1 and $x \in \mathbb{R}^n$ be a data point with predicted class $k$. Assumed that $f_p$ is Lipschitz continuous with modulus $L_p$, any distortion $\Delta \in \mathbb{R}^n$ which changes the prediction of point $\tilde{x} = x + \Delta$ to another class $l \neq k$ has a minimum value of:}
\begin{equation*}
\|\Delta\| \geq 
\frac{\|Z_{\cdot l} - Z_{\cdot k} \| - \|f_p(\tilde{x}) - Z_{\cdot l} \| - \|f_p(x) - Z_{\cdot k} \|}
{L_p}
\end{equation*}
These two theorems exemplify that the noise sensitivity may be relieved from three aspects: (1) reduce the Lipschitz modulus of neural networks, (2) maximize the mutual distance of the 
centroids of different classes, and (3) map $x$ or $\tilde{x}$ close enough to their class centroids.

\begin{figure}[h]
\includegraphics[width=\linewidth]{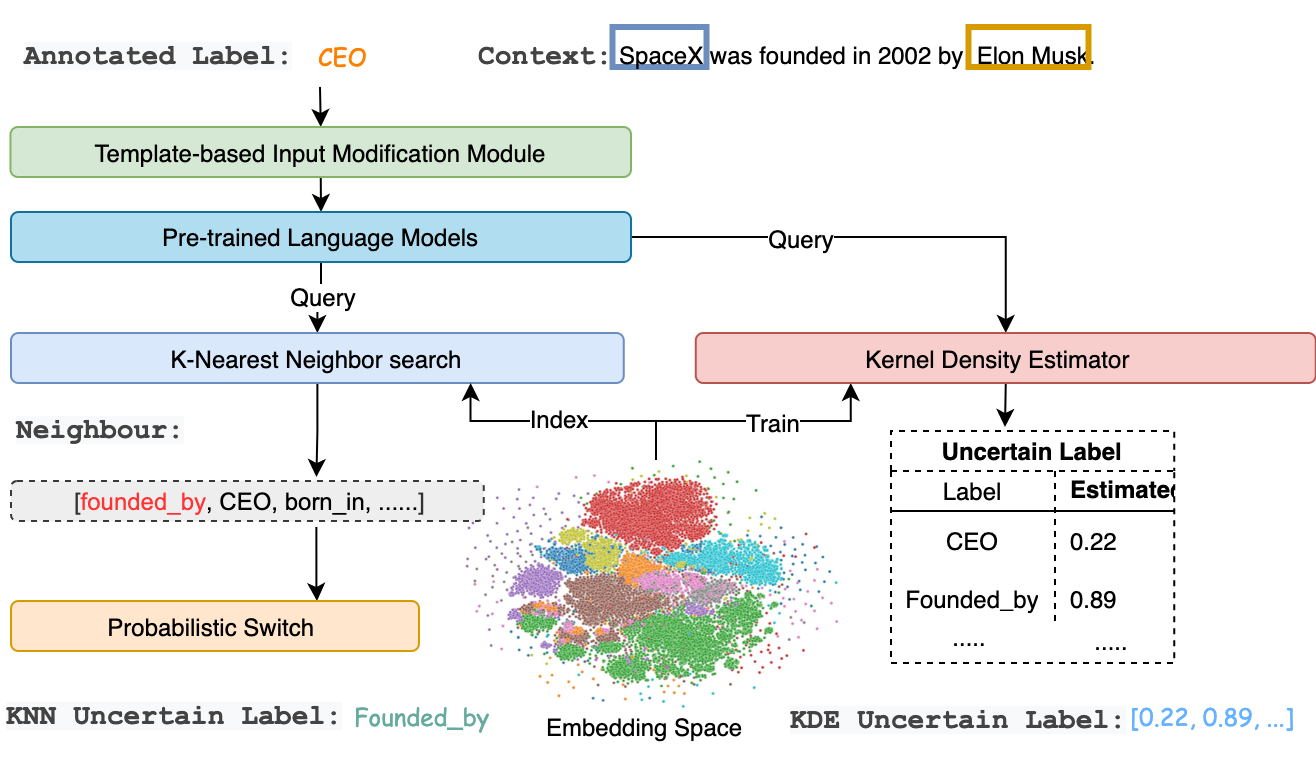}
\caption{The pipeline that converts the certain labels in the datasets into the labels with uncertainty by K-Nearest Neighbours and Kernel Density Estimation.}
\label{fig:uncertain}
\end{figure}

Considering the annotation errors are widely spread in the datasets and the noise sensitivity of the softmax classifier, we follow the second and third insights to prevent AEC models from unsuspectingly depending on the observed labels in the training folds. Following the concept of reappraising existing labels of an example based on the annotation tendency of its neighbours (Section \ref{sec:neigh}), we propose two approaches that intentionally introduce the uncertainty to the observed labels in datasets to enable trained models to amend the annotations on the held-out folds prudently (Figure \ref{fig:uncertain}): (1) let some observed labels adopt the majority labels of their neighbours to map the marginalised $x$ close to proximal class centroids, which may not be their original class centroids, by alternating their class membership, and (2) soft-label with Kernel Density Estimation for maximising mutual distance of the centroids of different classes by assigning probabilistic class membership.

\subsubsection*{Label Replacement by K-Nearest Neighbours}
As discussed in Section \ref{sec:neigh}, retrieved neighbours presumably hint at the underlying true label of the query example. Therefore, letting a suitable fraction of the labels adopt the true labels predicted by the KNN classifier may introduce the right amount of helpful uncertainty to the Train folds, because some marginalised $x$ after alternating class membership can be considered less noisy by softmax classifiers.

Given the observed relation label $r'$ and neighbouring relations $\{r_{n_0}, r_{n_1}...\}$, the uncertain label $\bar{r}$ are controlled by the probabilistic switch with a manually appointed threshold $\phi \in [0, 1]$ as follow:
\begin{gather}
\bar{r} = 
\begin{cases}
    KNN(\{r_{n_0}, r_{n_1}...\}), &  p_u  < \phi\\
    r', &     p_u \geq \phi
\end{cases}
\label{eq:knn_label}
\end{gather}, where $KNN$ is the K-Nearest Neighbours classifier, and $p_u$ is a random value following the uniform distribution in range $[0, 1]$. Hence, the threshold $\phi$ decides the ratio of label replacement in the whole datasets. 

\subsubsection*{Soft-label with Kernel Density Estimation}
 
Soft labels are widely accepted as the paradigm for handling noisy data in supervised learning \citep{DBLP:conf/kes/Thiel08, DBLP:journals/jamia/NguyenVH14, DBLP:conf/emnlp/LiuWCS17, DBLP:journals/nn/ZhaoZCL14, DBLP:journals/corr/abs-2103-10869}. In contrast to hard labels where class membership is certain, soft labels express uncertainty in which single label should be assigned. \citep{DBLP:conf/ilp/GalstyanC07}. Combined with the probabilistic cross-entropy loss, soft-label enables the model to consider the supervised signals from multiple possible classes.
 
Intuitively, we could use the similar idea of soft-label to prevent the AEC models from blindly trusting the annotation in the Train folds. According to Section \ref{sec:neigh}, given the embedded example and the relation type $t \in \mathcal{R}$, the probability density estimated by Equation \ref{eq:kde} can be regarded as the likelihood that the example indeed belongs to this relation type $t$. Therefore, the KDE model with appropriate bandwidth $h$ could convert the one-hot hard label $r'$ on the unchecked training folds into a soft label vector $\tilde{\textbf{r}} \in \mathbb{R}^{|\mathcal{R}|}$ with its estimated probability density regarding each relation class:
\begin{equation}
\tilde{\textbf{r}} = \{f_\mathcal{K}^{t_0}(\textbf{e}_r), ..., f_\mathcal{K}^{t_n}(\textbf{e}_r) \} , t_i \in \mathcal{R} \label{eq:kde_label}
\end{equation}, where $\textbf{e}_r$ is the relation embedding, $f_\mathcal{K}^{t}$ is the KDE models with the Gaussian kernel regarding relation type $t$, and $\mathcal{R}$ is the set of relation types.

\subsection{Neighbour-aware Correction}
\label{sec:classifier}
The neural classifiers are built by the fully-connected feed-forward layers followed by a softmax layer. It is able to better predict annotation inconsistencies by supervised learning on the entire training sets, in contrast to the KNN classifiers which only rely on the local geometry of the distribution of annotations. The empirical experiments in AID tasks indicate that static relation representations derived with properly crafted prompts or markers from the off-shelf PLMs are semantically sensitive to annotation noise. Fine-tuning the PLMs and neural classifier would further enhance this preponderance by subtly adapting the static relation representation to the relation extraction task. Hence, the neural classifier stacked on top of the PLMs trained with cross-entropy loss by the cross-validation is a strong baseline for the AEC task.

However, a vanilla neural relation classifier is not immediately aware of neighbouring relations when making its decisions. Such information would be helpful since the neighbouring consistency hints the essence of observed annotations as discussed in Chapter \ref{chap:AID}. We therefore propose to augment the vanilla neural relation classifier into neighbour-aware classifiers with two alternatives: (1) rank-aware neighbouring encoders or (2) contrastive learning with distant-peer positive examples.

\subsubsection*{Rank-aware Neighbouring Encoder}
\begin{figure}[h]
\includegraphics[width=\linewidth]{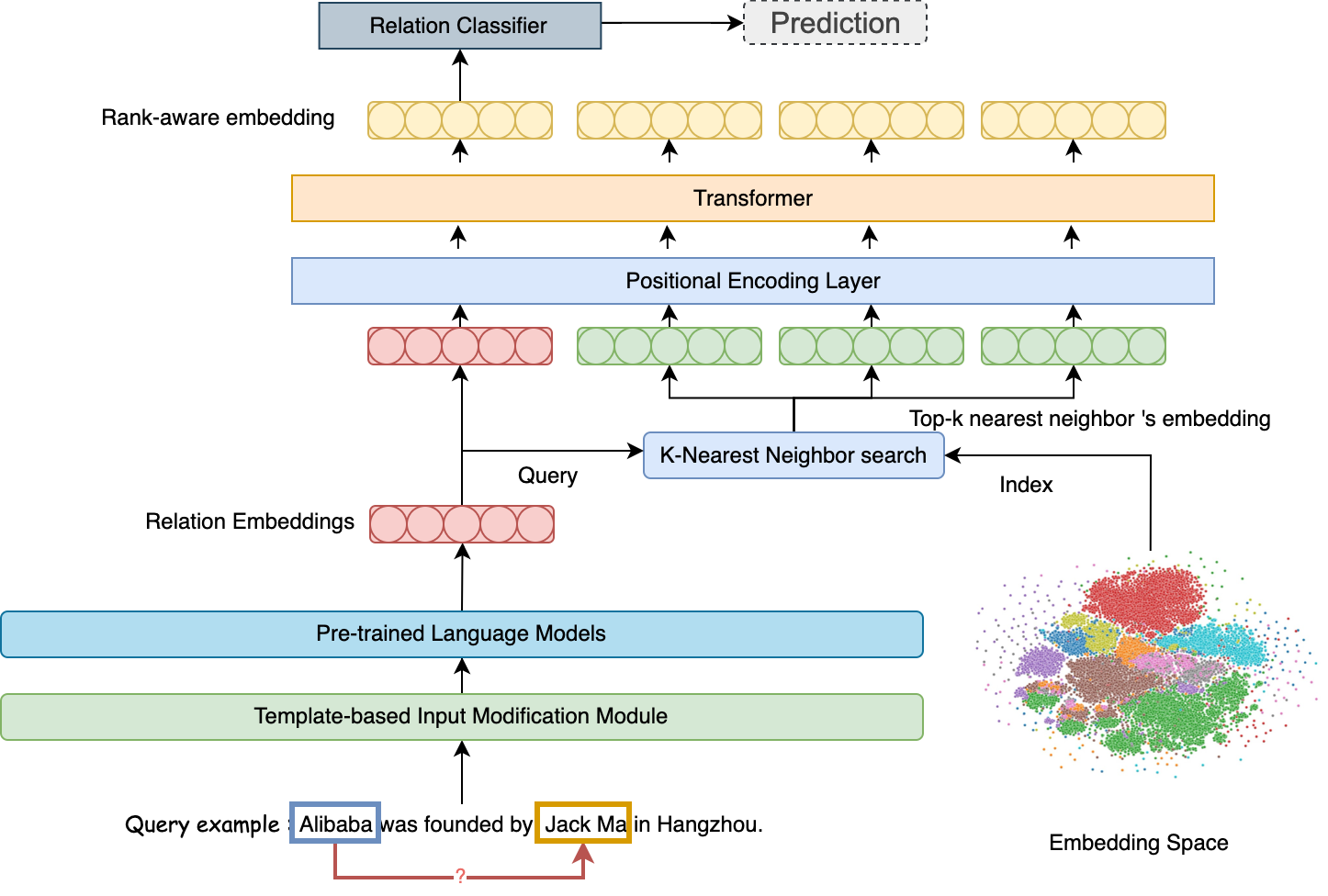}
\caption{The architecture of rank-aware neighbouring encoder.}
\label{fig:knn_trans}
\end{figure}

The KNN classifier treats each retrieved neighbour as an atomic individual but ignores their crucial mutual interactions. Therefore, we regard the query example and retrieved neighbours analogously as the sequential textual inputs and acquire their contextualized embedding with the rank-aware Transformer encoder \citep{DBLP:conf/nips/VaswaniSPUJGKP17} (Figure \ref{fig:knn_trans}). The inputs to the Transformer encoder is the sequence of relation representations of both the query example and its neighbours in rank order. The self-attention multi-head encoder acquires the neighbour-aware relation embedding of the query example by jointly aggregating the information from its neighbours and capturing the mutual interaction among neighbours. Meanwhile, the positional encoding utilizes the hints behind the order of its neighbours.

The rank-aware neighbouring encoder intends to replace the vanilla embedding vector $\mathbf{e}_r$ in Equation \ref{eq:hid_layer} with the neighbour-aware embedding vector $\mathbf{e'}_r$. Each relation representation $\textbf{e}_r$ obtained from PLMs will be first merged with the embeddings of its neighbours $\{\textbf{e}_{n_0}, ..., \textbf{e}_{n_i}\}$ into the sequence $\mathbf{z} = \{\textbf{e}_r, \textbf{e}_{n_0}, ..., \textbf{e}_{n_i}\}$ in order. As the transformers are neither recurrent nor convolutional, the model architecture restricts its capability of encoding the sequential information in the input. Hence, augmenting the input embeddings with the positional information is crucial for rank-aware learning. We follow the techniques proposed by \cite{DBLP:conf/nips/VaswaniSPUJGKP17}, to get the positional encoding $\textbf{g}_i$ of each input position $i$ as follows:
\begin{gather*}
\omega_k = \frac{1}{10000^{2k/d}} \\
\textbf{g}_i = f(i) = 
\begin{cases}
    sin(\omega_k, i), \text{if i = 2k} \\
    cos(\omega_k, i), \text{if i = 2k +1}
\end{cases}
\end{gather*}, where d is the encoding dimension. Then, the sequential neighbouring matrix $\mathbf{z}$ is added with the position encoding $\mathbf{g} = \{\textbf{g}_0, ..., \textbf{g}_i\} $ into the rank-aware input $\mathbf{x}$. Then, we use the Transformer encoder with a multi-head attention layer and a fully-connected feed-forward layer with ReLU activation to acquire the contextualized encoding of the neighbouring sequence:
\begin{gather*}
\mathbf{h} = MultiHead(\mathbf{x}) \\
\mathbf{y} = FeedForward(\mathbf{h})
\end{gather*}We take the first contextual embedding which is in the position of the query example from the output encoding $\mathbf{y}$ as the neighbour-aware embedding vector $\mathbf{e'}_r$:
\begin{equation}
\textbf{e}'_r = \mathbf{y} _0
\end{equation}Consequentially, based on the rank-aware neighbouring vector $\mathbf{e'}_r$, we augment the neural relation classifier discussed in Section \ref{sec:classifier} with neighbouring attention.

\subsubsection*{Contrastive Learning with Distant-Peer }
Cross-entropy loss, as the most popular loss function for neural classification models in supervised learning, has been long criticized for inducing poor decision margins \citep{DBLP:conf/nips/ElsayedKMRB18, DBLP:conf/icml/LiuWYY16} and lacking robustness to noisy annotations \citep{DBLP:conf/nips/ZhangS18, sukhbaatar2015training}. These two shortcomings are especially fatal for AEC tasks because the target of the neural relation classifier is to learn the correct annotations from the noisy observed annotations. The supervised contrastive loss \citep{DBLP:conf/nips/KhoslaTWSTIMLK20} is an appealing supplement of cross-entropy loss to enhance the noise-tolerant learning for the AEC task. 

Contrastive learning has shown signs of resurgence by achieving competitive performance in unsupervised learning in various machine learning domains \citep{DBLP:conf/cvpr/WuXYL18, DBLP:conf/icml/Henaff20, DBLP:journals/corr/abs-1807-03748, DBLP:conf/iclr/HjelmFLGBTB19, DBLP:conf/icml/ChenK0H20, DBLP:conf/cvpr/He0WXG20, DBLP:conf/cvpr/StojnicR21, DBLP:journals/access/Le-KhacHS20}. The intuition behind contrastive learning is to pull together an anchor and positive examples meanwhile push apart the anchor from negative examples, where the anchor is the target example. The positive examples are the samples that share the fundamental homogeneity with the anchor example, while the negative examples should be essentially different from the anchor example. Under the unsupervised setting, the positive examples are usually obtained by the data augmentation of the sample, and negative examples are randomly selected from the training batch. Supervised contrastive loss \citep{DBLP:conf/nips/KhoslaTWSTIMLK20} generalizes this idea by introducing the supervising signal to the contrastive process to exploit the label information fully. It pulls closer the normalized embedding from the same class than the embeddings from different classes by taking both the self-augmented examples and the examples with the same label as the positive targets. According to the experimental results presented in \cite{DBLP:conf/nips/KhoslaTWSTIMLK20}, supervised contrastive loss empirically endows the deep neural networks with better robustness and hyperparameter stability in the context of image corruption and training data reduction regarding Computer Vision (CV).

Contrastive learning also shows its potential advantages in several Natural Language Processing (NLP) tasks, such as text generation \citep{DBLP:conf/iclr/LeeLH21} and information extraction \citep{DBLP:conf/aaai/YeZDCTHC21, DBLP:conf/emnlp/PengGHLLLSZ20}. However, controllable self-augmentation is usually harder in most NLP tasks than CV because of the flexibility of natural languages. Therefore, in this project, we merely regard the examples with the same relation label in the batch as positive while taking the example with the different relation label as negative. First, a batch of relation representations $\mathbf{e}_r = \{\textbf{e}_i\}_{i=1}^I$ , where $\mathbf{e}_i \in \mathbb{R}^d$, obtained by prompt-based or entity-based approaches are mapped to the dimension reduced projection space $p$ with multi-layer perceptron $f_{proj}$:
\begin{equation}
\mathbf{z} = f_{proj}(\mathbf{e}_r)
\end{equation}The vanilla contrastive loss $\mathcal{L}_{cl}$ for the AEC task is:
\begin{equation}
\mathcal{L}_{cl} = \sum_{i \in I} \frac{-1}{|P(i)|}\sum_{p \in P(i)} 
log \frac{exp(\textbf{z}_i \cdot \textbf{z}_p / \tau)}
{\sum_{n \in N(i)}exp(\textbf{z}_i \cdot \textbf{z}_n / \tau)}
\label{eq:cl}
\end{equation}, where $i$ is the index of anchor, $P(i)$ and $N(i)$ are the set of indices of all positives and negatives regarding the anchor example in the batch, respectively, $|P(i)|$ is the cardinality of the positive set, and $\tau \in \mathbb{R}^+$ is a scalar temperature hyperparameter. Ideally, the vanilla contrastive loss $\mathcal{L}_{cl}$ lets the relation embedding get closer to the examples from the same class while staying away from the examples from disparate classes. Nevertheless, as the instances in a batch are randomly sampled, the model may not fully leverage the most informative knowledge from the adjoining embeddings of the anchor examples. It is especially problematic when the computational resources limit the batch size, making the sampling less representative and leading to poor classification margins. Therefore, we alleviate this shortcoming by adding neighbouring information to the contrastive loss and involving the supervised signals from cross-entropy learning.

Inserting the positive neighbours to $P(i)$ or negative neighbours to $N(i)$ injects the neighbouring knowledge into the contrastive losses. However, the quality of inserted neighbours strongly impacts the final learning output. Furthermore, there is no guarantee that the observed labels of neighbours are correct. As argued by \citet{DBLP:conf/iclr/LeeLH21,DBLP:conf/nips/KhoslaTWSTIMLK20}, the gradient contributions from harder positives or negatives are more conducive to the encoder. 

Therefore, instead of randomly choosing the neighbours or inserting all neighbours, we propose a new method to introduce the distant-peer to the positive set for computing contrastive loss. The distant-peer is the farthest positive neighbour according to our defined peer distance $\lambda$. The peer distance evaluates if a neighbour is positive or not without considering its own observed labels, but the co-occurrence of its label among all neighbours and the distance from the anchor. Let $\mathbf{L} = \{\tilde{\textbf{r}}_i\}_{i = 0}^N$ be the soft-label matrix of neighbours where $\tilde{\textbf{r}}$ is the KDE-based soft-label of each neighbour computed by Equation \ref{eq:kde_label} and $N$ is the number of searched neighbours, and the distance vector of neighbours $\mathbf{D} = \{d_i\}_{i=0}^{N}$ denotes the distances between the anchor and each of its neighbours. Given $\mathbf{L} \in \mathbb{R}^{N \times |\mathcal{R}|}$ and $\mathbf{D} \in \mathbb{R}^{N}$, the peer distance $\mathbf{\lambda} \in \mathbb{R}^{N}$ is defined as:
\begin{equation}
\mathbf{\lambda} = \mathbf{L}\mathbf{L}^\intercal log(\mathbf{D})
\label{eq:peer_score}
\end{equation}, where $\mathbf{L}\mathbf{L}^\intercal \in \mathbb{R}^{N \times N} $ could be regarded as the co-occurrence of the annotations of neighbours. Multiplying the co-occurrence matrix $\mathbf{L}\mathbf{L}^\intercal$ with $ log(\mathbf{D})$ embodies our assumption that  the dense concentration of concurrent annotations in the distant neighbours may imply the existence of positives. Then, we add the top $k$ neighbours selected by the criterion of peer distance $\lambda$ to the positive set $P(i)$ in Equation \ref{eq:cl} to get the neighbour-aware contrastive loss $\mathcal{L}_{ncl}$.

\begin{figure}[h]
\includegraphics[width=\linewidth]{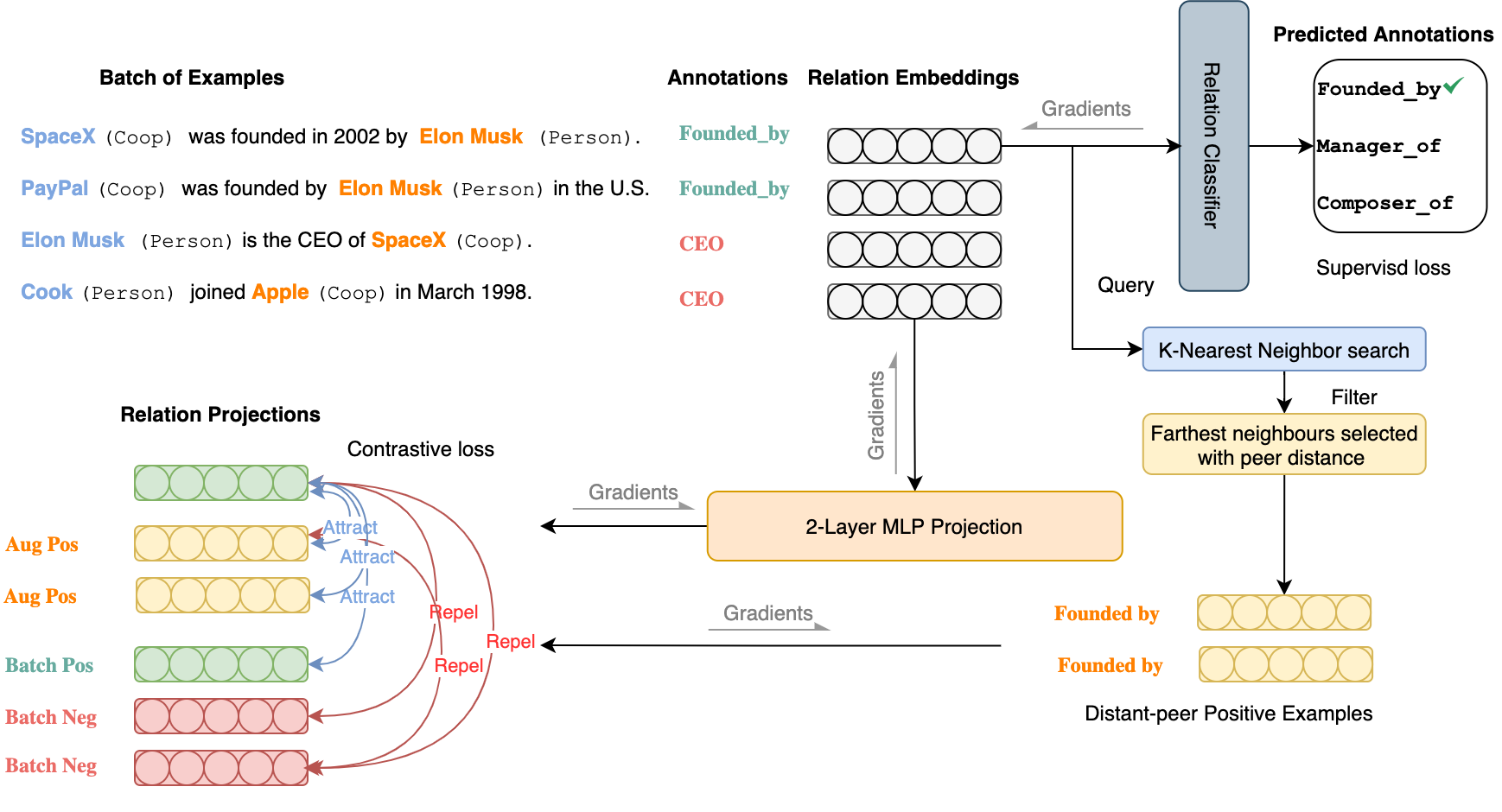}
\caption{The framework of contrastive learning with distant-peer.}
\label{fig:contrast}
\end{figure}

Finally, we apply the auxiliary loss technique of multi-task learning \citep{DBLP:journals/corr/ZhangY17aa, DBLP:journals/corr/abs-2007-01126} which combines the strengths of cross-entropy loss and contrastive loss, to incentivize AEC classifier to learn the semantically rich and highly differential relation representations for predicting the corrected labels (Figure \ref{fig:contrast}). The standard cross-entropy loss $\mathcal{L}_{ce}$ is computed with the ground-truth revisions and the predictions given by the vanilla softmax-based relation classifier described in Section \ref{sec:cross_val}, while the contrastive loss $\mathcal{L}_{ncl} $ is computed by the distant-peer contrastive learning. We leverage the weighted loss combination $\mathcal{L}$ to introduce the neighbouring knowledge to the neural relation classifier during training:
\begin{equation}
\mathcal{L} = \mathcal{L}_{ce} + \mu \mathcal{L}_{ncl} 
\label{eq:multi_loss}
\end{equation}, where $\mu$ is a hyper-parameter, the contrastive loss weight.
 
\section{Evaluation}
\label{sec:downstream}
We thoroughly evaluate the performance of AEC models with both upstream and downstream methods. Section \textit{Upstream Evaluations},  introduces the classification metrics, accuracy and macro $F_1$ for upstream evaluation while regarding human revisions as ground truths. Section \textit{Downstream Evaluations}, describes the relation extraction datasets and selected state-of-the-art models for downstream evaluating the de-noising outcomes of our proposed AEC models.

\subsection*{Upstream Evaluations}
In essence, the Annotation Error Correction task is the multi-class classification task that requires models to give the revising suggestions as to the same as the human revisers. Hence, naturally, we can take the human revisions in the TACRev and Re-TACRED datasets as the ground truth to evaluate the performance of proposed AEC systems with classification metrics, like accuracy and macro $F_1$ score \citep{grishman1996message}. Accuracy can give the general idea of how many predictions are exactly the same as human re-annotations, and $F_1$ score comprehensively reflect both the correctness and misclassifications. The macro $F_1$ score is the average of the independent $F_1$ scores of every class, which is especially conducive to AEC tasks because we expect the AEC models to have evenly performed in rectifying the annotations from all classes. The value of Micro $F_1$score may be largely increased if the AEC models are good at correcting the major classes, such as the \texttt{no\_relation} in TACRED, which is misleading for assessing the model performance of dealing with the annotation noise from minor classes.

\subsection*{Downstream Evaluations}
While the classification metrics are simple and straightforward approaches to quantifying the performance of the AEC models, they may not adequately represent the real-world performance of proposed systems directly. Hence, downstream evaluation with the state-of-the-art (SOTA) models in relation extraction may present more convincing evaluation results: (1) Using the optimized AEC systems to automatically denoise the original Train set of TACRED \citep{zhang2017tacred} and DocRED dataset \citep{yao-etal-2019-docred}; (2) Training the same SOTA relation extraction models on both the raw Train set and the denoised Train set; (3) Evaluating the models trained on different Train sets with the same Dev and Test sets on TACRED, TACRev \citep{DBLP:conf/acl/AltGH20a}, and DocRED \citep{yao-etal-2019-docred}. 

Table \ref{tab:downstream} shows that both TACRED and TACRev share the same size Train, Dev, and Test set with 68,124 examples, 22,631 examples, and 15,509 examples, respectively. While the Train sets of TACRED and TACRev are identical, TACRev has 1,656 examples in its Dev set and 998 examples in its Test set that are revised from the original TACRED Dev and Test examples. The authors of TACRev indicate that the evaluation quality is largely improved by these 7.3\% and 6.4\% label revisions on TACRED Dev and Test sets because the erroneous labels contribute up to 8\% test error. The best AEC model selected by the upstream classification metrics changes the annotations of 7,267 examples in the TACRED Train set, which accounts for 10.7\% of the Train examples. The DocRED datasets include 38,269 Train examples, 12,332 Dev examples, and 12,842 Test examples. The optimal AEC models automatically revise 7,676 examples on the Train set, accounting for 20.0\% Train instances. 

\begin{table}[h]
\centering
\resizebox{\textwidth}{!}{%
\begin{tabular}{lccccc}
\hline
Dataset & \multicolumn{1}{l}{\#Train} & \multicolumn{1}{l}{\#Dev} & \multicolumn{1}{l}{\#Test} & \multicolumn{1}{l}{\#AEC Revisions} & \multicolumn{1}{l}{\#RelTypes} \\ \hline
\multicolumn{6}{l}{\textit{Sentence-level Relation Extraction Datasets}} \\
TACRED \citep{zhang2017tacred}    & 68,124      & 22,631     & 15,509     & 7,267     & 42     \\
TACRev \citep{DBLP:conf/acl/AltGH20a}     & 68,124      & 22,631     & 15,509     & 7,267     & 42     \\ \hline
\multicolumn{6}{l}{\textit{Document-level Relation Extraction Datasets}}          \\
DocRED \citep{yao-etal-2019-docred}    & 38,269      & 12,332     & 12,842     & 7,676     & 96     \\ \hline
\end{tabular}%
}
\caption{The statistics of the original TACRED, DocRED datasets and the TACRev dataset partially revised from TACRED. Our proposed AEC models revised 10.7\% examples on the TACRED Train set, and 20.0\% examples on DocRED Train set.}
\label{tab:downstream}
\end{table}

To evaluate our proposed AEC models on downstream performance we first select the SOTA sentence-level relation extraction model by \cite{DBLP:journals/corr/abs-2102-01373} and the SOTA document-level ATLOP models \citep{WenxuanZhou2020DocumentLevelRE}. For instance, as for sentence-level models, we first train a model on the original Train set of TACRED and another model on an automatically denoised Train set of TACRED. Finally, by comparing the performance differences between these two trained models on the same original Test set of TACRED, we can testify whether correcting the training set by AEC models is conducive to the downstream tasks.

Details regarding these two investigated models are as follows:
\paragraph*{Entity Marker Model}
The entity marker model proposed by \citet{DBLP:journals/corr/abs-2102-01373}, has almost the same architecture as the vanilla PLM-based relation classifiers described in Sections \ref{sec:cross_val}, with the entity-based relation embedder applying the \textsc{entity marker} techniques demonstrated in Section \ref{sec:marker}. This simple but strong baseline offers the new SOTA performance in the sentence-level RE task. It even outperforms the competitive knowledge-enhanced PLM KnowBERT \citep{DBLP:conf/emnlp/PetersNLSJSS19} on TACRED.

\paragraph*{ATLOP model}
ATLOP, Adaptive Thresholding and Localized Context Pooling, model presented by \citet{WenxuanZhou2020DocumentLevelRE} tackles the document-level RE tasks with two techniques:\begin{itemize}
\item Adaptive-thresholding loss: Most RE models on DocRED apply a threshold to the output probability for deciding if a certain relation holds between given head and tail entities. But the threshold needs to be manually specified and so can potentially result in decision errors. Adaptive-thresholding enables the model to learn an adaptive threshold independently, depending on entity pairs.
\item  Localized context pooling: Normally, the entity representations for document-level RE tasks are acquired by aggregating the embeddings of all occurrences for a given entity over entire documents, whereas some context of the entities may not be relevant and may even distract from the RE targets. Hence, the localized context pooling reinforced the capability of capturing related context for entity pairs of PLMs by transferring pre-trained attention, which leads to better entity representations. 
\end{itemize}
The ATLOP model reaches an $F_1$ score of 63.4 on DocRED, surpassing other representative document-level RE models such as HIN-BERT-base \citep{DBLP:conf/pakdd/TangC0CFWY20} and CorefBERT-base \citep{ye-etal-2020-coreferential}.

\section{Experiments}
\begin{itemize}
\item Section \ref{sec:aec_implement} gives the detailed description of experimental implementations in AEC. 
\item Section \ref{exp:cross_val} contrasts the fine-tuned relation representations introduced in Section \ref{sec:cross_val} with the zero-shot relation representations previously discussed in Section \ref{sec:rel_rep_meth} in AEC through comprehensive experiments.
\item Section \ref{exp:uncertain_label} presents the results of different types of uncertain labels introduced in Section \ref{sec:uncertain_label}.
\item Section \ref{exp:neigh_class} demonstrates the performance of neighbour-aware AEC models discussed in Section \ref{sec:classifier}.
\item Section \ref{exp:cl_with_uncertain} shows the impressive performance gain with the combination of uncertain labels (Section \ref{sec:uncertain_label}) and distant-peer contrastive loss (Section \ref{sec:classifier}).
\item Section \ref{exp:downstream} reveals the downstream performance of our proposed AEC models following the descriptions in Section \ref{sec:downstream}.
\end{itemize}

\subsection{Implementation Details}
\label{sec:aec_implement}
The Pre-trained Language Model, long sequence handling techniques and experimental environments for the AEC task are the same as the AID task described in Section \ref{sec:aid_implement}. The reported experimental results on the Test sets in this section are regarding the models that reach the highest macro $F_1$ scores on corresponding Dev sets within the training epochs.

\paragraph*{Datasets}
The proposed uncertain labeling methods and neighbour-aware classifiers rely on neighbouring information by retrieving the neighbours for the query example. Identically to the development of the datastore used in AID task, the key of embeddings and value of labels are initially acquired based on the entire TACRED dataset with 103,738 examples and DocRED dataset with 50,503 examples. The upstream experiments are conducted with the TACRev, Re-TACRED and Re-DocRED datasets which contain manual revisions of annotations of partial examples in TACRED and DocRED. We take the revised labels as the gold labels to evaluate the performance of proposed AEC methods. The TACRev dataset contains 1,263 Dev examples and 1,263 Test examples, Re-TACRED contains 5,364 Dev examples and 5,365 Test examples, and Re-TACRED contains 206 Test examples and 205 Dev examples. As for the downstream evaluations, we leverage the initial TACRED and DocRED with the original data split depicted in Table \ref{tab:downstream} to get the comparable results with other existing research on relation extraction.

\paragraph*{Embedding of Neighbours}
The parameters of the PLM are fine-tuned along with the iterations of cross-validation. This means that the initial relation embedding acquired by the off-shelf PLM stored in the keybase for the KNN retriever may be out-of-date. Therefore, we investigate the influence of the static and dynamic neighbouring embeddings: 
\begin{itemize}
\item \textbf{Static Embedding of Neighbours}: The embeddings in the keybase always keep the same initialization from the off-shelf PLM.
\item \textbf{Dynamic Embedding of Neighbours}: The embeddings in the keybase are updated at the end of each epoch with the new embeddings acquired by latest fine-tuned models.
\end{itemize}
Empirically, we found that the rank-aware neighbouring encoders work better with static neighbour embeddings, whereas the contrastive learning approaches work better with the dynamic neighbour embeddings.

\paragraph*{Hyper-parameters of Cross Validation}
Both TACRED and DocRED are split into 4 folds for cross validation, and each iteration of cross validation has 5 training epochs.The optimizer for all types of neighbour-aware classifiers is AdamW \citep{DBLP:conf/iclr/LoshchilovH19}. The learning rate for fine-tuning the AEC models is 5e-4, the dropout probability is 0.2, and the warm-up ratio is 0.1 for both datasets.

\paragraph*{Hyper-parameters of Uncertain Labelling Models}
With the KNN-based label replacement methods, we only study the setting of $k=1$. With Bayesian optimization \citep{snoek2012practical} on the Dev sets, we optimize the replacement threshold $\phi$ of KNN-based methods in Equation \ref{eq:knn_label} as 0.3 for TACRED and 0.15 for DocRED, and the bandwidth $h$ for KDE-based soft labelling in Equation \ref{eq:kde_label} as 0.25 for TACRED and 0.1 for DocRED.

\paragraph*{Hyper-parameters of Neighbour-aware Classifiers}
\begin{itemize}
\item \textbf{Rank-aware neighbouring Encoder}: The backbone model is the Transformer encoder with 6 layers where each attention layer has 8 attention heads. The dropout probability of each layer is 0.1 as default. The number of neighbours retrieved for providing contextual information is 10.
\item \textbf{Contrastive Learning with Distant-Peer}: The dimension of the projection space for conducting contrastive learning is 189, which is one fourth of the embedding dimension of \texttt{bert-base-cased} models. In total, 100 neighbours are retrieved with their peer distance computed  according to Equation \ref{eq:peer_score}, and only 5 of them are selected as the positives that would be involved into the contrastive learning. The contrastive learning weight $\mu$ is 0.35 for TACRED and 0.02 for DocRED. The number of selected distant-peers and the contrastive learning weight $\mu$ are tuned with Bayesian optimization \citep{snoek2012practical} on the Dev sets.
\end{itemize}

\paragraph*{Hyper-parameters of Downstream Models}
The hyper-parameters of the downstream models on TACRED are identical to the original settings reported by \citet{DBLP:journals/corr/abs-2102-01373}, regardless of whether the models are trained on the original Train set or the denoised Train set. As for ATLOP model on DocRED, we modified the learning rate from its original configured 5e-5 into 2e-5 for \texttt{roberta-large} based models. The warm-up ratio is set as 0.36 for \texttt{roberta-large} based models and 0.04 for \texttt{bert-base-cased} based models.

\subsection{Zero-shot KNN vs. Fine-tuned Neural Corrector }
\label{exp:cross_val}
According to the experimental results in Table \ref{tab:baseline}, the PLM-based neural classifiers fine-tuned by cross-validation show a large advantage in AEC tasks compared to zero-shot KNN classifiers. It demonstrates that fine-tuning the PLM with the task of relation extraction is conducive to correcting annotations, because of mitigating the ambiguity of the AEC tasks for the PLMs. The neural classifiers reach the highest macro $F_1$ scores of 64.4 on TACRev, of 46.4 on Re-TACRED, and of 44.5 on Re-DocRED. Therefore, we regard the vanilla PLM-based neural classifiers as a strong baseline for contrasting our further amelioration in AEC learning.

\begin{table}[h!]
\centering
\resizebox{\textwidth}{!}{%
\begin{tabular}{ll|ll|ll|ll}
\hline
 &  & \multicolumn{2}{l|}{TACRev} & \multicolumn{2}{l|}{Re-TACRED} & \multicolumn{2}{l}{Re-DocRED} \\ \cline{3-8} 
Detection Methods & Relation Representation & Acc & $F_1$ & Acc & $F_1$ & Acc & $F_1$ \\ \hline
\multirow{3}{*}{\begin{tabular}[c]{@{}l@{}} KNN Classifier (k=1)\\ +Static Representations\end{tabular}} & Relation Prompt & 56.4 & 39.8 & 39.0 & 31.0 & 47.8 & 34.9 \\
 & Entity marker & 54.2 & 38.5 & 36.8 & 26.5 & 44.4 & 33.1 \\
 & Entity marker (punct) & 56.6 & 39.5 & 36.8 & 27.3 & 43.9 & 30.9 \\ \hline
\multirow{3}{*}{\begin{tabular}[c]{@{}l@{}}KNN Classifier (k=3)\\ +Static Representations\end{tabular}} & Relation Prompt & 45.9 & 26.0 & 27.9 & 18.4 & 50.7 & 36.1 \\
 & Entity marker & 46.1 & 25.4 & 26.4 & 13.6 & 44.9 & 29.8 \\
 & Entity marker (punct) & 49.1 & 25.5 & 27.1 & 16.5 & 43.4 & 29.2 \\ \hline
\multirow{3}{*}{\begin{tabular}[c]{@{}l@{}}Neural Classifier\\ Dynamic Representations \end{tabular}} & Relation Prompt & 75.2 & 60.6 & 49.4 & 44.4 & \textbf{57.3} & \textbf{44.5} \\
 & Entity marker & \textbf{75.7} & \textbf{64.4} & 49.7 & \textbf{46.4} & 52.4 & 28.7 \\
 & Entity marker (punct) & 72.6 & 59.1 & \textbf{50.8} & 44.5 & 55.3 & 32.7 \\ \hline
\end{tabular}%
}
\caption{The accuracy and macro $F_1$ of the KNN classifiers with static representations and neural classifiers with the dynamic representations on the Test set of TACRev, Re-TACRED and Re-DocRED. The PLM-based neural classifiers fine-tuned by cross-validation consistently outperform the static methods. }
\label{tab:baseline}
\end{table}

Among the KNN classifiers, we found that the models with $k=1$ slightly outperform the models with $k=3$ on TACRev and Re-TACRED, which differs from the observations in AID tasks (Section \ref{sec:aid_classify}). As for the AID tasks, we reckon that having more voting neighbours helps detect abnormal annotations on TACRED because more local geometry information in the embedding space is available. However, the empirical results indicate that the successful methods in AID are not necessarily effective in AEC equally. According to our analysis, the \texttt{no\_relation} examples are pervasive in TACRED. Therefore those \texttt{no\_relation} neighbours that are considered helpful in AID may mislead the KNN-based classifier in AEC. In contrast, the KNN classifiers with $k=3$ still outperform those with $k=1$ on DocRED if without distracting \texttt{no\_relation} examples.

As for the KNN-based classifiers, the \textsc{relation prompt} embedders consistently outperform the \textsc{entity marker} and \textsc{entity marker (punct)} embedders, regardless of $k=1$ or $k=3$, which supports the implications in AID task. Nevertheless, this tendency is not held under the circumstance of cross-validation. Since fine-tuning with cross-validation lets models learn the task-specific contextualized embeddings for the newly inserted tokens needed for \textsc{entity marker} and \textsc{entity marker (punct)} representations, they show competitive abilities on TACRev and Re-TACRED. The \textsc{entity marker} based neural classifiers even increases the macro $F_1$ by 3.8 on TACRev, and by 2.0 on Re-TACRED, compared to \textsc{relation prompt} based methods. However, cross validated on Re-DocRED, the \textsc{relation prompt} representations show overwhelming improvement over \textsc{entity marker} and \textsc{entity marker (punct)} representations, with a macro $F_1$ gain of 15.8 and 11.8 respectively. It is also noticeable that different entity marker types, namely new special tokens or punctuation, have their own advantages and shortcomings after cross-validation learning.

\subsection{Uncertain Labeling}
\label{exp:uncertain_label}
Furthermore, as shown in Table \ref{tab:uncertain_label}, the uncertain labels proposed in Section \ref{exp:uncertain_label} potentially enable neural classifier to better correcting suspicious annotations than certain labels. Empirically, we find that the models with \textsc{entity marker} representations are less sensitive to the certainty of learning objectives, than the \textsc{entity marker (punct)} and \textsc{relation prompt} representations. For instance, for the models with \textsc{Entity marker (punct)} representations, the KNN-based uncertain labels grant the increase of macro $F_1$ of 2.6 on TACRev, 1.8 on Re-TACRED, and 1.7 on Re-DocRED, but merely 0.1 on TACRev, 0 on Re-TACRED, and 0.4 on Re-TACRED for those with \textsc{Entity marker} representations. Similar phenomena can be observed with the approach of KDE-based soft labels, where the uncertain labels even compromise the performance of the model with \textsc{Entity marker} representations by decreasing macro $F_1$ by 0.4. We suspect that it is caused by properties of the contextualized embeddings of the newly introduced tokens after fine-tuning by cross-validation, which will be left as future work.

\begin{table}[h!]
\centering
\resizebox{\textwidth}{!}{%
\begin{tabular}{ll|ll|ll|ll}
\hline
 &  & \multicolumn{2}{l|}{TACRev} & \multicolumn{2}{l|}{Re-TACRED} & \multicolumn{2}{l}{Re-DocRED} \\ \cline{3-8} 
Labeling Methods & Relation Representation & Acc & $F_1$ & Acc & $F_1$ & Acc & $F_1$ \\ \hline
\multirow{3}{*}{Certain Label} & Relation Prompt & 75.2 & 60.6 & 49.4 & 44.4 & 57.3 & 44.5 \\
 & Entity marker & 75.7 & 64.4 & 49.7 & 46.4 & 52.4 & 28.7 \\
 & Entity marker (punct) & 72.6 & 59.1 & 50.8 & 44.5 & 55.3 & 32.7 \\ \hline
\multirow{3}{*}{KNN Uncertain Label} & Relation Prompt & 70.8 & 61.7 & 49.6 & 44.9 & 57.3 & 43.9 \\
 & Entity marker & \textbf{75.8} & \textbf{64.5} & 49.7 & 46.4 & 50.5 & 29.1 \\
 & Entity marker (punct) & 71.1 & 61.7 & 50.6 & 45.8 & 53.4 & 34.4 \\ \hline
\multirow{3}{*}{KDE Uncertain Label} & Relation Prompt & 71.1 & 64.2 & 50.2 & 46.3 & \textbf{65.0} & \textbf{52.9} \\
 & Entity marker & 74.5 & 64.0 &\textbf{51.4} &\textbf{46.9 }& 52.4 & 29.9 \\
 & Entity marker (punct) & 73.6 & 62.7 & 49.4 & 45.2 & 58.3 & 40.5 \\ \hline
\end{tabular}%
}
\caption{The accuracy and macro $F_1$ of the vanilla AEC models learnt with the labels with different certainty on the Test set of TACRev, Re-TACRED and Re-DocRED. The labels with uncertainty likely contribute to better performance in correcting annotations than overconfident hard labels.}
\label{tab:uncertain_label}
\end{table}

Generally, the KDE-based approaches of deducing the uncertain labels provide more benefits for the AEC models than KNN-based methods. Taking the models based on \textsc{relation prompt} as examples, the KDE methods improve the macro $F_1$ by 5.9 \% on TACRev, 4.2 \% on Re-TACRED and 18.8 \% on Re-DocRED.  The KNN-based label replacement only improves macro $F_1$  by 1.8 \% on TACRev,  1.1 \% on Re-TACRED, but result in a decrease of 0.1\% on Re-DocRED.

\subsection{Neighbour-aware Classifiers}
\label{exp:neigh_class}

Considering the results in Table \ref{tab:classifier}, we would argue that the distant-peer contrastive learning is a better mechanism to inject the neighbouring knowledge into the AEC classifiers than the rank-aware neighbouring encoder. As observed, the rank-aware neighbouring encoder only improves the macro $F_1$ of the AEC models with the \textsc{Entity marker (punct)} on TACRev, \textsc{Entity marker (punct)} on both Re-TACRED and Re-DocRED. Apart from these three settings, the rank-aware neighbouring encoders at most time disappointedly worse the performance of AEC models, especially for Re-DocRED. On the opposite, the distant-peer contrastive learning is considered to have the desired property of aiding the representation learning for tackling the AEC task. Aside from slightly perturbing the performance of the AEC models with \textsc{Relation Prompt} on TACRev and Re-TACRED, the multi-task loss combing the distant-peer contrastive loss and cross-entropy loss guides the models to correcting the annotation errors with enhanced prudence.

Regardless of the disillusionary overall performance of rank-aware neighbouring encoder, it reaches the highest macro $F_1$ on Re-DocRED with the \textsc{Relation Prompt} representations, exceeding the baseline by 9.2\% and surpassing the contrastive models by 11.7\%. This unexpected finding arouses the suspicion that the contrastive methods may be more vulnerable to the overconfident labels, which are supported by the results presented in the next section.

\begin{table}[h]
\centering
\resizebox{\textwidth}{!}{%
\begin{tabular}{ll|ll|ll|ll}
\hline
 &  & \multicolumn{2}{l|}{TACRev} & \multicolumn{2}{l|}{Re-TACRED} & \multicolumn{2}{l}{Re-DocRED} \\ \cline{3-8} 
Labeling Methods & Relation Representation & Acc & $F_1$ & Acc & $F_1$ & Acc & $F_1$ \\ \hline
\multirow{3}{*}{Neural Classifier} & Relation Prompt & 75.2 & 60.6 & 49.4 & 44.4 & 57.3 & 44.5 \\
 & Entity marker & \textbf{75.7} & 64.4 & 49.7 & 46.4 & 52.4 & 28.7 \\
 & Entity marker (punct) & 72.6 & 59.1 & \textbf{50.8} & 44.5 & 55.3 & 32.7 \\ \hline
\multirow{3}{*}{+ Neighbouring Encoder} & Relation Prompt & 70.1 & 58.8 & 49.7 & 44.8 & \textbf{59.2} & \textbf{48.6} \\
 & Entity marker & 69.4 & 61.3 & 49.0 & 45.8 & 37.9 & 19.7 \\
 & Entity marker (punct) & 67.6 & 60.3 & 47.5 & 43.0 & 33.0 & 17.4 \\ \hline
\multirow{3}{*}{+ Contrastive Learning} & Relation Prompt & 70.3 & 60.3 & 50.3 & 45.5 & 58.3 & 43.5 \\
 & Entity marker & 74.0 & 64.4 & 50.2 & \textbf{47.0} & 56.3 & 37.5 \\
 & Entity marker (punct) & 72.8 & \textbf{65.1} & 49.8 & 46.9 & 55.3 & 34.2 \\ \hline
\end{tabular}%
}
\caption{The accuracy and macro $F_1$ of different approaches that endow the neural relation classifiers with the neighbouring awareness with the certain labels on the Test set of TACRev, Re-TACRED and Re-DocRED. The neighbouring encoding and contrastive learning potentially make the classier to perform better at rectifying the annotations.}
\label{tab:classifier}
\end{table}

\subsection{Contrastive Learning with Uncertainty}
\label{exp:cl_with_uncertain}
After a comprehensive search for the best setup to combine the strengths of the uncertain labelling and neighbour-aware classifiers, we found that the distant-peer contrastive models learnt with KDE-based soft labels always lead to the best results among all configurations (Table \ref{tab:best_comb}). The increment of macro $F_1$ is decomposed by ablating the contribution made by the uncertain labels and neighbour-awareness.

On TACRev, solely applying the KDE uncertain labels decrease the macro $F_1$ by 0.3, but combing soft labels with distant-peer contrastive learning leads to the highest accuracy of 75.8 and macro $F_1$of 66.2. Similarly, on Re-DocRED, independent usage of distant-peer contrastive learning reduces the macro $F_1$ by 1.0, whereas collaborating with the KDE-based soft labels result in a significant improvement that reaches the highest accuracy of 66.0 and macro $F_1$ of 57.8.

The observations reveal two implications: (1) Although the models with \textsc{Entity Marker} representations already show considerable robustness in dealing with the overconfident labels, properly introducing the uncertainty is still meaningful if the models are further ameliorated with distant-peer contrastive learning. (2) The distant-peer contrastive learning is sensitive to the noise in certain labels when the models are based on \textsc{relation prompt} representations, but the KDE-based soft labelling can mitigate the vulnerability to a large extent. 

\begin{table}[h]
\centering
\begin{tabular}{llll}
\hline
Methods & Acc & $F_1$ & $\Delta$ $F_1$ \\ \hline
\multicolumn{4}{l}{\textit{Best Combination on TACRev}} \\
Neural Classifier with Entity Marker Representation & 75.6 & 64.3 &  \\
+ KDE Uncertain Label & 74.5 & 64.0 & - 0.3 \\
+ Distant-peer Contrastive Learning & 74.0 & 64.5 & + 0.2 \\
\hspace{2mm} + KDE Uncertain Label & \textbf{75.8} & \textbf{66.2} & \textbf{+ 1.9} \\ \hline
\multicolumn{4}{l}{\textit{Best Combination on Re-TACRED}} \\
Neural Classifier with Entity Marker Representation & 49.3 & 44.3 &  \\
+ KDE Uncertain Label & 51.4 & 46.9 & + 2.6 \\
+ Distant-peer Contrastive Learning & 50.2 & 47.0 & + 2.7 \\
\hspace{2mm}  + KDE Uncertain Label &\textbf{52.2} & \textbf{47.7} & \textbf{+ 3.4} \\\hline
\multicolumn{4}{l}{\textit{Best Combination on Re-DocRED}} \\
Neural Classifier with Relation Prompt Representation & 57.2 & 44.5 &  \\
+ KDE Uncertain Label & 65.0 & 52.9 & + 8.4 \\
 + Distant-peer Contrastive Learning& 58.2 & 43.5 & -1.0 \\
\hspace{2mm}+ KDE Uncertain Label  & \textbf{66.0} & \textbf{57.8} & \textbf{+ 13.3} \\ \hline
\end{tabular} 
\caption{The macro $F_1$ and the improvement of the macro $F_1$ under the best combinations of proposed methods on different datasets. The combination of KDE uncertain labels and distant-peer contrastive learning evidently unleashes the true potential of AEC models.}
\label{tab:best_comb}
\end{table}

\begin{figure}[h]
\includegraphics[width=\linewidth]{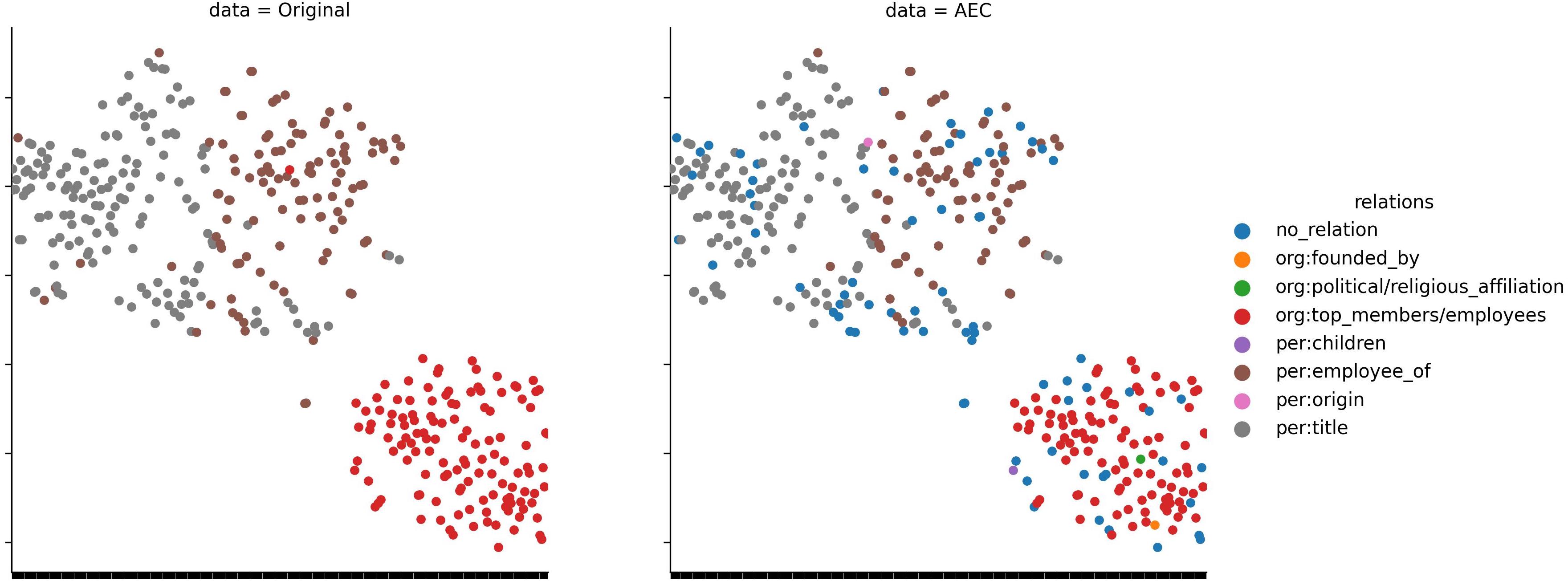}
\caption{Illustration of the revising outcome on TACRED by the optimal AEC models, presented by visualizing the \textsc{relation prompt} embedding of randomly sampled instances from relations \texttt{per:title}, \texttt{per:employee\_of} and \texttt{per:top\_members/employees} classes. \textbf{Left}: examples with the original annotations. \textbf{Right}: examples with the revised annotations predicted by the AEC models.}
\label{fig:aec_example}
\end{figure}

The statistics in Table \ref{tab:downstream} implies that the optimized AEC models would totally revise 10.7\% examples on the TACRED Train set, and 20.0\% examples on DocRED Train set. Furthermore, we visualize a part of decisions made by our proposed AEC models to have a more intuitive sense of the consequence. Figure \ref{fig:aec_example} depicts the partial revising outcome by the optimal models on TACRED. The examples were initially labeled with the relations \texttt{per:title}, \texttt{per:employee\_of} and \texttt{per:top\_members/employees}, that are ambiguous and error-prone to annotate as discussed in Section \ref{sec:trade_off}. After automatic rectification, most examples have been re-labeled as \texttt{no\_relation}, which is compatible with the propensity of revisions made in TACRev. The examples initially labeled as \texttt{per:top\_members/employees} (a red point) with the representation located in the cluster of \texttt{per:employee\_of} (brown points) is revised into \texttt{per:employee\_of} (brown points) by the AEC models. Several \texttt{per:top\_members/employees} labels are reasonably changed into \texttt{org:founded\_by}, \texttt{org:political/religious\_affiliation} or \texttt{per:children}. Overall, the automatic revisions are congenial with reason and common sense.

\subsection{Learning on Denoised Train Sets}
\label{exp:downstream}
Assuming the annotation noise is evenly spread across all data splits, the test error caused by the noise in Test sets may not fully reflect the true contribution of AEC. Still, training on the Train set denoised by our proposed AEC model empirically led to the enhancement of both sentence-level and document-level relation extraction models, even evaluated on the probably flawed Test sets of TACRED, TACRev and DocRED. According to \citet{yao-etal-2019-docred}, the Ign $F_1$ reported on. DocRED is the Micro $F_1$ excluding those relational facts shared by the Train, Dev and Test sets. 

\subsubsection*{TACRED}
In Table \ref{tab:down_tacred}, we recognize the persistent advancement by training the models with \textsc{entity marker} representations on denoised TACRED Train set. Especially, the model based on BERT-base receives the maximum gain of Micro $F_1$ by 3.5 \% on TACRED and by 3.4 \% on TACRev by learning with denoised annotations. The model based on RoBERTa-large initially is inferior to the KnowBERT \citep{DBLP:conf/emnlp/PetersNLSJSS19} when the predictions are evaluated with TACRED Test sets. However, the enhancement brought by the denoised data can be captured by evaluating with both the Test sets of TACRED and TACRev, enabling the model to reach higher Micro $F_1$ on TACRED Test set than the KnowBERT \citep{DBLP:conf/emnlp/PetersNLSJSS19}.

\begin{table}[h]
\centering
\begin{tabular}{lll}
\hline
Models & TACRED & TACRev \\
 & Micro $F_1$ & Micro $F_1$ \\ \hline
\multicolumn{3}{l}{\textit{Other Models Trained on TACRED Train Set}} \\
PA-LSTM \citep{YuhaoZhang2017PositionawareAA} & 65.1 & 73.3 \\
C-GCN \citep{zhang-etal-2018-graph} & 66.3 & 74.6 \\
SpanBERT\citep{joshi-etal-2020-spanbert} & 70.8 & 78.0 \\
KnowBERT \citep{DBLP:conf/emnlp/PetersNLSJSS19} & 71.5 & 79.3 \\ \hline
\multicolumn{3}{l}{\textit{Investigated Models Trained on TACRED Train Set}} \\
BERT-base + entity marker \citep{DBLP:journals/corr/abs-2102-01373} & 68.4 & 77.2 \\
BERT-large + entity marker \citep{DBLP:journals/corr/abs-2102-01373} & 69.7 & 77.9 \\
RoBERTa-large + entity marker \citep{DBLP:journals/corr/abs-2102-01373} & 70.7 & 81.2 \\ \hline
\multicolumn{3}{l}{\textit{Investigated Models Trained on De-noised TACRED Train Set}} \\
BERT-base + entity marker & 70.8 & 79.8 \\
BERT-large + entity marker & 71.1 & 80.4 \\
RoBERTa-large + entity marker & \textbf{72.1} & \textbf{81.9} \\ \hline
\end{tabular}
\caption{The Micro $F_1$ Test scores on TACRED and TACRev of the sentence-level relation extraction models trained on original Train set and automatically denoised Train set. Our proposed AEC models can further enhance the SOTA models by improving the quality of training data. }
\label{tab:down_tacred}
\end{table}

\subsubsection*{DocRED}
The results in Table \ref{tab:down_docred} demonstrate that the denoised data are also beneficial to the SOTA document-level relation extraction models. However, the advantage of denoised data is less obvious than we observed with the investigated models on TACRED. Automatically denoised data slightly increases the Ign $F_1$ and $F_1$ of ATLOP BERT-base model  by 1.1\% and 0.7\% respectively, and improve the Ign $F_1$ and $F_1$ of ATLOP RoBERTa-large model by 0.6\% and 0.8\%. Therefore, it is reasonable to believe that our currently implemented AEC model may be more successful in correcting the relation annotation in the sentence-level than document-level. Since document-level relation extraction task usually makes greater demands of capturing the inter-sentence interactions, adding hierarchical context modelling to AEC models is a promising direction for future exploration.  
\begin{table}[h]
\centering
\begin{tabular}{lll}
\hline
Models & \multicolumn{2}{l}{DocRED} \\
 & Ign $F_1$ & $F_1$ \\ \hline
\multicolumn{3}{l}{\textit{Other Models Trained on DocRED Train Set}} \\
CNN \citep{yao-etal-2019-docred} & 40.33 & 42.26 \\
BiLSTM \citep{yao-etal-2019-docred} & 48.78 & 51.06 \\
BERT-LSR-base \citep{DBLP:conf/acl/NanGSL20} & 56.97 & 59.05 \\
HIN-BERT-base \citep{DBLP:conf/pakdd/TangC0CFWY20} & 53.70 & 55.60 \\
CorefBERT-base \citep{ye-etal-2020-coreferential}  & 54.54 & 56.96 \\
CorefRoBERTa-large \citep{ye-etal-2020-coreferential} & 57.90 & 60.25 \\ \hline
\multicolumn{3}{l}{Investigated Models Trained on DocRED Train Set} \\
ATLOP-BERT-base \citep{WenxuanZhou2020DocumentLevelRE} & 59.31 & 61.30 \\
ATLOP-RoBERTa-large \citep{WenxuanZhou2020DocumentLevelRE} & 61.39 & 63.40 \\ \hline
\multicolumn{3}{l}{Investigated Models Trained on De-noised DocRED Train Set} \\
ATLOP-BERT-base &   59.98 & 61.73 \\
ATLOP-RoBERTa-large & \textbf{61.75} & \textbf{63.88} \\ \hline
\end{tabular}
\caption{The Micro $F_1$ and Ign Micro $F_1$ Test scores on DocRED of the document-level relation extraction models trained with the original Train examples and the denoised Train examples. Similarly, automatically rectifying the Train set leads to the boost of performance in document-level relation extraction. }
\label{tab:down_docred}
\end{table}

\section{Discussions}
\subsection{Revision Quality of Automatic Re-Annotator}
We further analyze the revision quality of our proposed AEC methods from two aspects: (1) quantification of the agreement between automatic revisions and manual revisions, and (2) case analysis of the positive and negative predictions by AEC models.

\subsubsection*{Agreement between Human and AEC Model }

Cohen Kappa score $\kappa$ \citep{cohen1960coefficient, DBLP:journals/coling/ArtsteinP08} is a widely accepted method to assess the rate of agreement between two different annotators, which is also informative to quantify the consistency between human and machine revisions. Compared to the $F_1$ score we used for evaluating the AEC performance while regarding the human revision as the ground truth, the Cohen Kappa score tries to contrast the observed agreement between the human and AEC model with the expected agreement when both of them revise labels randomly. The range of Cohen Kappa score is $[-1, 1]$, and the score above 0 indicates that there is an agreement between two raters, and the score above 0.8 generally means the agreement is considerable. 

The Cohen Kappa score $\kappa$ reported in Table \ref{tab:cohen} indicates that: (1) The annotators composing TACRev and Re-TACRED are in excellent agreement and awarded with the Cohen Kappa score of 0.882. (2) The Cohen Kappa score between TACRev and the revisions by the AEC model is 0.587. It reveals that though the agreement between humans and the AEC model is worse than inter-human, the decisions made by the AEC model still demonstrate good consistency with human revisions. Generally, on the basis of the Cohen Kappa score $\kappa$, the overall agreement between humans and our proposed AEC model is acceptable.

\begin{table}[h!]
\centering
\begin{tabular}{@{}lllc@{}}
\toprule
Dataset & Revision A & Revision B & Cohen Kappa $\kappa$ \\ \midrule
\multirow{3}{*}{TACRED} & TACRev & Re-TACRED & 0.882 \\
 & TACRev & AEC TACRED & 0.587 \\ \midrule
DocRED & Re-DocRED & AEC DocRED & 0.604 \\ \bottomrule
\end{tabular}
\caption{The Cohen kappa score between different revisions on TACRED and DocRED. The corrections predicted by the optimized AEC models demonstrate promising agreement with the manual revisions. }
\label{tab:cohen}
\end{table}

\subsubsection*{Case Study of Automatic Revision}
To better understand the actual performance of our proposed AEC models, we manually re-examine around 100 revising decisions made on TACRED and DocRED, and the typical positive and negative corrections are listed in Table \ref{tab:case}. 

The positive example in TACRED was originally annotated as \texttt{per:other family}, but the word "daughter" in the context strongly hints that \texttt{per:children} may be a better label in this case, which has been nicely apprehended by our proposed model. The context of exemplified negative example in TACRED is relatively intricate even for the human annotators because it would easily mislead us to believe that "City Council" is the workplace of "Dixon". However, if we read carefully, we could notice that this assumption is actually incorrect in the given context, so \texttt{no\_relation} would be a more appropriate label here. Although the AEC model also could not give the correct answer, letting the relation label remain the same as the original one is a rational choice that would not introduce more noise after automatic revision.

The head and tail entities of the positive example in DocRED are the famous English writer "John Fowles" and one of his well-known novels "The French Lieutenant's Woman". Possibly due to the distracting word "French" in the book title, its original relation label in DocRED is \texttt{country}, which is incorrect. Conversely, our proposed AEC model reasonably revised the label into \texttt{notable work} in agreement with human revisers, which implicitly reflects that pre-training on the large-scale general corpus like Wikipedia educates the PLMs with common-sense knowledge. As for the negative example in DocRED, the originally labelled relation between "Thailand" and "Thai" is \texttt{located in}. This is incorrect because the word "Thai" can only refer to a native or inhabitant of Thailand or to the official language of Thailand. In contrast, the AEC revision of \texttt{ethnic group} is more understandable compared to the original annotation, though it may not be as accurate as of the human revision of \texttt{official language} in the given context.

\begin{table}[h!]
\centering
\resizebox{\textwidth}{!}{%
\begin{tabular}{@{}llll@{}}
\toprule
\multicolumn{4}{l}{\textit{Examples of Automatic Corrections on TACRED}} \\
\multirow{2}{*}{Pos} & \multicolumn{3}{l}{\textbf{{[}Tessa Dahl{]} }'s daughter is the model and writer \textbf{\textless{}Sophie Dahl\textgreater{}}.} \\
 & \multicolumn{1}{l}{\textbf{Annotation:} \texttt{per:other\_family} } & \textbf{AEC Model:} \texttt{per:children} & \textbf{Human: }\texttt{per:children} \\ \cmidrule(l){2-4} 
\multirow{2}{*}{Neg} & \multicolumn{3}{l}{\begin{tabular}[c]{@{}l@{}}If those efforts fail, \textbf{{[}Dixon{]}} would probably be forced from office and the \textbf{\textless{}City Council\textgreater} president, \\Stephanie Rawlings-Blake, would succeed her.\end{tabular}} \\
 & \multicolumn{1}{l}{\textbf{Annotation:} \texttt{per:employee\_of}} & \textbf{AEC Model:} \texttt{per:employee\_of} & \textbf{Human: }\texttt{no\_relation} \\ \midrule
\multicolumn{4}{l}{\textit{Examples of Automatic Corrections on DocRED}} \\
\multirow{2}{*}{Pos} & \multicolumn{3}{l}{\begin{tabular}[c]{@{}l@{}}Works often described as examples of historiographical metafiction include: William Shakespeare 's \\Pericles, Prince of Tyre (c.1608), \textbf{{[}John Fowles{]}}'s \textbf{\textless{}The French Lieutenant 's Woman\textgreater} (1969), ....\end{tabular}} \\
 & \multicolumn{1}{l}{\textbf{Annotation:} \texttt{country}} & \textbf{AEC Model:} \texttt{notable work} & \textbf{Human:} \texttt{notable work} \\ \cmidrule(l){2-4} 
\multirow{2}{*}{Neg} & \multicolumn{3}{l}{\begin{tabular}[c]{@{}l@{}}Chupong Changprung (RTGS: Dan Chupong); born March 23, 1981 in Kalasin Province, \textbf{{[}Thailand{]}}, \\ \textbf{\textless{}Thai\textgreater} nickname: "Deaw" is a \textbf{\textless{}Thai\textgreater} martial arts film actor ......\end{tabular}} \\
 & \multicolumn{1}{l}{\textbf{Annotation:} \texttt{located in}} & \textbf{AEC Model:} \texttt{ethnic group} & \textbf{Human:} \texttt{official language} \\ \bottomrule
\end{tabular}%
}
\caption{The case study of positive and negative annotation corrections made by our proposed AEC models. The models have the due capability in making the reasonable decision of rectifying annotations, though they may not be as precise as human revisions.}
\label{tab:case}
\end{table}

\subsection{Efficiency of Automatic Re-Annotator}
Table \ref{tab:time} shows the statistics of the average time in seconds required for human and automatic re-annotators to reexamine and possibly revise each sample of TACRED and DocRED, respectively. We estimate the efficiency of human revisers through the annotation progress of the re-annotator who helped us revise 411 examples of DocRED in 35 hours. A more detailed analysis of the efficiency of human re-annotators will be conducted in the future. The annotation speed of the annotation inconsistency detector and annotation error corrector is computed by the averaging run time of all experiments, including different configurations. Although the estimated efficiency may not be precise, we still can compare performance in terms of orders of magnitude.

\begin{table}[h!]
\centering
\begin{tabular}{lll}
\hline
Re-Annotator & TACRED & DocRED \\ \hline
Human Reviser & 47.68 & 306.63 \\
Annotation Inconsistency Detector & 4.27 $\times 10^{-3}$  & 6.69 $\times 10^{-3}$ \\
Annotation Error Corrector & 0.32 & 3.33 \\ \hline
\end{tabular}
\caption{Comparison the efficiency between human revisers and our proposed two types of automatic re-annotators, by the average time (in seconds) of relabeling each example. The PLM-based re-annotators are at least hundreds of times faster than the human revisers.}
\label{tab:time}
\end{table}

On average, the human revisers need to spend around 47.68 seconds to revise each annotation on sentence-level relation extraction corpus TACRED and 306.63 seconds, ($\sim$5 minutes), to revise each annotation on document-level corpus DocRED. In comparison, our proposed automatic re-annotators are significantly faster. As the annotation inconsistency detectors only involve the inference stage of PLMs, they only take $4.27 \times 10^{-3}$ seconds and $6.69 \times 10^{-3}$ seconds to assess the validity of each annotation on TACRED and DocRED, respectively, which is over 10$^4$ times faster than the human. In contrast, the annotation error corrector depends on the time-consuming fine-tuning stage of PLMs to give exact suggestions in revising annotations by cross-validation. It would cost about 0.32 seconds to revise on TACRED and 3.33 seconds on DocRED. The difference between re-annotation efficiency on TACRED and DocRED is due to the difference in their input length. Despite the annotation error corrector being slower than annotation inconsistency detectors, it still demonstrates the re-annotation speed at least hundreds of times higher than the human re-annotators.

In sum, our proposed pair of re-annotators could improve the effective efficiency of re-annotation. The annotation inconsistency detectors can help spot the most suspicious labels for human revisers, in order to reduce the overall time for dataset revision or filter out the unreliable samples from the training set to alleviate the negative effects. Moreover, the annotation error corrector based on the prior knowledge in PLMs shows both competitive re-annotation efficiency and accuracy, even when compared with human revisers.

\chapter{Conclusions and Future Directions}
\label{chap:conclusion}
The re-annotation tasks we investigated, Annotation Inconsistency Detection and Annotation Error Correction, have significant implications for prospective data-driven Natural Language Processing researches. Through empirical study, we demonstrate the non-negligible potential of Pre-trained Language Models in reducing annotation noise. Moreover, we point out the appealing directions to further extend our work in the future.

\section{Conclusions}
In conclusion, Pre-trained Language Models (PLMs) show impressive potential as automatic re-annotators. Through detailed experiments conducted on sentence-level and document-level relation extraction datasets (TACRED and DocRED) and their human revised versions, we have shown that a PLM-based annotation inconsistency detector and annotation error corrector can be used to improve annotation quality more efficiently.

Our investigations in Annotation Inconsistency Detection indicate a well-designed prompt can induce PLMs to acquire instance representations with favourable properties to detect inconsistencies in zero-shot scenarios, which profits from prior knowledge in PLMs. Compared to the K-Nearest Neighbour classifier, our proposed credibility scores jointly consider both the distance and trustworthiness of K-Nearest neighbours. The credibility-based detector yields better sensitivity and specificity of detecting annotation inconsistencies than the vote-based K-Nearest Neighbour classifier. The best binary $F_1$ scores that our proposed inconsistency detector can reach are 92.4 on TACRED and 72.5 on DocRED.

Our observations in Annotation Error Correction suggest that fine-tuning with proper domain-specific knowledge by cross-validation further improves the capability of PLMs in recommending precise revisions for the suspicious annotations. Given the vulnerability of the softmax classifier to annotation noise, we mitigate the overconfidence of the automatic corrector by introducing uncertainty to observed hard labels. We combine a novel distant-peer contrastive loss with the ordinary cross-entropy loss to improve neighbour awareness while learning to correct annotations. The distant-peer contrastive framework selects the positive neighbours by their label co-occurrence and distance from the anchor example. The results demonstrate that the distant-peer contrastive corrector learnt with Kernel Density Estimation based soft label obtains the best performance in annotation correction throughout the study. The highest macro $F_1$ scores we observed throughout the study of error corrector are 66.2 on TACRED and 57.8 on DocRED.

In practice, we prove that the state-of-the-art model can benefit from learning using data automatically denoised by our proposed annotation corrector. Automatic correction at most leads to a gain of 3.6\% to the downstream state-of-the-art relation extraction models. In terms of time cost, an automatic inconsistency detector is capable of reexamining each sample in the dataset over ten thousand times more efficiently than a human. Moreover, an automatic error corrector can even suggest the recommended revision for annotations in question hundreds of times faster compared to manual revision with acceptable accuracy. We believe that our findings are extremely valuable to prospective data-driven NLP research.

\section{Future Directions}

This dissertation project opens up the following potential directions for future research in data-driven NLP:
\begin{itemize}
\item For example, we did not explore re-annotating very domain-specific datasets, such as medical reports or legal documents. Annotation noise is also assumed to be pervasive in those datasets, so it would be of interest to investigate if domain-specific PLMs \citep{DBLP:journals/bioinformatics/LeeYKKKSK20, DBLP:conf/emnlp/FengGTDFGS0LJZ20, DBLP:journals/corr/abs-2004-03760} can help to reveal the annotation inconsistencies and errors in such domains.
\item For the credibility score, we manually tuned the threshold for the credibility-based detectors, but learning the threshold automatically would be worth exploring as well.
\item Through our experiments contrasting the influence of different input formats to the annotation corrector, we surprisingly found that the \texttt{Entity Marker} with newly introduced special tokens show outstanding robustness to training noise. Considering there is no systematic analysis on the effect of freshly introduced special tokens, we are also curious about their potential advantages in noise-tolerant learning.
\item Sentence-level and document-level annotation revisions are treated similarly by the automatic re-annotators presented in this dissertation, without explicit consideration of complex inter-sentence interactions and document structure. Therefore, we believe that methods for modelling the document hierarchy or discourse relations will further improve the performance of document-level re-annotation.
\item The comparison of human and automatic revision was carried out using rough estimates. The annotation efficiency of humans can be impacted by various factors, such as the interface of the annotation platform, the level of skill and training of the annotators as well as their mental condition. The efficiency of automatic re-annotation can also vary depending on the chosen PLMs, the implementation details and the experimental setup. Hence, a more  systematic analysis of the difference between human and automatic re-annotators may be conducive to prospective human-central NLP.
\end{itemize}

Aside from the aforementioned directions, the promising results obtained using our proposed prompt-based credibility score have led us to come up with ideas for extending this work further. Recently, prompt-based methods demonstrate impressive zero-shot learning capability in diverse downstream tasks with impressive zero-shot learning capability \citep{DBLP:journals/corr/abs-2107-13586}. Theoretically, we can derive credibility scores on almost all NLP datasets with the appropriate prompts. Under the human-in-the-loop setting, a prompt-based credibility score could help human re-annotators to focus on the most suspicious annotations to speed up their work. Under the noise-tolerant learning setting, the prompt-based credibility score has the potential to improve model performance in various NLP tasks. For instance, when applying curriculum learning \citep{DBLP:conf/icml/BengioLCW09}, the model can greedily learn on the relatively clean data first, and then prudently learn on data that is considered less accurate and consistent by the prompt-based credibility score. We can also design the sampling weight based on this prompt-based approach, encouraging the models to learn on reliable samples most of the time.



\appendix

\bibliographystyle{apalike}


\bibliography{thesis, anthology}

\begin{thebibliography}{}

\bibitem[Abney et~al., 1999]{DBLP:conf/emnlp/AbneySS99}
Abney, S., Schapire, R.~E., and Singer, Y. (1999).
\newblock Boosting applied to tagging and {PP} attachment.
\newblock In Fung, P. and Zhou, J., editors, {\em Joint {SIGDAT} Conference on
  Empirical Methods in Natural Language Processing and Very Large Corpora,
  {EMNLP} 1999, College Park, MD, USA, June 21-22, 1999}. Association for
  Computational Linguistics.

\bibitem[Alex et~al., 2010]{DBLP:conf/acllaw/AlexGSK10}
Alex, B., Grover, C., Shen, R., and Kabadjov, M.~A. (2010).
\newblock Agile corpus annotation in practice: An overview of manual and
  automatic annotation of cvs.
\newblock In Xue, N. and Poesio, M., editors, {\em Proceedings of the Fourth
  Linguistic Annotation Workshop, {LAW} 2010, Uppsala, Sweden, July 15-16,
  2010}, pages 29--37. Association for Computational Linguistics.

\bibitem[Algan and Ulusoy, 2021]{DBLP:journals/corr/abs-2103-10869}
Algan, G. and Ulusoy, I. (2021).
\newblock Metalabelnet: Learning to generate soft-labels from noisy-labels.
\newblock {\em CoRR}, abs/2103.10869.

\bibitem[Allen, 1974]{doi:10.1080/00401706.1974.10489157}
Allen, D.~M. (1974).
\newblock The relationship between variable selection and data agumentation and
  a method for prediction.
\newblock {\em Technometrics}, 16(1):125--127.

\bibitem[Alt et~al., 2020]{DBLP:conf/acl/AltGH20a}
Alt, C., Gabryszak, A., and Hennig, L. (2020).
\newblock {TACRED} revisited: {A} thorough evaluation of the {TACRED} relation
  extraction task.
\newblock In Jurafsky, D., Chai, J., Schluter, N., and Tetreault, J.~R.,
  editors, {\em Proceedings of the 58th Annual Meeting of the Association for
  Computational Linguistics, {ACL} 2020, Online, July 5-10, 2020}, pages
  1558--1569. Association for Computational Linguistics.

\bibitem[Angluin and Laird, 1987]{DBLP:journals/ml/AngluinL87}
Angluin, D. and Laird, P.~D. (1987).
\newblock Learning from noisy examples.
\newblock {\em Mach. Learn.}, 2(4):343--370.

\bibitem[Argyris, 2004]{argyris2004reasons}
Argyris, C. (2004).
\newblock {\em Reasons and rationalizations: The limits to organizational
  knowledge}.
\newblock Oxford University Press on Demand.

\bibitem[Artstein and Poesio, 2008]{DBLP:journals/coling/ArtsteinP08}
Artstein, R. and Poesio, M. (2008).
\newblock Inter-coder agreement for computational linguistics.
\newblock {\em Comput. Linguistics}, 34(4):555--596.

\bibitem[Aydar et~al., 2020]{DBLP:journals/corr/abs-2007-04247}
Aydar, M., Bozal, O., and {\"{O}}zbay, F. (2020).
\newblock Neural relation extraction: a survey.
\newblock {\em CoRR}, abs/2007.04247.

\bibitem[Barnett, 1978]{barnett1978study}
Barnett, V. (1978).
\newblock The study of outliers: purpose and model.
\newblock {\em Journal of the Royal Statistical Society: Series C (Applied
  Statistics)}, 27(3):242--250.

\bibitem[Beck et~al., 2020]{beck2020representation}
Beck, C., Booth, H., El-Assady, M., and Butt, M. (2020).
\newblock Representation problems in linguistic annotations: Ambiguity,
  variation, uncertainty, error and bias.
\newblock In {\em 14th Linguistic Annotation Workshop}, pages 60--73.

\bibitem[Bengio et~al., 2009]{DBLP:conf/icml/BengioLCW09}
Bengio, Y., Louradour, J., Collobert, R., and Weston, J. (2009).
\newblock Curriculum learning.
\newblock In Danyluk, A.~P., Bottou, L., and Littman, M.~L., editors, {\em
  Proceedings of the 26th Annual International Conference on Machine Learning,
  {ICML} 2009, Montreal, Quebec, Canada, June 14-18, 2009}, volume 382 of {\em
  {ACM} International Conference Proceeding Series}, pages 41--48. {ACM}.

\bibitem[Bhadra and Hein, 2015]{DBLP:journals/ijon/BhadraH15}
Bhadra, S. and Hein, M. (2015).
\newblock Correction of noisy labels via mutual consistency check.
\newblock {\em Neurocomputing}, 160:34--52.

\bibitem[Brodley and Friedl, 1999]{DBLP:journals/jair/BrodleyF99}
Brodley, C.~E. and Friedl, M.~A. (1999).
\newblock Identifying mislabeled training data.
\newblock {\em J. Artif. Intell. Res.}, 11:131--167.

\bibitem[Brown et~al., 2020]{DBLP:conf/nips/BrownMRSKDNSSAA20}
Brown, T.~B., Mann, B., Ryder, N., Subbiah, M., Kaplan, J., Dhariwal, P.,
  Neelakantan, A., Shyam, P., Sastry, G., Askell, A., Agarwal, S.,
  Herbert{-}Voss, A., Krueger, G., Henighan, T., Child, R., Ramesh, A.,
  Ziegler, D.~M., Wu, J., Winter, C., Hesse, C., Chen, M., Sigler, E., Litwin,
  M., Gray, S., Chess, B., Clark, J., Berner, C., McCandlish, S., Radford, A.,
  Sutskever, I., and Amodei, D. (2020).
\newblock Language models are few-shot learners.
\newblock In Larochelle, H., Ranzato, M., Hadsell, R., Balcan, M., and Lin, H.,
  editors, {\em Advances in Neural Information Processing Systems 33: Annual
  Conference on Neural Information Processing Systems 2020, NeurIPS 2020,
  December 6-12, 2020, virtual}.

\bibitem[Bryant, 2019]{DBLP:phd/ethos/Bryant19}
Bryant, C.~J. (2019).
\newblock {\em Automatic annotation of error types for grammatical error
  correction}.
\newblock PhD thesis, University of Cambridge, {UK}.

\bibitem[Cao et~al., 2021]{DBLP:conf/iclr/CaoI0P21}
Cao, N.~D., Izacard, G., Riedel, S., and Petroni, F. (2021).
\newblock Autoregressive entity retrieval.
\newblock In {\em 9th International Conference on Learning Representations,
  {ICLR} 2021, Virtual Event, Austria, May 3-7, 2021}. OpenReview.net.

\bibitem[Chandola et~al., 2009]{DBLP:journals/csur/ChandolaBK09}
Chandola, V., Banerjee, A., and Kumar, V. (2009).
\newblock Anomaly detection: {A} survey.
\newblock {\em {ACM} Comput. Surv.}, 41(3):15:1--15:58.

\bibitem[Chaudhuri et~al., 2021]{DBLP:conf/esws/ChaudhuriR021}
Chaudhuri, D., Rony, M. R. A.~H., and Lehmann, J. (2021).
\newblock Grounding dialogue systems via knowledge graph aware decoding with
  pre-trained transformers.
\newblock In Verborgh, R., Hose, K., Paulheim, H., Champin, P., Maleshkova, M.,
  Corcho, {\'{O}}., Ristoski, P., and Alam, M., editors, {\em The Semantic Web
  - 18th International Conference, {ESWC} 2021, Virtual Event, June 6-10, 2021,
  Proceedings}, volume 12731 of {\em Lecture Notes in Computer Science}, pages
  323--339. Springer.

\bibitem[Chen et~al., 2019]{DBLP:conf/icml/ChenLCZ19}
Chen, P., Liao, B., Chen, G., and Zhang, S. (2019).
\newblock Understanding and utilizing deep neural networks trained with noisy
  labels.
\newblock In Chaudhuri, K. and Salakhutdinov, R., editors, {\em Proceedings of
  the 36th International Conference on Machine Learning, {ICML} 2019, 9-15 June
  2019, Long Beach, California, {USA}}, volume~97 of {\em Proceedings of
  Machine Learning Research}, pages 1062--1070. {PMLR}.

\bibitem[Chen et~al., 2020]{DBLP:conf/icml/ChenK0H20}
Chen, T., Kornblith, S., Norouzi, M., and Hinton, G.~E. (2020).
\newblock A simple framework for contrastive learning of visual
  representations.
\newblock In {\em Proceedings of the 37th International Conference on Machine
  Learning, {ICML} 2020, 13-18 July 2020, Virtual Event}, pages 1597--1607.

\bibitem[Chen et~al., 2021]{DBLP:journals/corr/abs-2104-07650}
Chen, X., Zhang, N., Xie, X., Deng, S., Yao, Y., Tan, C., Huang, F., Si, L.,
  and Chen, H. (2021).
\newblock Knowprompt: Knowledge-aware prompt-tuning with synergistic
  optimization for relation extraction.
\newblock {\em CoRR}, abs/2104.07650.

\bibitem[Chi et~al., 2020]{DBLP:conf/acl/ChiHM20}
Chi, E.~A., Hewitt, J., and Manning, C.~D. (2020).
\newblock Finding universal grammatical relations in multilingual {BERT}.
\newblock In Jurafsky, D., Chai, J., Schluter, N., and Tetreault, J.~R.,
  editors, {\em Proceedings of the 58th Annual Meeting of the Association for
  Computational Linguistics, {ACL} 2020, Online, July 5-10, 2020}, pages
  5564--5577. Association for Computational Linguistics.

\bibitem[Clark et~al., 2019]{DBLP:conf/blackboxnlp/ClarkKLM19}
Clark, K., Khandelwal, U., Levy, O., and Manning, C.~D. (2019).
\newblock What does {BERT} look at? an analysis of bert's attention.
\newblock In Linzen, T., Chrupala, G., Belinkov, Y., and Hupkes, D., editors,
  {\em Proceedings of the 2019 {ACL} Workshop BlackboxNLP: Analyzing and
  Interpreting Neural Networks for NLP, BlackboxNLP@ACL 2019, Florence, Italy,
  August 1, 2019}, pages 276--286. Association for Computational Linguistics.

\bibitem[Cohen, 1960]{cohen1960coefficient}
Cohen, J. (1960).
\newblock A coefficient of agreement for nominal scales.
\newblock {\em Educational and psychological measurement}, 20(1):37--46.

\bibitem[Crowston, 2012]{DBLP:conf/ifip8-2/Crowston12}
Crowston, K. (2012).
\newblock Amazon mechanical turk: {A} research tool for organizations and
  information systems scholars.
\newblock In Bhattacherjee, A. and Fitzgerald, B., editors, {\em Shaping the
  Future of {ICT} Research. Methods and Approaches - {IFIP} {WG} 8.2, Working
  Conference, Tampa, FL, USA, December 13-14, 2012. Proceedings}, volume 389 of
  {\em {IFIP} Advances in Information and Communication Technology}, pages
  210--221. Springer.

\bibitem[Cui et~al., 2021]{DBLP:conf/acl/CuiWLYZ21}
Cui, L., Wu, Y., Liu, J., Yang, S., and Zhang, Y. (2021).
\newblock Template-based named entity recognition using {BART}.
\newblock In Zong, C., Xia, F., Li, W., and Navigli, R., editors, {\em Findings
  of the Association for Computational Linguistics: {ACL/IJCNLP} 2021, Online
  Event, August 1-6, 2021}, volume {ACL/IJCNLP} 2021 of {\em Findings of
  {ACL}}, pages 1835--1845. Association for Computational Linguistics.

\bibitem[Cui et~al., 2017]{DBLP:conf/ccks/CuiLWY17}
Cui, M., Li, L., Wang, Z., and You, M. (2017).
\newblock A survey on relation extraction.
\newblock In Li, J., Zhou, M., Qi, G., Lao, N., Ruan, T., and Du, J., editors,
  {\em Knowledge Graph and Semantic Computing. Language, Knowledge, and
  Intelligence - Second China Conference, {CCKS} 2017, Chengdu, China, August
  26-29, 2017, Revised Selected Papers}, volume 784 of {\em Communications in
  Computer and Information Science}, pages 50--58. Springer.

\bibitem[Davison et~al., 2019]{DBLP:conf/emnlp/DavisonFR19}
Davison, J., Feldman, J., and Rush, A.~M. (2019).
\newblock Commonsense knowledge mining from pretrained models.
\newblock In Inui, K., Jiang, J., Ng, V., and Wan, X., editors, {\em
  Proceedings of the 2019 Conference on Empirical Methods in Natural Language
  Processing and the 9th International Joint Conference on Natural Language
  Processing, {EMNLP-IJCNLP} 2019, Hong Kong, China, November 3-7, 2019}, pages
  1173--1178. Association for Computational Linguistics.

\bibitem[Devlin et~al., 2019]{devlin-etal-2019-bert}
Devlin, J., Chang, M.-W., Lee, K., and Toutanova, K. (2019).
\newblock {BERT}: Pre-training of deep bidirectional transformers for language
  understanding.
\newblock In {\em Proceedings of the 2019 Conference of the North {A}merican
  Chapter of the Association for Computational Linguistics: Human Language
  Technologies, Volume 1 (Long and Short Papers)}, pages 4171--4186,
  Minneapolis, Minnesota. Association for Computational Linguistics.

\bibitem[Dickinson, 2010]{DBLP:conf/acl/Dickinson10}
Dickinson, M. (2010).
\newblock Detecting errors in automatically-parsed dependency relations.
\newblock In Hajic, J., Carberry, S., and Clark, S., editors, {\em {ACL} 2010,
  Proceedings of the 48th Annual Meeting of the Association for Computational
  Linguistics, July 11-16, 2010, Uppsala, Sweden}, pages 729--738. The
  Association for Computer Linguistics.

\bibitem[Dickinson and Lee, 2008]{DBLP:conf/lrec/DickinsonL08}
Dickinson, M. and Lee, C.~M. (2008).
\newblock Detecting errors in semantic annotation.
\newblock In {\em Proceedings of the International Conference on Language
  Resources and Evaluation, {LREC} 2008, 26 May - 1 June 2008, Marrakech,
  Morocco}. European Language Resources Association.

\bibitem[Dickinson and Meurers, 2003]{DBLP:conf/eacl/DickinsonM03}
Dickinson, M. and Meurers, D. (2003).
\newblock Detecting errors in part-of-speech annotation.
\newblock In {\em {EACL} 2003, 10th Conference of the European Chapter of the
  Association for Computational Linguistics, April 12-17, 2003, Agro Hotel,
  Budapest, Hungary}, pages 107--114. The Association for Computer Linguistics.

\bibitem[Dickinson and Smith, 2011]{DBLP:conf/iwpt/DickinsonS11}
Dickinson, M. and Smith, A. (2011).
\newblock Detecting dependency parse errors with minimal resources.
\newblock In {\em Proceedings of the 12th International Conference on Parsing
  Technologies, {IWPT} 2011, October 5-7, 2011, Dublin City University, Dubin,
  Ireland}, pages 241--252. The Association for Computational Linguistics.

\bibitem[Dligach and Palmer, 2011]{DBLP:conf/acllaw/DligachP11}
Dligach, D. and Palmer, M. (2011).
\newblock Reducing the need for double annotation.
\newblock In {\em Proceedings of the Fifth Linguistic Annotation Workshop,
  {LAW} 2011, June 23-24, 2011, Portland, Oregon, {USA}}, pages 65--73. The
  Association for Computer Linguistics.

\bibitem[Dobbie et~al., 2021]{DBLP:journals/fdgth/DobbieSPFJATL21}
Dobbie, S., Strafford, H., Pickrell, W.~O., Fonferko{-}Shadrach, B., Jones, C.,
  Akbari, A., Thompson, S., and Lacey, A. (2021).
\newblock Markup: {A} web-based annotation tool powered by active learning.
\newblock {\em Frontiers Digit. Health}, 3:598916.

\bibitem[Dozat and Manning, 2017]{Dozat2017DeepBA}
Dozat, T. and Manning, C.~D. (2017).
\newblock Deep biaffine attention for neural dependency parsing.
\newblock {\em ArXiv}, abs/1611.01734.

\bibitem[Dubey, 2021]{DBLP:phd/dnb/Dubey21}
Dubey, M. (2021).
\newblock {\em Towards Complex Question Answering over Knowledge Graphs}.
\newblock PhD thesis, University of Bonn, Germany.

\bibitem[Elsayed et~al., 2018]{DBLP:conf/nips/ElsayedKMRB18}
Elsayed, G.~F., Krishnan, D., Mobahi, H., Regan, K., and Bengio, S. (2018).
\newblock Large margin deep networks for classification.
\newblock In {\em Advances in Neural Information Processing Systems 31: Annual
  Conference on Neural Information Processing Systems 2018, NeurIPS 2018,
  December 3-8, 2018, Montr{\'{e}}al, Canada}, pages 850--860.

\bibitem[Eskin, 2000]{DBLP:conf/anlp/Eskin00}
Eskin, E. (2000).
\newblock Detecting errors within a corpus using anomaly detection.
\newblock In {\em 6th Applied Natural Language Processing Conference, {ANLP}
  2000, Seattle, Washington, USA, April 29 - May 4, 2000}, pages 148--153.
  {ACL}.

\bibitem[Ettinger, 2020]{DBLP:journals/tacl/Ettinger20}
Ettinger, A. (2020).
\newblock What {BERT} is not: Lessons from a new suite of psycholinguistic
  diagnostics for language models.
\newblock {\em Trans. Assoc. Comput. Linguistics}, 8:34--48.

\bibitem[Feng et~al., 2020]{DBLP:conf/emnlp/FengGTDFGS0LJZ20}
Feng, Z., Guo, D., Tang, D., Duan, N., Feng, X., Gong, M., Shou, L., Qin, B.,
  Liu, T., Jiang, D., and Zhou, M. (2020).
\newblock Codebert: {A} pre-trained model for programming and natural
  languages.
\newblock In Cohn, T., He, Y., and Liu, Y., editors, {\em Findings of the
  Association for Computational Linguistics: {EMNLP} 2020, Online Event, 16-20
  November 2020}, volume {EMNLP} 2020 of {\em Findings of {ACL}}, pages
  1536--1547. Association for Computational Linguistics.

\bibitem[Fix and Hodges, 1989]{fix1989discriminatory}
Fix, E. and Hodges, J.~L. (1989).
\newblock Discriminatory analysis. nonparametric discrimination: Consistency
  properties.
\newblock {\em International Statistical Review/Revue Internationale de
  Statistique}, 57(3):238--247.

\bibitem[Fornaciari et~al., 2021]{DBLP:conf/naacl/FornaciariUPPHP21}
Fornaciari, T., Uma, A., Paun, S., Plank, B., Hovy, D., and Poesio, M. (2021).
\newblock Beyond black {\&} white: Leveraging annotator disagreement via
  soft-label multi-task learning.
\newblock In Toutanova, K., Rumshisky, A., Zettlemoyer, L.,
  Hakkani{-}T{\"{u}}r, D., Beltagy, I., Bethard, S., Cotterell, R.,
  Chakraborty, T., and Zhou, Y., editors, {\em Proceedings of the 2021
  Conference of the North American Chapter of the Association for Computational
  Linguistics: Human Language Technologies, {NAACL-HLT} 2021, Online, June
  6-11, 2021}, pages 2591--2597. Association for Computational Linguistics.

\bibitem[Franklin, 2005]{franklin2005elements}
Franklin, J. (2005).
\newblock The elements of statistical learning: data mining, inference and
  prediction.
\newblock {\em The Mathematical Intelligencer}, 27(2):83--85.

\bibitem[Fr{\'{e}}nay and Verleysen, 2014]{DBLP:journals/tnn/FrenayV14}
Fr{\'{e}}nay, B. and Verleysen, M. (2014).
\newblock Classification in the presence of label noise: {A} survey.
\newblock {\em {IEEE} Trans. Neural Networks Learn. Syst.}, 25(5):845--869.

\bibitem[Galstyan and Cohen, 2007]{DBLP:conf/ilp/GalstyanC07}
Galstyan, A. and Cohen, P.~R. (2007).
\newblock Empirical comparison of "hard" and "soft" label propagation for
  relational classification.
\newblock In {\em Inductive Logic Programming, 17th International Conference,
  {ILP} 2007, Corvallis, OR, USA, June 19-21, 2007, Revised Selected Papers},
  pages 98--111.

\bibitem[Gao et~al., 2021a]{DBLP:conf/semweb/GaoZZFL21}
Gao, M., Zhang, S., Zhang, X., Feng, Z., and Lu, W. (2021a).
\newblock Graphs and commonsense knowledge improve the dialogue reasoning
  ability.
\newblock In Seneviratne, O., Pesquita, C., Sequeda, J., and Etcheverry, L.,
  editors, {\em Proceedings of the {ISWC} 2021 Posters, Demos and Industry
  Tracks: From Novel Ideas to Industrial Practice co-located with 20th
  International Semantic Web Conference {(ISWC} 2021), Virtual Conference,
  October 24-28, 2021}, volume 2980 of {\em {CEUR} Workshop Proceedings}.
  CEUR-WS.org.

\bibitem[Gao et~al., 2021b]{DBLP:conf/acl/GaoFC20}
Gao, T., Fisch, A., and Chen, D. (2021b).
\newblock Making pre-trained language models better few-shot learners.
\newblock In Zong, C., Xia, F., Li, W., and Navigli, R., editors, {\em
  Proceedings of the 59th Annual Meeting of the Association for Computational
  Linguistics and the 11th International Joint Conference on Natural Language
  Processing, {ACL/IJCNLP} 2021, (Volume 1: Long Papers), Virtual Event, August
  1-6, 2021}, pages 3816--3830. Association for Computational Linguistics.

\bibitem[Grishman and Sundheim, 1996]{grishman1996message}
Grishman, R. and Sundheim, B.~M. (1996).
\newblock Message understanding conference-6: A brief history.
\newblock In {\em COLING 1996 Volume 1: The 16th International Conference on
  Computational Linguistics}.

\bibitem[Grivas et~al., 2020]{DBLP:conf/acl-louhi/GrivasAGTW20}
Grivas, A., Alex, B., Grover, C., Tobin, R., and Whiteley, W. (2020).
\newblock Not a cute stroke: Analysis of rule- and neural network-based
  information extraction systems for brain radiology reports.
\newblock In Holderness, E., Jimeno{-}Yepes, A., Lavelli, A., Minard, A.,
  Pustejovsky, J., and Rinaldi, F., editors, {\em Proceedings of the 11th
  International Workshop on Health Text Mining and Information Analysis,
  LOUHI@EMNLP 2020, Online, November 20, 2020}, pages 24--37. Association for
  Computational Linguistics.

\bibitem[Han et~al., 2021]{DBLP:journals/corr/abs-2105-11259}
Han, X., Zhao, W., Ding, N., Liu, Z., and Sun, M. (2021).
\newblock {PTR:} prompt tuning with rules for text classification.
\newblock {\em CoRR}, abs/2105.11259.

\bibitem[Harris et~al., 2020]{harris2020array}
Harris, C.~R., Millman, K.~J., van~der Walt, S.~J., Gommers, R., Virtanen, P.,
  Cournapeau, D., Wieser, E., Taylor, J., Berg, S., Smith, N.~J., Kern, R.,
  Picus, M., Hoyer, S., van Kerkwijk, M.~H., Brett, M., Haldane, A., del
  R{\'{i}}o, J.~F., Wiebe, M., Peterson, P., G{\'{e}}rard-Marchant, P.,
  Sheppard, K., Reddy, T., Weckesser, W., Abbasi, H., Gohlke, C., and Oliphant,
  T.~E. (2020).
\newblock Array programming with {NumPy}.
\newblock {\em Nature}, 585(7825):357--362.

\bibitem[Haverinen et~al., 2011]{DBLP:conf/depling/HaverinenGLKVNS11}
Haverinen, K., Ginter, F., Laippala, V., Kohonen, S., Viljanen, T., Nyblom, J.,
  and Salakoski, T. (2011).
\newblock A dependency-based analysis of treebank annotation errors.
\newblock In Gerdes, K., Hajicov{\'{a}}, E., and Wanner, L., editors, {\em
  Computational Dependency Theory [papers from the International Conference on
  Dependency Linguistics, Depling 2011, Barcelona, Spain, September 2011]},
  volume 258 of {\em Frontiers in Artificial Intelligence and Applications},
  pages 47--61. {IOS} Press.

\bibitem[Hawkins, 1980]{DBLP:books/sp/Hawkins80}
Hawkins, D.~M. (1980).
\newblock {\em Identification of Outliers}.
\newblock Monographs on Applied Probability and Statistics. Springer.

\bibitem[Hayton et~al., 2000]{DBLP:conf/nips/HaytonSTA00}
Hayton, P.~M., Sch{\"{o}}lkopf, B., Tarassenko, L., and Anuzis, P. (2000).
\newblock Support vector novelty detection applied to jet engine vibration
  spectra.
\newblock In Leen, T.~K., Dietterich, T.~G., and Tresp, V., editors, {\em
  Advances in Neural Information Processing Systems 13, Papers from Neural
  Information Processing Systems {(NIPS)} 2000, Denver, CO, {USA}}, pages
  946--952. {MIT} Press.

\bibitem[He et~al., 2020]{DBLP:conf/cvpr/He0WXG20}
He, K., Fan, H., Wu, Y., Xie, S., and Girshick, R.~B. (2020).
\newblock Momentum contrast for unsupervised visual representation learning.
\newblock In {\em 2020 {IEEE/CVF} Conference on Computer Vision and Pattern
  Recognition, {CVPR} 2020, Seattle, WA, USA, June 13-19, 2020}, pages
  9726--9735.

\bibitem[H{\'{e}}naff, 2020]{DBLP:conf/icml/Henaff20}
H{\'{e}}naff, O.~J. (2020).
\newblock Data-efficient image recognition with contrastive predictive coding.
\newblock In {\em Proceedings of the 37th International Conference on Machine
  Learning, {ICML} 2020, 13-18 July 2020, Virtual Event}, pages 4182--4192.

\bibitem[Hess et~al., 2020]{DBLP:journals/corr/abs-2001-01987}
Hess, S., Duivesteijn, W., and Mocanu, D. (2020).
\newblock Softmax-based classification is k-means clustering: Formal proof,
  consequences for adversarial attacks, and improvement through centroid based
  tailoring.
\newblock {\em CoRR}, abs/2001.01987.

\bibitem[Hewitt and Manning, 2019a]{hewitt-manning-2019-structural}
Hewitt, J. and Manning, C.~D. (2019a).
\newblock {A} structural probe for finding syntax in word representations.
\newblock In {\em Proceedings of the 2019 Conference of the North {A}merican
  Chapter of the Association for Computational Linguistics: Human Language
  Technologies, Volume 1 (Long and Short Papers)}, pages 4129--4138,
  Minneapolis, Minnesota. Association for Computational Linguistics.

\bibitem[Hewitt and Manning, 2019b]{DBLP:conf/naacl/HewittM19}
Hewitt, J. and Manning, C.~D. (2019b).
\newblock A structural probe for finding syntax in word representations.
\newblock In Burstein, J., Doran, C., and Solorio, T., editors, {\em
  Proceedings of the 2019 Conference of the North American Chapter of the
  Association for Computational Linguistics: Human Language Technologies,
  {NAACL-HLT} 2019, Minneapolis, MN, USA, June 2-7, 2019, Volume 1 (Long and
  Short Papers)}, pages 4129--4138. Association for Computational Linguistics.

\bibitem[Hickey, 1996]{DBLP:journals/ai/Hickey96}
Hickey, R.~J. (1996).
\newblock Noise modelling and evaluating learning from examples.
\newblock {\em Artif. Intell.}, 82(1-2):157--179.

\bibitem[Higuchi et~al., 2021]{DBLP:conf/icassp/HiguchiSSTDD21}
Higuchi, T., Saxena, S., Souden, M., Tran, T.~D., Delfarah, M., and Dhir, C.
  (2021).
\newblock Dynamic curriculum learning via data parameters for noise robust
  keyword spotting.
\newblock In {\em {IEEE} International Conference on Acoustics, Speech and
  Signal Processing, {ICASSP} 2021, Toronto, ON, Canada, June 6-11, 2021},
  pages 6848--6852. {IEEE}.

\bibitem[Hjelm et~al., 2019]{DBLP:conf/iclr/HjelmFLGBTB19}
Hjelm, R.~D., Fedorov, A., Lavoie{-}Marchildon, S., Grewal, K., Bachman, P.,
  Trischler, A., and Bengio, Y. (2019).
\newblock Learning deep representations by mutual information estimation and
  maximization.
\newblock In {\em 7th International Conference on Learning Representations,
  {ICLR} 2019, New Orleans, LA, USA, May 6-9, 2019}.

\bibitem[Hodge and Austin, 2004]{DBLP:journals/air/HodgeA04}
Hodge, V.~J. and Austin, J. (2004).
\newblock A survey of outlier detection methodologies.
\newblock {\em Artif. Intell. Rev.}, 22(2):85--126.

\bibitem[Hoffmann, 2007]{DBLP:journals/pr/Hoffmann07}
Hoffmann, H. (2007).
\newblock Kernel {PCA} for novelty detection.
\newblock {\em Pattern Recognit.}, 40(3):863--874.

\bibitem[Hollenstein et~al., 2016]{DBLP:conf/lrec/HollensteinSW16}
Hollenstein, N., Schneider, N., and Webber, B.~L. (2016).
\newblock Inconsistency detection in semantic annotation.
\newblock In Calzolari, N., Choukri, K., Declerck, T., Goggi, S., Grobelnik,
  M., Maegaard, B., Mariani, J., Mazo, H., Moreno, A., Odijk, J., and
  Piperidis, S., editors, {\em Proceedings of the Tenth International
  Conference on Language Resources and Evaluation {LREC} 2016, Portoro{\v{z}},
  Slovenia, May 23-28, 2016}. European Language Resources Association {(ELRA)}.

\bibitem[Hovy et~al., 2013]{DBLP:conf/naacl/HovyBVH13}
Hovy, D., Berg{-}Kirkpatrick, T., Vaswani, A., and Hovy, E.~H. (2013).
\newblock Learning whom to trust with {MACE}.
\newblock In Vanderwende, L., III, H.~D., and Kirchhoff, K., editors, {\em
  Human Language Technologies: Conference of the North American Chapter of the
  Association of Computational Linguistics, Proceedings, June 9-14, 2013,
  Westin Peachtree Plaza Hotel, Atlanta, Georgia, {USA}}, pages 1120--1130. The
  Association for Computational Linguistics.

\bibitem[Jamison and Gurevych, 2015]{DBLP:conf/emnlp/JamisonG15}
Jamison, E. and Gurevych, I. (2015).
\newblock Noise or additional information? leveraging crowdsource annotation
  item agreement for natural language tasks.
\newblock In M{\`{a}}rquez, L., Callison{-}Burch, C., Su, J., Pighin, D., and
  Marton, Y., editors, {\em Proceedings of the 2015 Conference on Empirical
  Methods in Natural Language Processing, {EMNLP} 2015, Lisbon, Portugal,
  September 17-21, 2015}, pages 291--297. The Association for Computational
  Linguistics.

\bibitem[Jiang et~al., 2018]{DBLP:conf/icml/JiangZLLF18}
Jiang, L., Zhou, Z., Leung, T., Li, L., and Fei{-}Fei, L. (2018).
\newblock Mentornet: Learning data-driven curriculum for very deep neural
  networks on corrupted labels.
\newblock In Dy, J.~G. and Krause, A., editors, {\em Proceedings of the 35th
  International Conference on Machine Learning, {ICML} 2018,
  Stockholmsm{\"{a}}ssan, Stockholm, Sweden, July 10-15, 2018}, volume~80 of
  {\em Proceedings of Machine Learning Research}, pages 2309--2318. {PMLR}.

\bibitem[Jiang et~al., 2020a]{DBLP:conf/emnlp/JiangAADN20}
Jiang, Z., Anastasopoulos, A., Araki, J., Ding, H., and Neubig, G. (2020a).
\newblock {X-FACTR:} multilingual factual knowledge retrieval from pretrained
  language models.
\newblock In Webber, B., Cohn, T., He, Y., and Liu, Y., editors, {\em
  Proceedings of the 2020 Conference on Empirical Methods in Natural Language
  Processing, {EMNLP} 2020, Online, November 16-20, 2020}, pages 5943--5959.
  Association for Computational Linguistics.

\bibitem[Jiang et~al., 2020b]{DBLP:journals/tacl/JiangXAN20}
Jiang, Z., Xu, F.~F., Araki, J., and Neubig, G. (2020b).
\newblock How can we know what language models know.
\newblock {\em Trans. Assoc. Comput. Linguistics}, 8:423--438.

\bibitem[Johnson et~al., 2017]{JDH17}
Johnson, J., Douze, M., and J{\'e}gou, H. (2017).
\newblock Billion-scale similarity search with gpus.
\newblock {\em arXiv preprint arXiv:1702.08734}.

\bibitem[Joshi et~al., 2020]{joshi-etal-2020-spanbert}
Joshi, M., Chen, D., Liu, Y., Weld, D.~S., Zettlemoyer, L., and Levy, O.
  (2020).
\newblock {S}pan{BERT}: Improving pre-training by representing and predicting
  spans.
\newblock {\em Transactions of the Association for Computational Linguistics},
  8:64--77.

\bibitem[Khandelwal et~al., 2020]{DBLP:conf/iclr/KhandelwalLJZL20}
Khandelwal, U., Levy, O., Jurafsky, D., Zettlemoyer, L., and Lewis, M. (2020).
\newblock Generalization through memorization: Nearest neighbor language
  models.
\newblock In {\em 8th International Conference on Learning Representations,
  {ICLR} 2020, Addis Ababa, Ethiopia, April 26-30, 2020}.

\bibitem[Khayrallah and Koehn, 2018]{DBLP:conf/aclnmt/KhayrallahK18}
Khayrallah, H. and Koehn, P. (2018).
\newblock On the impact of various types of noise on neural machine
  translation.
\newblock In Birch, A., Finch, A.~M., Luong, M., Neubig, G., and Oda, Y.,
  editors, {\em Proceedings of the 2nd Workshop on Neural Machine Translation
  and Generation, NMT@ACL 2018, Melbourne, Australia, July 20, 2018}, pages
  74--83. Association for Computational Linguistics.

\bibitem[Khosla et~al., 2020]{DBLP:conf/nips/KhoslaTWSTIMLK20}
Khosla, P., Teterwak, P., Wang, C., Sarna, A., Tian, Y., Isola, P., Maschinot,
  A., Liu, C., and Krishnan, D. (2020).
\newblock Supervised contrastive learning.
\newblock In {\em Advances in Neural Information Processing Systems 33: Annual
  Conference on Neural Information Processing Systems 2020, NeurIPS 2020,
  December 6-12, 2020, virtual}.

\bibitem[Khurana et~al., 2017]{khurana2017natural}
Khurana, D., Koli, A., Khatter, K., and Singh, S. (2017).
\newblock Natural language processing: State of the art, current trends and
  challenges.
\newblock {\em arXiv preprint arXiv:1708.05148}.

\bibitem[Klie et~al., 2018]{tubiblio106270}
Klie, J.-C., Bugert, M., Boullosa, B., de~Castilho, R.~E., and Gurevych, I.
  (2018).
\newblock The inception platform: Machine-assisted and knowledge-oriented
  interactive annotation.
\newblock In {\em Proceedings of the 27th International Conference on
  Computational Linguistics: System Demonstrations}, pages 5--9. Association
  for Computational Linguistics.

\bibitem[K{\"u}bler et~al., 2009]{kubler2009dependency}
K{\"u}bler, S., McDonald, R., and Nivre, J. (2009).
\newblock Dependency parsing.
\newblock {\em Synthesis lectures on human language technologies}, 1(1):1--127.

\bibitem[Kusendov{\'{a}}, 2005]{DBLP:journals/glottometrics/Kusendova05}
Kusendov{\'{a}}, J. (2005).
\newblock Don mcnicol, \emph{A Primer of Signal Detection Theory.} london:
  Lawrence. erlbaum associates, publishers 2005.
\newblock {\em Glottometrics}, 9:89--90.

\bibitem[Larson et~al., 2020]{DBLP:conf/coling/LarsonCMLK20}
Larson, S., Cheung, A., Mahendran, A., Leach, K., and Kummerfeld, J.~K. (2020).
\newblock Inconsistencies in crowdsourced slot-filling annotations: {A}
  typology and identification methods.
\newblock In Scott, D., Bel, N., and Zong, C., editors, {\em Proceedings of the
  28th International Conference on Computational Linguistics, {COLING} 2020,
  Barcelona, Spain (Online), December 8-13, 2020}, pages 5035--5046.
  International Committee on Computational Linguistics.

\bibitem[Le{-}Khac et~al., 2020]{DBLP:journals/access/Le-KhacHS20}
Le{-}Khac, P.~H., Healy, G., and Smeaton, A.~F. (2020).
\newblock Contrastive representation learning: {A} framework and review.
\newblock {\em {IEEE} Access}, 8:193907--193934.

\bibitem[Lee et~al., 2020]{DBLP:journals/bioinformatics/LeeYKKKSK20}
Lee, J., Yoon, W., Kim, S., Kim, D., Kim, S., So, C.~H., and Kang, J. (2020).
\newblock Biobert: a pre-trained biomedical language representation model for
  biomedical text mining.
\newblock {\em Bioinform.}, 36(4):1234--1240.

\bibitem[Lee et~al., 2021]{DBLP:conf/iclr/LeeLH21}
Lee, S., Lee, D.~B., and Hwang, S.~J. (2021).
\newblock Contrastive learning with adversarial perturbations for conditional
  text generation.
\newblock In {\em 9th International Conference on Learning Representations,
  {ICLR} 2021, Virtual Event, Austria, May 3-7, 2021}. OpenReview.net.

\bibitem[Lester et~al., 2021]{DBLP:conf/emnlp/LesterAC21}
Lester, B., Al{-}Rfou, R., and Constant, N. (2021).
\newblock The power of scale for parameter-efficient prompt tuning.
\newblock In Moens, M., Huang, X., Specia, L., and Yih, S.~W., editors, {\em
  Proceedings of the 2021 Conference on Empirical Methods in Natural Language
  Processing, {EMNLP} 2021, Virtual Event / Punta Cana, Dominican Republic,
  7-11 November, 2021}, pages 3045--3059. Association for Computational
  Linguistics.

\bibitem[Lewis et~al., 2020]{DBLP:conf/acl/LewisLGGMLSZ20}
Lewis, M., Liu, Y., Goyal, N., Ghazvininejad, M., Mohamed, A., Levy, O.,
  Stoyanov, V., and Zettlemoyer, L. (2020).
\newblock {BART:} denoising sequence-to-sequence pre-training for natural
  language generation, translation, and comprehension.
\newblock In {\em Proceedings of the 58th Annual Meeting of the Association for
  Computational Linguistics, {ACL} 2020, Online, July 5-10, 2020}, pages
  7871--7880.

\bibitem[Li et~al., 2019]{DBLP:conf/apweb/LiWWZ019}
Li, A., Wang, X., Wang, W., Zhang, A., and Li, B. (2019).
\newblock A survey of relation extraction of knowledge graphs.
\newblock In Song, J. and Zhu, X., editors, {\em Web and Big Data - APWeb-WAIM
  2019 International Workshops, {KGMA} and DSEA, Chengdu, China, August 1-3,
  2019, Revised Selected Papers}, volume 11809 of {\em Lecture Notes in
  Computer Science}, pages 52--66. Springer.

\bibitem[Li and Liu, 2015]{DBLP:conf/hcomp/LiL15}
Li, H. and Liu, Q. (2015).
\newblock Cheaper and better: Selecting good workers for crowdsourcing.
\newblock In Gerber, E. and Ipeirotis, P., editors, {\em Proceedings of the
  Third {AAAI} Conference on Human Computation and Crowdsourcing, {HCOMP} 2015,
  November 8-11, 2015, San Diego, California, {USA}}, pages 20--21. {AAAI}
  Press.

\bibitem[Li et~al., 2020a]{DBLP:conf/icdm/LiQWYS20}
Li, P., Qin, Z., Wang, H., Yang, Q., and Shao, J. (2020a).
\newblock Exploiting inconsistency problem in multi-label classification via
  metric learning.
\newblock In Plant, C., Wang, H., Cuzzocrea, A., Zaniolo, C., and Wu, X.,
  editors, {\em 20th {IEEE} International Conference on Data Mining, {ICDM}
  2020, Sorrento, Italy, November 17-20, 2020}, pages 1100--1105. {IEEE}.

\bibitem[Li et~al., 2020b]{DBLP:journals/corr/abs-2004-03760}
Li, T., Gu, J., Zhu, X., Liu, Q., Ling, Z., Su, Z., and Wei, S. (2020b).
\newblock Dialbert: {A} hierarchical pre-trained model for conversation
  disentanglement.
\newblock {\em CoRR}, abs/2004.03760.

\bibitem[Li and Liang, 2021]{DBLP:conf/acl/LiL20}
Li, X.~L. and Liang, P. (2021).
\newblock Prefix-tuning: Optimizing continuous prompts for generation.
\newblock In Zong, C., Xia, F., Li, W., and Navigli, R., editors, {\em
  Proceedings of the 59th Annual Meeting of the Association for Computational
  Linguistics and the 11th International Joint Conference on Natural Language
  Processing, {ACL/IJCNLP} 2021, (Volume 1: Long Papers), Virtual Event, August
  1-6, 2021}, pages 4582--4597. Association for Computational Linguistics.

\bibitem[Li et~al., 2018]{li2018seq2seq}
Li, Z., Cai, J., He, S., and Zhao, H. (2018).
\newblock Seq2seq dependency parsing.
\newblock In {\em Proceedings of the 27th International Conference on
  Computational Linguistics}, pages 3203--3214.

\bibitem[Liu et~al., 2019a]{DBLP:conf/naacl/Liu0BPS19}
Liu, N.~F., Gardner, M., Belinkov, Y., Peters, M.~E., and Smith, N.~A. (2019a).
\newblock Linguistic knowledge and transferability of contextual
  representations.
\newblock In Burstein, J., Doran, C., and Solorio, T., editors, {\em
  Proceedings of the 2019 Conference of the North American Chapter of the
  Association for Computational Linguistics: Human Language Technologies,
  {NAACL-HLT} 2019, Minneapolis, MN, USA, June 2-7, 2019, Volume 1 (Long and
  Short Papers)}, pages 1073--1094. Association for Computational Linguistics.

\bibitem[Liu et~al., 2021a]{DBLP:journals/corr/abs-2107-13586}
Liu, P., Yuan, W., Fu, J., Jiang, Z., Hayashi, H., and Neubig, G. (2021a).
\newblock Pre-train, prompt, and predict: {A} systematic survey of prompting
  methods in natural language processing.
\newblock {\em CoRR}, abs/2107.13586.

\bibitem[Liu et~al., 2017]{DBLP:conf/emnlp/LiuWCS17}
Liu, T., Wang, K., Chang, B., and Sui, Z. (2017).
\newblock A soft-label method for noise-tolerant distantly supervised relation
  extraction.
\newblock In {\em Proceedings of the 2017 Conference on Empirical Methods in
  Natural Language Processing, {EMNLP} 2017, Copenhagen, Denmark, September
  9-11, 2017}, pages 1790--1795.

\bibitem[Liu et~al., 2021b]{DBLP:journals/ijon/LiuTLC21}
Liu, W., Tang, J., Liang, X., and Cai, Q. (2021b).
\newblock Heterogeneous graph reasoning for knowledge-grounded medical dialogue
  system.
\newblock {\em Neurocomputing}, 442:260--268.

\bibitem[Liu et~al., 2016]{DBLP:conf/icml/LiuWYY16}
Liu, W., Wen, Y., Yu, Z., and Yang, M. (2016).
\newblock Large-margin softmax loss for convolutional neural networks.
\newblock In {\em Proceedings of the 33nd International Conference on Machine
  Learning, {ICML} 2016, New York City, NY, USA, June 19-24, 2016}, pages
  507--516.

\bibitem[Liu et~al., 2019b]{DBLP:journals/corr/abs-1907-11692}
Liu, Y., Ott, M., Goyal, N., Du, J., Joshi, M., Chen, D., Levy, O., Lewis, M.,
  Zettlemoyer, L., and Stoyanov, V. (2019b).
\newblock Roberta: {A} robustly optimized {BERT} pretraining approach.
\newblock {\em CoRR}, abs/1907.11692.

\bibitem[Loshchilov and Hutter, 2019]{DBLP:conf/iclr/LoshchilovH19}
Loshchilov, I. and Hutter, F. (2019).
\newblock Decoupled weight decay regularization.
\newblock In {\em 7th International Conference on Learning Representations,
  {ICLR} 2019, New Orleans, LA, USA, May 6-9, 2019}. OpenReview.net.

\bibitem[Ma et~al., 2001]{DBLP:conf/icann/MaLMII01}
Ma, Q., Lu, B., Murata, M., Ichikawa, M., and Isahara, H. (2001).
\newblock On-line error detection of annotated corpus using modular neural
  networks.
\newblock In Dorffner, G., Bischof, H., and Hornik, K., editors, {\em
  Artificial Neural Networks - {ICANN} 2001, International Conference Vienna,
  Austria, August 21-25, 2001 Proceedings}, volume 2130 of {\em Lecture Notes
  in Computer Science}, pages 1185--1192. Springer.

\bibitem[Mairal, 2013]{DBLP:conf/icml/Mairal13}
Mairal, J. (2013).
\newblock Optimization with first-order surrogate functions.
\newblock In {\em Proceedings of the 30th International Conference on Machine
  Learning, {ICML} 2013, Atlanta, GA, USA, 16-21 June 2013}, volume~28 of {\em
  {JMLR} Workshop and Conference Proceedings}, pages 783--791. JMLR.org.

\bibitem[Malossini et~al., 2006]{DBLP:journals/bioinformatics/MalossiniBN06}
Malossini, A., Blanzieri, E., and Ng, R.~T. (2006).
\newblock Detecting potential labeling errors in microarrays by data
  perturbation.
\newblock {\em Bioinform.}, 22(17):2114--2121.

\bibitem[M{\`a}rquez et~al., 2008]{marquez2008semantic}
M{\`a}rquez, L., Carreras, X., Litkowski, K.~C., and Stevenson, S. (2008).
\newblock Semantic role labeling: an introduction to the special issue.

\bibitem[Matousek and Tihelka, 2017]{DBLP:journals/csl/MatousekT17}
Matousek, J. and Tihelka, D. (2017).
\newblock Anomaly-based annotation error detection in speech-synthesis corpora.
\newblock {\em Comput. Speech Lang.}, 46:1--35.

\bibitem[Matsumoto and Yamashita, 2000]{DBLP:conf/lrec/MatsumotoY00}
Matsumoto, Y. and Yamashita, T. (2000).
\newblock Using machine learning methods to improve quality of tagged corpora
  and learning models.
\newblock In {\em Proceedings of the Second International Conference on
  Language Resources and Evaluation, {LREC} 2000, 31 May - June 2, 2000,
  Athens, Greece}. European Language Resources Association.

\bibitem[McLachlan, 1999]{mclachlan1999mahalanobis}
McLachlan, G.~J. (1999).
\newblock Mahalanobis distance.
\newblock {\em Resonance}, 4(6):20--26.

\bibitem[Mikolov et~al., 2013]{NIPS2013_9aa42b31}
Mikolov, T., Sutskever, I., Chen, K., Corrado, G.~S., and Dean, J. (2013).
\newblock Distributed representations of words and phrases and their
  compositionality.
\newblock In Burges, C. J.~C., Bottou, L., Welling, M., Ghahramani, Z., and
  Weinberger, K.~Q., editors, {\em Advances in Neural Information Processing
  Systems}, volume~26. Curran Associates, Inc.

\bibitem[Min et~al., 2021]{DBLP:journals/corr/abs-2111-01243}
Min, B., Ross, H., Sulem, E., Veyseh, A. P.~B., Nguyen, T.~H., Sainz, O.,
  Agirre, E., Heintz, I., and Roth, D. (2021).
\newblock Recent advances in natural language processing via large pre-trained
  language models: {A} survey.
\newblock {\em CoRR}, abs/2111.01243.

\bibitem[Mucherino et~al., 2009]{Mucherino2009}
Mucherino, A., Papajorgji, P.~J., and Pardalos, P.~M. (2009).
\newblock {\em k-Nearest Neighbor Classification}, pages 83--106.
\newblock Springer New York, New York, NY.

\bibitem[Murphy, 2012]{DBLP:books/lib/Murphy12}
Murphy, K.~P. (2012).
\newblock {\em Machine learning - a probabilistic perspective}.
\newblock Adaptive computation and machine learning series. {MIT} Press.

\bibitem[Nakagawa and Matsumoto, 2002]{DBLP:conf/coling/NakagawaM02}
Nakagawa, T. and Matsumoto, Y. (2002).
\newblock Detecting errors in corpora using support vector machines.
\newblock In {\em 19th International Conference on Computational Linguistics,
  {COLING} 2002, Howard International House and Academia Sinica, Taipei,
  Taiwan, August 24 - September 1, 2002}.

\bibitem[Nan et~al., 2020]{DBLP:conf/acl/NanGSL20}
Nan, G., Guo, Z., Sekulic, I., and Lu, W. (2020).
\newblock Reasoning with latent structure refinement for document-level
  relation extraction.
\newblock In {\em Proceedings of the 58th Annual Meeting of the Association for
  Computational Linguistics, {ACL} 2020, Online, July 5-10, 2020}, pages
  1546--1557.

\bibitem[Nghiem et~al., 2021]{DBLP:conf/eacl/NghiemBA21}
Nghiem, M., Baylis, P., and Ananiadou, S. (2021).
\newblock Paladin: an annotation tool based on active and proactive learning.
\newblock In Gkatzia, D. and Seddah, D., editors, {\em Proceedings of the 16th
  Conference of the European Chapter of the Association for Computational
  Linguistics: System Demonstrations, {EACL} 2021, Online, April 19-23, 2021},
  pages 238--243. Association for Computational Linguistics.

\bibitem[Nguyen et~al., 2014]{DBLP:journals/jamia/NguyenVH14}
Nguyen, Q., Valizadegan, H., and Hauskrecht, M. (2014).
\newblock Learning classification models with soft-label information.
\newblock {\em J. Am. Medical Informatics Assoc.}, 21(3):501--508.

\bibitem[Nicholson et~al., 2015]{DBLP:conf/dsaa/NicholsonZSW15}
Nicholson, B., Zhang, J., Sheng, V.~S., and Wang, Z. (2015).
\newblock Label noise correction methods.
\newblock In {\em 2015 {IEEE} International Conference on Data Science and
  Advanced Analytics, {DSAA} 2015, Campus des Cordeliers, Paris, France,
  October 19-21, 2015}, pages 1--9. {IEEE}.

\bibitem[Niu et~al., 2011]{DBLP:conf/aici/NiuSSH11}
Niu, Z., Shi, S., Sun, J., and He, X. (2011).
\newblock A survey of outlier detection methodologies and their applications.
\newblock In Deng, H., Miao, D., Lei, J., and Wang, F.~L., editors, {\em
  Artificial Intelligence and Computational Intelligence - Third International
  Conference, {AICI} 2011, Taiyuan, China, September 24-25, 2011, Proceedings,
  Part {I}}, volume 7002 of {\em Lecture Notes in Computer Science}, pages
  380--387. Springer.

\bibitem[Nivre, 2005]{nivre2005dependency}
Nivre, J. (2005).
\newblock Dependency grammar and dependency parsing.
\newblock {\em MSI report}, 5133(1959):1--32.

\bibitem[Northcutt et~al., 2021a]{DBLP:journals/corr/abs-2103-14749}
Northcutt, C.~G., Athalye, A., and Mueller, J. (2021a).
\newblock Pervasive label errors in test sets destabilize machine learning
  benchmarks.
\newblock {\em CoRR}, abs/2103.14749.

\bibitem[Northcutt et~al., 2021b]{DBLP:journals/jair/NorthcuttJC21}
Northcutt, C.~G., Jiang, L., and Chuang, I.~L. (2021b).
\newblock Confident learning: Estimating uncertainty in dataset labels.
\newblock {\em J. Artif. Intell. Res.}, 70:1373--1411.

\bibitem[Nowak and R{\"{u}}ger, 2010]{DBLP:conf/mir/NowakR10}
Nowak, S. and R{\"{u}}ger, S.~M. (2010).
\newblock How reliable are annotations via crowdsourcing: a study about
  inter-annotator agreement for multi-label image annotation.
\newblock In Wang, J.~Z., Boujemaa, N., Ramirez, N.~O., and Natsev, A.,
  editors, {\em Proceedings of the 11th {ACM} {SIGMM} International Conference
  on Multimedia Information Retrieval, {MIR} 2010, Philadelphia, Pennsylvania,
  USA, March 29-31, 2010}, pages 557--566. {ACM}.

\bibitem[Oppenheimer et~al., 2009]{oppenheimer2009instructional}
Oppenheimer, D.~M., Meyvis, T., and Davidenko, N. (2009).
\newblock Instructional manipulation checks: Detecting satisficing to increase
  statistical power.
\newblock {\em Journal of experimental social psychology}, 45(4):867--872.

\bibitem[Palmer et~al., 2010]{palmer2010semantic}
Palmer, M., Gildea, D., and Xue, N. (2010).
\newblock Semantic role labeling.
\newblock {\em Synthesis Lectures on Human Language Technologies}, 3(1):1--103.

\bibitem[Parde and Nielsen, 2017]{DBLP:conf/emnlp/PardeN17}
Parde, N. and Nielsen, R.~D. (2017).
\newblock Finding patterns in noisy crowds: Regression-based annotation
  aggregation for crowdsourced data.
\newblock In Palmer, M., Hwa, R., and Riedel, S., editors, {\em Proceedings of
  the 2017 Conference on Empirical Methods in Natural Language Processing,
  {EMNLP} 2017, Copenhagen, Denmark, September 9-11, 2017}, pages 1907--1912.
  Association for Computational Linguistics.

\bibitem[Passonneau and Carpenter, 2014]{DBLP:journals/tacl/PassonneauC14}
Passonneau, R.~J. and Carpenter, B. (2014).
\newblock The benefits of a model of annotation.
\newblock {\em Trans. Assoc. Comput. Linguistics}, 2:311--326.

\bibitem[Paszke et~al., 2019]{DBLP:journals/corr/abs-1912-01703}
Paszke, A., Gross, S., Massa, F., Lerer, A., Bradbury, J., Chanan, G., Killeen,
  T., Lin, Z., Gimelshein, N., Antiga, L., Desmaison, A., K{\"{o}}pf, A., Yang,
  E.~Z., DeVito, Z., Raison, M., Tejani, A., Chilamkurthy, S., Steiner, B.,
  Fang, L., Bai, J., and Chintala, S. (2019).
\newblock Pytorch: An imperative style, high-performance deep learning library.
\newblock {\em CoRR}, abs/1912.01703.

\bibitem[Pechenizkiy et~al., 2006]{DBLP:conf/cbms/PechenizkiyTPP06}
Pechenizkiy, M., Tsymbal, A., Puuronen, S., and Pechenizkiy, O. (2006).
\newblock Class noise and supervised learning in medical domains: The effect of
  feature extraction.
\newblock In {\em 19th {IEEE} International Symposium on Computer-Based Medical
  Systems {(CBMS} 2006), 22-23 June 2006, Salt Lake City, Utah, {USA}}, pages
  708--713. {IEEE} Computer Society.

\bibitem[Pedregosa et~al., 2011]{scikit-learn}
Pedregosa, F., Varoquaux, G., Gramfort, A., Michel, V., Thirion, B., Grisel,
  O., Blondel, M., Prettenhofer, P., Weiss, R., Dubourg, V., Vanderplas, J.,
  Passos, A., Cournapeau, D., Brucher, M., Perrot, M., and Duchesnay, E.
  (2011).
\newblock Scikit-learn: Machine learning in {P}ython.
\newblock {\em Journal of Machine Learning Research}, 12:2825--2830.

\bibitem[Peng et~al., 2020]{DBLP:conf/emnlp/PengGHLLLSZ20}
Peng, H., Gao, T., Han, X., Lin, Y., Li, P., Liu, Z., Sun, M., and Zhou, J.
  (2020).
\newblock Learning from context or names? an empirical study on neural relation
  extraction.
\newblock In Webber, B., Cohn, T., He, Y., and Liu, Y., editors, {\em
  Proceedings of the 2020 Conference on Empirical Methods in Natural Language
  Processing, {EMNLP} 2020, Online, November 16-20, 2020}, pages 3661--3672.
  Association for Computational Linguistics.

\bibitem[Pennington et~al., 2014]{pennington-etal-2014-glove}
Pennington, J., Socher, R., and Manning, C. (2014).
\newblock {G}lo{V}e: Global vectors for word representation.
\newblock In {\em Proceedings of the 2014 Conference on Empirical Methods in
  Natural Language Processing ({EMNLP})}, pages 1532--1543, Doha, Qatar.
  Association for Computational Linguistics.

\bibitem[Peters et~al., 2019]{DBLP:conf/emnlp/PetersNLSJSS19}
Peters, M.~E., Neumann, M., IV, R. L.~L., Schwartz, R., Joshi, V., Singh, S.,
  and Smith, N.~A. (2019).
\newblock Knowledge enhanced contextual word representations.
\newblock In {\em Proceedings of the 2019 Conference on Empirical Methods in
  Natural Language Processing and the 9th International Joint Conference on
  Natural Language Processing, {EMNLP-IJCNLP} 2019, Hong Kong, China, November
  3-7, 2019}, pages 43--54.

\bibitem[Petroni et~al., 2019a]{petroni-etal-2019-language}
Petroni, F., Rockt{\"a}schel, T., Riedel, S., Lewis, P., Bakhtin, A., Wu, Y.,
  and Miller, A. (2019a).
\newblock Language models as knowledge bases?
\newblock In {\em Proceedings of the 2019 Conference on Empirical Methods in
  Natural Language Processing and the 9th International Joint Conference on
  Natural Language Processing (EMNLP-IJCNLP)}, pages 2463--2473, Hong Kong,
  China. Association for Computational Linguistics.

\bibitem[Petroni et~al., 2019b]{DBLP:conf/emnlp/PetroniRRLBWM19}
Petroni, F., Rockt{\"{a}}schel, T., Riedel, S., Lewis, P. S.~H., Bakhtin, A.,
  Wu, Y., and Miller, A.~H. (2019b).
\newblock Language models as knowledge bases?
\newblock In Inui, K., Jiang, J., Ng, V., and Wan, X., editors, {\em
  Proceedings of the 2019 Conference on Empirical Methods in Natural Language
  Processing and the 9th International Joint Conference on Natural Language
  Processing, {EMNLP-IJCNLP} 2019, Hong Kong, China, November 3-7, 2019}, pages
  2463--2473. Association for Computational Linguistics.

\bibitem[Portelas et~al., 2020]{DBLP:conf/ijcai/PortelasCWHO20}
Portelas, R., Colas, C., Weng, L., Hofmann, K., and Oudeyer, P. (2020).
\newblock Automatic curriculum learning for deep {RL:} {A} short survey.
\newblock In Bessiere, C., editor, {\em Proceedings of the Twenty-Ninth
  International Joint Conference on Artificial Intelligence, {IJCAI} 2020},
  pages 4819--4825. ijcai.org.

\bibitem[Qian et~al., 2021]{DBLP:conf/sigdial/QianBLDGYS21}
Qian, K., Beirami, A., Lin, Z., De, A., Geramifard, A., Yu, Z., and Sankar, C.
  (2021).
\newblock Annotation inconsistency and entity bias in multiwoz.
\newblock In Li, H., Levow, G., Yu, Z., Gupta, C., Sisman, B., Cai, S.,
  Vandyke, D., Dethlefs, N., Wu, Y., and Li, J.~J., editors, {\em Proceedings
  of the 22nd Annual Meeting of the Special Interest Group on Discourse and
  Dialogue, SIGdial 2021, Singapore and Online, July 29-31, 2021}, pages
  326--337. Association for Computational Linguistics.

\bibitem[Qiu et~al., 2020]{DBLP:journals/corr/abs-2003-08271}
Qiu, X., Sun, T., Xu, Y., Shao, Y., Dai, N., and Huang, X. (2020).
\newblock Pre-trained models for natural language processing: {A} survey.
\newblock {\em CoRR}, abs/2003.08271.

\bibitem[Radev et~al., 2002]{DBLP:conf/lrec/RadevQWF02}
Radev, D.~R., Qi, H., Wu, H., and Fan, W. (2002).
\newblock Evaluating web-based question answering systems.
\newblock In {\em Proceedings of the Third International Conference on Language
  Resources and Evaluation, {LREC} 2002, May 29-31, 2002, Las Palmas, Canary
  Islands, Spain}.

\bibitem[Raffel et~al., 2020]{DBLP:journals/jmlr/RaffelSRLNMZLL20}
Raffel, C., Shazeer, N., Roberts, A., Lee, K., Narang, S., Matena, M., Zhou,
  Y., Li, W., and Liu, P.~J. (2020).
\newblock Exploring the limits of transfer learning with a unified text-to-text
  transformer.
\newblock {\em J. Mach. Learn. Res.}, 21:140:1--140:67.

\bibitem[Raykar et~al., 2010]{DBLP:journals/jmlr/RaykarYZVFBM10}
Raykar, V.~C., Yu, S., Zhao, L.~H., Valadez, G.~H., Florin, C., Bogoni, L., and
  Moy, L. (2010).
\newblock Learning from crowds.
\newblock {\em J. Mach. Learn. Res.}, 11:1297--1322.

\bibitem[Reif et~al., 2019]{DBLP:conf/nips/ReifYWVCPK19}
Reif, E., Yuan, A., Wattenberg, M., Vi{\'{e}}gas, F.~B., Coenen, A., Pearce,
  A., and Kim, B. (2019).
\newblock Visualizing and measuring the geometry of {BERT}.
\newblock In Wallach, H.~M., Larochelle, H., Beygelzimer, A.,
  d'Alch{\'{e}}{-}Buc, F., Fox, E.~B., and Garnett, R., editors, {\em Advances
  in Neural Information Processing Systems 32: Annual Conference on Neural
  Information Processing Systems 2019, NeurIPS 2019, December 8-14, 2019,
  Vancouver, BC, Canada}, pages 8592--8600.

\bibitem[Reiss et~al., 2020]{DBLP:conf/conll/ReissXCME20}
Reiss, F., Xu, H., Cutler, B., Muthuraman, K., and Eichenberger, Z. (2020).
\newblock Identifying incorrect labels in the conll-2003 corpus.
\newblock In Fern{\'{a}}ndez, R. and Linzen, T., editors, {\em Proceedings of
  the 24th Conference on Computational Natural Language Learning, CoNLL 2020,
  Online, November 19-20, 2020}, pages 215--226. Association for Computational
  Linguistics.

\bibitem[Roit et~al., 2020]{DBLP:conf/acl/RoitKSMMSZD20}
Roit, P., Klein, A., Stepanov, D., Mamou, J., Michael, J., Stanovsky, G.,
  Zettlemoyer, L., and Dagan, I. (2020).
\newblock Controlled crowdsourcing for high-quality {QA-SRL} annotation.
\newblock In Jurafsky, D., Chai, J., Schluter, N., and Tetreault, J.~R.,
  editors, {\em Proceedings of the 58th Annual Meeting of the Association for
  Computational Linguistics, {ACL} 2020, Online, July 5-10, 2020}, pages
  7008--7013. Association for Computational Linguistics.

\bibitem[Saffari et~al., 2021]{DBLP:conf/emnlp/SaffariOSA21}
Saffari, A., Oliya, A., Sen, P., and Ayoola, T. (2021).
\newblock End-to-end entity resolution and question answering using
  differentiable knowledge graphs.
\newblock In Moens, M., Huang, X., Specia, L., and Yih, S.~W., editors, {\em
  Proceedings of the 2021 Conference on Empirical Methods in Natural Language
  Processing, {EMNLP} 2021, Virtual Event / Punta Cana, Dominican Republic,
  7-11 November, 2021}, pages 4193--4200. Association for Computational
  Linguistics.

\bibitem[Saunshi et~al., 2021]{DBLP:conf/iclr/SaunshiMA21}
Saunshi, N., Malladi, S., and Arora, S. (2021).
\newblock A mathematical exploration of why language models help solve
  downstream tasks.
\newblock In {\em 9th International Conference on Learning Representations,
  {ICLR} 2021, Virtual Event, Austria, May 3-7, 2021}. OpenReview.net.

\bibitem[Sch{\"{o}}lkopf et~al., 2001]{DBLP:journals/neco/ScholkopfPSSW01}
Sch{\"{o}}lkopf, B., Platt, J.~C., Shawe{-}Taylor, J., Smola, A.~J., and
  Williamson, R.~C. (2001).
\newblock Estimating the support of a high-dimensional distribution.
\newblock {\em Neural Comput.}, 13(7):1443--1471.

\bibitem[Sch{\"{o}}lkopf et~al., 1999]{DBLP:conf/nips/ScholkopfWSSP99}
Sch{\"{o}}lkopf, B., Williamson, R.~C., Smola, A.~J., Shawe{-}Taylor, J., and
  Platt, J.~C. (1999).
\newblock Support vector method for novelty detection.
\newblock In Solla, S.~A., Leen, T.~K., and M{\"{u}}ller, K., editors, {\em
  Advances in Neural Information Processing Systems 12, {[NIPS} Conference,
  Denver, Colorado, USA, November 29 - December 4, 1999]}, pages 582--588. The
  {MIT} Press.

\bibitem[Sculley and Cormack, 2008]{DBLP:conf/ceas/SculleyC08}
Sculley, D. and Cormack, G.~V. (2008).
\newblock Filtering email spam in the presence of noisy user feedback.
\newblock In {\em {CEAS} 2008 - The Fifth Conference on Email and Anti-Spam,
  21-22 August 2008, Mountain View, California, {USA}}.

\bibitem[Sebert, 1997]{sebert1997outliers}
Sebert, D.~M. (1997).
\newblock Outliers in statistical data.
\newblock {\em Journal of Quality Technology}, 29(2):230.

\bibitem[Sellam et~al., 2020]{DBLP:conf/acl/SellamDP20}
Sellam, T., Das, D., and Parikh, A.~P. (2020).
\newblock {BLEURT:} learning robust metrics for text generation.
\newblock In Jurafsky, D., Chai, J., Schluter, N., and Tetreault, J.~R.,
  editors, {\em Proceedings of the 58th Annual Meeting of the Association for
  Computational Linguistics, {ACL} 2020, Online, July 5-10, 2020}, pages
  7881--7892. Association for Computational Linguistics.

\bibitem[Sen et~al., 2021]{DBLP:conf/emnlp/SenOS21}
Sen, P., Oliya, A., and Saffari, A. (2021).
\newblock Expanding end-to-end question answering on differentiable knowledge
  graphs with intersection.
\newblock In Moens, M., Huang, X., Specia, L., and Yih, S.~W., editors, {\em
  Proceedings of the 2021 Conference on Empirical Methods in Natural Language
  Processing, {EMNLP} 2021, Virtual Event / Punta Cana, Dominican Republic,
  7-11 November, 2021}, pages 8805--8812. Association for Computational
  Linguistics.

\bibitem[Sharou et~al., 2021]{DBLP:conf/ranlp/SharouLS21}
Sharou, K.~A., Li, Z., and Specia, L. (2021).
\newblock Towards a better understanding of noise in natural language
  processing.
\newblock In Angelova, G., Kunilovskaya, M., Mitkov, R., and Nikolova{-}Koleva,
  I., editors, {\em Proceedings of the International Conference on Recent
  Advances in Natural Language Processing {(RANLP} 2021), Held Online,
  1-3September, 2021}, pages 53--62. {INCOMA} Ltd.

\bibitem[Shin et~al., 2020]{shin-etal-2020-autoprompt}
Shin, T., Razeghi, Y., Logan~IV, R.~L., Wallace, E., and Singh, S. (2020).
\newblock {A}uto{P}rompt: {E}liciting {K}nowledge from {L}anguage {M}odels with
  {A}utomatically {G}enerated {P}rompts.
\newblock In {\em Proceedings of the 2020 Conference on Empirical Methods in
  Natural Language Processing (EMNLP)}, pages 4222--4235, Online. Association
  for Computational Linguistics.

\bibitem[Smyth, 1996]{DBLP:journals/prl/Smyth96}
Smyth, P. (1996).
\newblock Bounds on the mean classification error rate of multiple experts.
\newblock {\em Pattern Recognit. Lett.}, 17(12):1253--1257.

\bibitem[Snoek et~al., 2012]{snoek2012practical}
Snoek, J., Larochelle, H., and Adams, R.~P. (2012).
\newblock Practical bayesian optimization of machine learning algorithms.
\newblock {\em Advances in neural information processing systems}, 25.

\bibitem[Snow et~al., 2008]{DBLP:conf/emnlp/SnowOJN08}
Snow, R., O'Connor, B., Jurafsky, D., and Ng, A.~Y. (2008).
\newblock Cheap and fast - but is it good? evaluating non-expert annotations
  for natural language tasks.
\newblock In {\em 2008 Conference on Empirical Methods in Natural Language
  Processing, {EMNLP} 2008, Proceedings of the Conference, 25-27 October 2008,
  Honolulu, Hawaii, USA, {A} meeting of SIGDAT, a Special Interest Group of the
  {ACL}}, pages 254--263. {ACL}.

\bibitem[Soares et~al., 2019]{DBLP:conf/acl/SoaresFLK19}
Soares, L.~B., FitzGerald, N., Ling, J., and Kwiatkowski, T. (2019).
\newblock Matching the blanks: Distributional similarity for relation learning.
\newblock In Korhonen, A., Traum, D.~R., and M{\`{a}}rquez, L., editors, {\em
  Proceedings of the 57th Conference of the Association for Computational
  Linguistics, {ACL} 2019, Florence, Italy, July 28- August 2, 2019, Volume 1:
  Long Papers}, pages 2895--2905. Association for Computational Linguistics.

\bibitem[Sohrab, 2003]{sohrab2003basic}
Sohrab, H. (2003).
\newblock {\em Basic Real Analysis}.
\newblock Birkh{\"a}user Boston.

\bibitem[Soviany et~al., 2021]{DBLP:journals/corr/abs-2101-10382}
Soviany, P., Ionescu, R.~T., Rota, P., and Sebe, N. (2021).
\newblock Curriculum learning: {A} survey.
\newblock {\em CoRR}, abs/2101.10382.

\bibitem[Stoica et~al., 2021]{DBLP:conf/aaai/StoicaPP21}
Stoica, G., Platanios, E.~A., and P{\'{o}}czos, B. (2021).
\newblock Re-tacred: Addressing shortcomings of the {TACRED} dataset.
\newblock In {\em Thirty-Fifth {AAAI} Conference on Artificial Intelligence,
  {AAAI} 2021, Thirty-Third Conference on Innovative Applications of Artificial
  Intelligence, {IAAI} 2021, The Eleventh Symposium on Educational Advances in
  Artificial Intelligence, {EAAI} 2021, Virtual Event, February 2-9, 2021},
  pages 13843--13850. {AAAI} Press.

\bibitem[Stojnic and Risojevic, 2021]{DBLP:conf/cvpr/StojnicR21}
Stojnic, V. and Risojevic, V. (2021).
\newblock Self-supervised learning of remote sensing scene representations
  using contrastive multiview coding.
\newblock In {\em {IEEE} Conference on Computer Vision and Pattern Recognition
  Workshops, {CVPR} Workshops 2021, virtual, June 19-25, 2021}, pages
  1182--1191.

\bibitem[Stone, 1977]{stone1977asymptotic}
Stone, M. (1977).
\newblock An asymptotic equivalence of choice of model by cross-validation and
  akaike's criterion.
\newblock {\em Journal of the Royal Statistical Society: Series B
  (Methodological)}, 39(1):44--47.

\bibitem[Sukhbaatar et~al., 2015]{sukhbaatar2015training}
Sukhbaatar, S., Bruna, J., Paluri, M., Bourdev, L., and Fergus, R. (2015).
\newblock Training convolutional networks with noisy labels.
\newblock In {\em 3rd International Conference on Learning Representations,
  ICLR 2015}.

\bibitem[Suzuki et~al., 2017]{DBLP:journals/ieicet/SuzukiKM17}
Suzuki, K., Kato, Y., and Matsubara, S. (2017).
\newblock Correcting syntactic annotation errors based on tree mining.
\newblock {\em {IEICE} Trans. Inf. Syst.}, 100-D(5):1106--1113.

\bibitem[Talmor et~al., 2020]{DBLP:journals/tacl/TalmorEGB20}
Talmor, A., Elazar, Y., Goldberg, Y., and Berant, J. (2020).
\newblock olmpics - on what language model pre-training captures.
\newblock {\em Trans. Assoc. Comput. Linguistics}, 8:743--758.

\bibitem[Tang et~al., 2020]{DBLP:conf/pakdd/TangC0CFWY20}
Tang, H., Cao, Y., Zhang, Z., Cao, J., Fang, F., Wang, S., and Yin, P. (2020).
\newblock {HIN:} hierarchical inference network for document-level relation
  extraction.
\newblock In {\em Advances in Knowledge Discovery and Data Mining - 24th
  Pacific-Asia Conference, {PAKDD} 2020, Singapore, May 11-14, 2020,
  Proceedings, Part {I}}, pages 197--209.

\bibitem[Taylor and Gollwitzer, 1995]{taylor1995effects}
Taylor, S.~E. and Gollwitzer, P.~M. (1995).
\newblock Effects of mindset on positive illusions.
\newblock {\em Journal of personality and social psychology}, 69(2):213.

\bibitem[Tenney et~al., 2019]{DBLP:conf/acl/TenneyDP19}
Tenney, I., Das, D., and Pavlick, E. (2019).
\newblock {BERT} rediscovers the classical {NLP} pipeline.
\newblock In Korhonen, A., Traum, D.~R., and M{\`{a}}rquez, L., editors, {\em
  Proceedings of the 57th Conference of the Association for Computational
  Linguistics, {ACL} 2019, Florence, Italy, July 28- August 2, 2019, Volume 1:
  Long Papers}, pages 4593--4601. Association for Computational Linguistics.

\bibitem[Thiel, 2008]{DBLP:conf/kes/Thiel08}
Thiel, C. (2008).
\newblock Classification on soft labels is robust against label noise.
\newblock In {\em Knowledge-Based Intelligent Information and Engineering
  Systems, 12th International Conference, {KES} 2008, Zagreb, Croatia,
  September 3-5, 2008, Proceedings, Part {I}}, pages 65--73.

\bibitem[Tibshirani, 1996]{tibshirani1996journal}
Tibshirani, R. (1996).
\newblock Journal of the royal statistical society. series b (methodological).

\bibitem[Tsuzuku et~al., 2018]{NEURIPS2018_48584348}
Tsuzuku, Y., Sato, I., and Sugiyama, M. (2018).
\newblock Lipschitz-margin training: Scalable certification of perturbation
  invariance for deep neural networks.
\newblock In Bengio, S., Wallach, H., Larochelle, H., Grauman, K.,
  Cesa-Bianchi, N., and Garnett, R., editors, {\em Advances in Neural
  Information Processing Systems}, volume~31. Curran Associates, Inc.

\bibitem[Vafaeikia et~al., 2020]{DBLP:journals/corr/abs-2007-01126}
Vafaeikia, P., Namdar, K., and Khalvati, F. (2020).
\newblock A brief review of deep multi-task learning and auxiliary task
  learning.
\newblock {\em CoRR}, abs/2007.01126.

\bibitem[van~den Oord et~al., 2018]{DBLP:journals/corr/abs-1807-03748}
van~den Oord, A., Li, Y., and Vinyals, O. (2018).
\newblock Representation learning with contrastive predictive coding.
\newblock {\em CoRR}, abs/1807.03748.

\bibitem[van~der Maaten and Hinton, 2008]{JMLR:v9:vandermaaten08a}
van~der Maaten, L. and Hinton, G. (2008).
\newblock Visualizing data using t-sne.
\newblock {\em Journal of Machine Learning Research}, 9(86):2579--2605.

\bibitem[van Rooyen et~al., 2015]{DBLP:conf/nips/RooyenMW15}
van Rooyen, B., Menon, A.~K., and Williamson, R.~C. (2015).
\newblock Learning with symmetric label noise: The importance of being
  unhinged.
\newblock In Cortes, C., Lawrence, N.~D., Lee, D.~D., Sugiyama, M., and
  Garnett, R., editors, {\em Advances in Neural Information Processing Systems
  28: Annual Conference on Neural Information Processing Systems 2015, December
  7-12, 2015, Montreal, Quebec, Canada}, pages 10--18.

\bibitem[Vaswani et~al., 2017]{DBLP:conf/nips/VaswaniSPUJGKP17}
Vaswani, A., Shazeer, N., Parmar, N., Uszkoreit, J., Jones, L., Gomez, A.~N.,
  Kaiser, L., and Polosukhin, I. (2017).
\newblock Attention is all you need.
\newblock In Guyon, I., von Luxburg, U., Bengio, S., Wallach, H.~M., Fergus,
  R., Vishwanathan, S. V.~N., and Garnett, R., editors, {\em Advances in Neural
  Information Processing Systems 30: Annual Conference on Neural Information
  Processing Systems 2017, December 4-9, 2017, Long Beach, CA, {USA}}, pages
  5998--6008.

\bibitem[Wang et~al., 2021]{DBLP:conf/icsim/WangLYQ21}
Wang, H., Lu, G., Yin, J., and Qin, K. (2021).
\newblock Relation extraction: {A} brief survey on deep neural network based
  methods.
\newblock In Li, Y. and Nishi, H., editors, {\em {ICSIM} 2021: 2021 The 4th
  International Conference on Software Engineering and Information Management,
  Yokohama Japan, January 16-18, 2021}, pages 220--228. {ACM}.

\bibitem[Wang et~al., 2020]{DBLP:journals/corr/abs-2010-13166}
Wang, X., Chen, Y., and Zhu, W. (2020).
\newblock A comprehensive survey on curriculum learning.
\newblock {\em CoRR}, abs/2010.13166.

\bibitem[Wang et~al., 2019]{DBLP:conf/emnlp/WangSLLLH19}
Wang, Z., Shang, J., Liu, L., Lu, L., Liu, J., and Han, J. (2019).
\newblock Crossweigh: Training named entity tagger from imperfect annotations.
\newblock In {\em Proceedings of the 2019 Conference on Empirical Methods in
  Natural Language Processing and the 9th International Joint Conference on
  Natural Language Processing, {EMNLP-IJCNLP} 2019, Hong Kong, China, November
  3-7, 2019}, pages 5153--5162.

\bibitem[Weeber et~al., 2021]{DBLP:journals/corr/abs-2112-11914}
Weeber, F., Hamborg, F., Donnay, K., and Gipp, B. (2021).
\newblock Assisted text annotation using active learning to achieve high
  quality with little effort.
\newblock {\em CoRR}, abs/2112.11914.

\bibitem[Winkens et~al., 2020]{DBLP:journals/corr/abs-2007-05566}
Winkens, J., Bunel, R., Roy, A.~G., Stanforth, R., Natarajan, V., Ledsam,
  J.~R., MacWilliams, P., Kohli, P., Karthikesalingam, A., Kohl, S., Cemgil,
  A.~T., Eslami, S. M.~A., and Ronneberger, O. (2020).
\newblock Contrastive training for improved out-of-distribution detection.
\newblock {\em CoRR}, abs/2007.05566.

\bibitem[Wolf et~al., 2020]{wolf-etal-2020-transformers}
Wolf, T., Debut, L., Sanh, V., Chaumond, J., Delangue, C., Moi, A., Cistac, P.,
  Rault, T., Louf, R., Funtowicz, M., Davison, J., Shleifer, S., von Platen,
  P., Ma, C., Jernite, Y., Plu, J., Xu, C., Le~Scao, T., Gugger, S., Drame, M.,
  Lhoest, Q., and Rush, A. (2020).
\newblock Transformers: State-of-the-art natural language processing.
\newblock In {\em Proceedings of the 2020 Conference on Empirical Methods in
  Natural Language Processing: System Demonstrations}, pages 38--45, Online.
  Association for Computational Linguistics.

\bibitem[Wu et~al., 2016]{DBLP:journals/corr/WuSCLNMKCGMKSJL16}
Wu, Y., Schuster, M., Chen, Z., Le, Q.~V., Norouzi, M., Macherey, W., Krikun,
  M., Cao, Y., Gao, Q., Macherey, K., Klingner, J., Shah, A., Johnson, M., Liu,
  X., Kaiser, L., Gouws, S., Kato, Y., Kudo, T., Kazawa, H., Stevens, K.,
  Kurian, G., Patil, N., Wang, W., Young, C., Smith, J., Riesa, J., Rudnick,
  A., Vinyals, O., Corrado, G., Hughes, M., and Dean, J. (2016).
\newblock Google's neural machine translation system: Bridging the gap between
  human and machine translation.
\newblock {\em CoRR}, abs/1609.08144.

\bibitem[Wu et~al., 2021]{DBLP:conf/aaai/WuSX0M21}
Wu, Y., Shu, J., Xie, Q., Zhao, Q., and Meng, D. (2021).
\newblock Learning to purify noisy labels via meta soft label corrector.
\newblock In {\em Thirty-Fifth {AAAI} Conference on Artificial Intelligence,
  {AAAI} 2021, Thirty-Third Conference on Innovative Applications of Artificial
  Intelligence, {IAAI} 2021, The Eleventh Symposium on Educational Advances in
  Artificial Intelligence, {EAAI} 2021, Virtual Event, February 2-9, 2021},
  pages 10388--10396. {AAAI} Press.

\bibitem[Wu et~al., 2018]{DBLP:conf/cvpr/WuXYL18}
Wu, Z., Xiong, Y., Yu, S.~X., and Lin, D. (2018).
\newblock Unsupervised feature learning via non-parametric instance
  discrimination.
\newblock In {\em 2018 {IEEE} Conference on Computer Vision and Pattern
  Recognition, {CVPR} 2018, Salt Lake City, UT, USA, June 18-22, 2018}, pages
  3733--3742.

\bibitem[Xu et~al., 2021]{DBLP:conf/aaai/XuWLZM21}
Xu, B., Wang, Q., Lyu, Y., Zhu, Y., and Mao, Z. (2021).
\newblock Entity structure within and throughout: Modeling mention dependencies
  for document-level relation extraction.
\newblock In {\em Thirty-Fifth {AAAI} Conference on Artificial Intelligence,
  {AAAI} 2021, Thirty-Third Conference on Innovative Applications of Artificial
  Intelligence, {IAAI} 2021, The Eleventh Symposium on Educational Advances in
  Artificial Intelligence, {EAAI} 2021, Virtual Event, February 2-9, 2021},
  pages 14149--14157.

\bibitem[Yang et~al., 2019]{DBLP:conf/nips/YangDYCSL19}
Yang, Z., Dai, Z., Yang, Y., Carbonell, J.~G., Salakhutdinov, R., and Le, Q.~V.
  (2019).
\newblock Xlnet: Generalized autoregressive pretraining for language
  understanding.
\newblock In {\em Advances in Neural Information Processing Systems 32: Annual
  Conference on Neural Information Processing Systems 2019, NeurIPS 2019,
  December 8-14, 2019, Vancouver, BC, Canada}, pages 5754--5764.

\bibitem[Yao et~al., 2019a]{DBLP:journals/corr/abs-1909-03193}
Yao, L., Mao, C., and Luo, Y. (2019a).
\newblock {KG-BERT:} {BERT} for knowledge graph completion.
\newblock {\em CoRR}, abs/1909.03193.

\bibitem[Yao et~al., 2019b]{yao-etal-2019-docred}
Yao, Y., Ye, D., Li, P., Han, X., Lin, Y., Liu, Z., Liu, Z., Huang, L., Zhou,
  J., and Sun, M. (2019b).
\newblock {D}oc{RED}: A large-scale document-level relation extraction dataset.
\newblock In {\em Proceedings of the 57th Annual Meeting of the Association for
  Computational Linguistics}, pages 764--777, Florence, Italy. Association for
  Computational Linguistics.

\bibitem[Ye et~al., 2020]{ye-etal-2020-coreferential}
Ye, D., Lin, Y., Du, J., Liu, Z., Li, P., Sun, M., and Liu, Z. (2020).
\newblock {C}oreferential {R}easoning {L}earning for {L}anguage
  {R}epresentation.
\newblock In {\em Proceedings of the 2020 Conference on Empirical Methods in
  Natural Language Processing (EMNLP)}, pages 7170--7186, Online. Association
  for Computational Linguistics.

\bibitem[Ye et~al., 2021]{DBLP:conf/aaai/YeZDCTHC21}
Ye, H., Zhang, N., Deng, S., Chen, M., Tan, C., Huang, F., and Chen, H. (2021).
\newblock Contrastive triple extraction with generative transformer.
\newblock In {\em Thirty-Fifth {AAAI} Conference on Artificial Intelligence,
  {AAAI} 2021, Thirty-Third Conference on Innovative Applications of Artificial
  Intelligence, {IAAI} 2021, The Eleventh Symposium on Educational Advances in
  Artificial Intelligence, {EAAI} 2021, Virtual Event, February 2-9, 2021},
  pages 14257--14265. {AAAI} Press.

\bibitem[Yuen et~al., 2011]{DBLP:conf/socialcom/YuenKL11}
Yuen, M., King, I., and Leung, K. (2011).
\newblock A survey of crowdsourcing systems.
\newblock In {\em PASSAT/SocialCom 2011, Privacy, Security, Risk and Trust
  (PASSAT), 2011 {IEEE} Third International Conference on and 2011 {IEEE} Third
  International Conference on Social Computing (SocialCom), Boston, MA, USA,
  9-11 Oct., 2011}, pages 766--773. {IEEE} Computer Society.

\bibitem[Zhang et~al., 2015]{DBLP:conf/cikm/ZhangSWFW15}
Zhang, J., Sheng, V.~S., Wu, J., Fu, X., and Wu, X. (2015).
\newblock Improving label quality in crowdsourcing using noise correction.
\newblock In Bailey, J., Moffat, A., Aggarwal, C.~C., de~Rijke, M., Kumar, R.,
  Murdock, V., Sellis, T.~K., and Yu, J.~X., editors, {\em Proceedings of the
  24th {ACM} International Conference on Information and Knowledge Management,
  {CIKM} 2015, Melbourne, VIC, Australia, October 19 - 23, 2015}, pages
  1931--1934. {ACM}.

\bibitem[Zhang et~al., 2020]{DBLP:conf/iclr/ZhangKWWA20}
Zhang, T., Kishore, V., Wu, F., Weinberger, K.~Q., and Artzi, Y. (2020).
\newblock Bertscore: Evaluating text generation with {BERT}.
\newblock In {\em 8th International Conference on Learning Representations,
  {ICLR} 2020, Addis Ababa, Ethiopia, April 26-30, 2020}. OpenReview.net.

\bibitem[Zhang et~al., 2018]{zhang-etal-2018-graph}
Zhang, Y., Qi, P., and Manning, C.~D. (2018).
\newblock Graph convolution over pruned dependency trees improves relation
  extraction.
\newblock In {\em Proceedings of the 2018 Conference on Empirical Methods in
  Natural Language Processing}, pages 2205--2215, Brussels, Belgium.
  Association for Computational Linguistics.

\bibitem[Zhang and Yang, 2017]{DBLP:journals/corr/ZhangY17aa}
Zhang, Y. and Yang, Q. (2017).
\newblock A survey on multi-task learning.
\newblock {\em CoRR}, abs/1707.08114.

\bibitem[Zhang et~al., 2017a]{zhang2017tacred}
Zhang, Y., Zhong, V., Chen, D., Angeli, G., and Manning, C.~D. (2017a).
\newblock Position-aware attention and supervised data improve slot filling.
\newblock In {\em Proceedings of the 2017 Conference on Empirical Methods in
  Natural Language Processing (EMNLP 2017)}, pages 35--45.

\bibitem[Zhang et~al., 2017b]{YuhaoZhang2017PositionawareAA}
Zhang, Y., Zhong, V., Chen, D., Angeli, G., and Manning, C.~D. (2017b).
\newblock Position-aware attention and supervised data improve slot filling.
\newblock In {\em Empirical Methods in Natural Language Processing}.

\bibitem[Zhang et~al., 2019]{DBLP:conf/acl/ZhangHLJSL19}
Zhang, Z., Han, X., Liu, Z., Jiang, X., Sun, M., and Liu, Q. (2019).
\newblock {ERNIE:} enhanced language representation with informative entities.
\newblock In Korhonen, A., Traum, D.~R., and M{\`{a}}rquez, L., editors, {\em
  Proceedings of the 57th Conference of the Association for Computational
  Linguistics, {ACL} 2019, Florence, Italy, July 28- August 2, 2019, Volume 1:
  Long Papers}, pages 1441--1451. Association for Computational Linguistics.

\bibitem[Zhang and Sabuncu, 2018]{DBLP:conf/nips/ZhangS18}
Zhang, Z. and Sabuncu, M.~R. (2018).
\newblock Generalized cross entropy loss for training deep neural networks with
  noisy labels.
\newblock In {\em Advances in Neural Information Processing Systems 31: Annual
  Conference on Neural Information Processing Systems 2018, NeurIPS 2018,
  December 3-8, 2018, Montr{\'{e}}al, Canada}, pages 8792--8802.

\bibitem[Zhao et~al., 2014]{DBLP:journals/nn/ZhaoZCL14}
Zhao, M., Zhang, Z., Chow, T. W.~S., and Li, B. (2014).
\newblock A general soft label based linear discriminant analysis for
  semi-supervised dimensionality reduction.
\newblock {\em Neural Networks}, 55:83--97.

\bibitem[Zheng et~al., 2021]{DBLP:conf/aaai/ZhengAD21}
Zheng, G., Awadallah, A.~H., and Dumais, S.~T. (2021).
\newblock Meta label correction for noisy label learning.
\newblock In {\em Thirty-Fifth {AAAI} Conference on Artificial Intelligence,
  {AAAI} 2021, Thirty-Third Conference on Innovative Applications of Artificial
  Intelligence, {IAAI} 2021, The Eleventh Symposium on Educational Advances in
  Artificial Intelligence, {EAAI} 2021, Virtual Event, February 2-9, 2021},
  pages 11053--11061. {AAAI} Press.

\bibitem[Zhong and Chen, 2020]{ZexuanZhong2020AFE}
Zhong, Z. and Chen, D. (2020).
\newblock A frustratingly easy approach for entity and relation extraction.
\newblock {\em arXiv: Computation and Language}.

\bibitem[Zhong et~al., 2021]{DBLP:conf/naacl/ZhongFC21}
Zhong, Z., Friedman, D., and Chen, D. (2021).
\newblock Factual probing is {[MASK]:} learning vs. learning to recall.
\newblock In Toutanova, K., Rumshisky, A., Zettlemoyer, L.,
  Hakkani{-}T{\"{u}}r, D., Beltagy, I., Bethard, S., Cotterell, R.,
  Chakraborty, T., and Zhou, Y., editors, {\em Proceedings of the 2021
  Conference of the North American Chapter of the Association for Computational
  Linguistics: Human Language Technologies, {NAACL-HLT} 2021, Online, June
  6-11, 2021}, pages 5017--5033. Association for Computational Linguistics.

\bibitem[Zhou et~al., 2021a]{DBLP:conf/iclr/ZhouWB21}
Zhou, T., Wang, S., and Bilmes, J.~A. (2021a).
\newblock Robust curriculum learning: from clean label detection to noisy label
  self-correction.
\newblock In {\em 9th International Conference on Learning Representations,
  {ICLR} 2021, Virtual Event, Austria, May 3-7, 2021}. OpenReview.net.

\bibitem[Zhou and Chen, 2021]{DBLP:journals/corr/abs-2102-01373}
Zhou, W. and Chen, M. (2021).
\newblock An improved baseline for sentence-level relation extraction.
\newblock {\em CoRR}, abs/2102.01373.

\bibitem[Zhou et~al., 2020]{WenxuanZhou2020DocumentLevelRE}
Zhou, W., Huang, K., Ma, T., and Huang, J. (2020).
\newblock Document-level relation extraction with adaptive thresholding and
  localized context pooling.
\newblock {\em arXiv: Computation and Language}.

\bibitem[Zhou et~al., 2021b]{DBLP:conf/emnlp/Zhou0C21}
Zhou, W., Liu, F., and Chen, M. (2021b).
\newblock Contrastive out-of-distribution detection for pretrained
  transformers.
\newblock In Moens, M., Huang, X., Specia, L., and Yih, S.~W., editors, {\em
  Proceedings of the 2021 Conference on Empirical Methods in Natural Language
  Processing, {EMNLP} 2021, Virtual Event / Punta Cana, Dominican Republic,
  7-11 November, 2021}, pages 1100--1111. Association for Computational
  Linguistics.

\bibitem[Zhu et~al., 2015]{Zhu_2015_ICCV}
Zhu, Y., Kiros, R., Zemel, R., Salakhutdinov, R., Urtasun, R., Torralba, A.,
  and Fidler, S. (2015).
\newblock Aligning books and movies: Towards story-like visual explanations by
  watching movies and reading books.
\newblock In {\em The IEEE International Conference on Computer Vision (ICCV)}.

\bibitem[Zou et~al., 2021]{DBLP:conf/sigir/ZouZHS21}
Zou, X., Zhang, Z., He, Z., and Shi, L. (2021).
\newblock Unsupervised ensemble learning with noisy label correction.
\newblock In Diaz, F., Shah, C., Suel, T., Castells, P., Jones, R., and Sakai,
  T., editors, {\em {SIGIR} '21: The 44th International {ACM} {SIGIR}
  Conference on Research and Development in Information Retrieval, Virtual
  Event, Canada, July 11-15, 2021}, pages 2308--2312. {ACM}.

\end{thebibliography}

\end{document}